\documentclass[sigconf, nonacm, table]{acmart}
\newcommand\vldbdoi{10.14778/3749646.3749682}
\newcommand\vldbpages{4131 - 4143}
\newcommand\vldbvolume{18}
\newcommand\vldbissue{11}
\newcommand\vldbyear{2025}
\newcommand\vldbauthors{\authors}
\newcommand\vldbtitle{\shorttitle} 
\newcommand\vldbavailabilityurl{https://github.com/DataResponsibly/ShaRP}

\newcommand\vldbpagestyle{empty} 

\usepackage{algorithm}
\usepackage{algorithmic}
\usepackage{amsmath, bbm}
\usepackage[normalem]{ulem}
\usepackage{multirow}

\usepackage{balance}
\usepackage{newtxmath}  % or mathptmx
\usepackage{dsfont}

\usepackage{color}
\usepackage{subfig}
\usepackage{xcolor}
\usepackage{colortbl}
\usepackage{xspace}

\usepackage{pdfpages}
\usepackage{stfloats}
\usepackage{makecell}

\newcommand{\queryScore}[1]{Y_{#1}}
\newcommand*{\ie}{i.e.,\xspace}
\newcommand*{\eg}{e.g.,\xspace}

\newcommand*{\sys}{ShaRP~\xspace}
\newcommand{\candidateSet}[1]{\mathcal D_{\mathit{#1}}}

\def\btau{\boldsymbol{\tau}}

\newcommand*{\ranking}[1]{\langle #1 \rangle}

\newcommand*{\lst}[2][n] {#2_1, \ldots, #2_{#1}}

\def\e#1{{\em #1}}
\def\val#1{\texttt{#1}}
\def\angs#1{\mathord{\langle #1 \rangle}}

\newcommand*{\grpRank}{Rank-Group\xspace}
\newcommand*{\grpScore}{Score-Group\xspace}
\newcommand*{\csr}{CSRankings\xspace}

\newtheorem{example}{Example}

\def\val#1{\texttt{#1}}

\begin{document}

\title{ShaRP: Explaining Rankings and Preferences with Shapley Values}

\author{Venetia Pliatsika}
\authornote{Both authors contributed equally to this research.}
\affiliation{%
  \institution{New York University}
  \city{New York, NY}
  \country{USA}}
\email{venetia@nyu.edu}

\author{Joao Fonseca}
\authornotemark[1]
\affiliation{%
  \institution{New York University}
  \city{New York, NY}
  \country{USA}}
\email{jpm9748@nyu.edu}

\author{Kateryna Akhynko}
\affiliation{%
 \institution{Ukrainian Catholic University}
 \city{Lviv}
 \country{Ukraine}}
\email{kateryna.akhynko@ucu.edu.ua}

\author{Ivan Shevchenko}
\affiliation{%
 \institution{Ukrainian Catholic University}
 \city{Lviv}
 \country{Ukraine}}
\email{ivan.shevchenko@ucu.edu.ua}

\author{Julia Stoyanovich}
\affiliation{%
  \institution{New York University}
  \city{New York, NY}
  \country{USA}}
\email{stoyanovich@nyu.edu}

\begin{abstract}
Algorithmic decisions in critical domains such as hiring, college admissions, and lending are often based on rankings. Given the impact of these decisions on individuals, organizations, and population groups, it is essential to understand them—to help individuals improve their ranking position, design better ranking procedures, and ensure legal compliance.  In this paper, we argue that explainability methods for classification and regression, such as SHAP, are insufficient for ranking tasks, and present ShaRP---Shapley Values for Rankings and Preferences---a framework that explains the contributions of features to various aspects of a ranked outcome. 

ShaRP computes feature contributions for various ranking-specific profit functions, such as rank and top-$k$, and also includes a novel Shapley value-based method for explaining pairwise preference outcomes.  We provide a flexible implementation of ShaRP, capable of efficiently and comprehensively explaining ranked and pairwise outcomes over tabular data, in score-based ranking and learning-to-rank tasks.  Finally, we develop a comprehensive evaluation methodology for ranking explainability methods, showing through qualitative, quantitative, and usability studies that our rank-aware QoIs offer complementary insights, scale effectively, and help users interpret ranked outcomes in practice.

\end{abstract}



\maketitle

\pagestyle{\vldbpagestyle}
\begingroup\small\noindent\raggedright\textbf{PVLDB Reference Format:}\\
\vldbauthors. \vldbtitle. PVLDB, \vldbvolume(\vldbissue): \vldbpages, \vldbyear.\\
\href{https://doi.org/\vldbdoi}{doi:\vldbdoi}
\endgroup
\begingroup
\renewcommand\thefootnote{}\footnote{\noindent
This work is licensed under the Creative Commons BY-NC-ND 4.0 International License. Visit \url{https://creativecommons.org/licenses/by-nc-nd/4.0/} to view a copy of this license. For any use beyond those covered by this license, obtain permission by emailing \href{mailto:info@vldb.org}{info@vldb.org}. Copyright is held by the owner/author(s). Publication rights licensed to the VLDB Endowment. \\
\raggedright Proceedings of the VLDB Endowment, Vol. \vldbvolume, No. \vldbissue\ %
ISSN 2150-8097. \\
\href{https://doi.org/\vldbdoi}{doi:\vldbdoi} \\
}\addtocounter{footnote}{-1}\endgroup

\ifdefempty{\vldbavailabilityurl}{}{
\vspace{.3cm}
\begingroup\small\noindent\raggedright\textbf{PVLDB Artifact Availability:}\\
The source code, data, and/or other artifacts have been made available at \url{\vldbavailabilityurl}.
\endgroup
}

\section{Introduction}
\label{sec:intro}
Rankings produced by data-driven algorithmic systems now influence a myriad of socio-technical applications, as part of automated or semi-automated decision-making, and with direct consequences to people's lives and aspirations. An \emph{algorithmic ranker}, or a \emph{ranker} for short, takes a database of candidates as input and produces a permutation of these candidates as output, see Figure~\ref{fig:admissions} for an example. We refer to the output of a ranker as a \emph{ranked outcome} or simply a \emph{ranking}. As an alternative to the full permutation, the best-ranked $k$ candidates, or the top-$k$, may be returned in rank order or as a set. In the latter case, we are dealing with a \emph{selection} task, which is a special case of ranking.  

Algorithmic rankers are broadly used to support decision-making in critical domains, including hiring and employment, school and college admissions, credit and lending, and, of course, college ranking.  Because of the impact rankers have on individuals, organizations, and population groups, there is a need to understand them: to know whether the decisions are correct and legally compliant (\emph{auditing} tasks), to help individuals improve their ranked outcomes (\emph{recourse} tasks), and to design better ranking procedures (\emph{design} tasks).  To make progress towards these tasks, we need ways to explain and interpret ranked outcomes.  In this paper, we present \emph{\sys}~---\emph{Shapley for Rankings and Preferences}---a framework that explains the contributions of features to different aspects of a ranked outcome, and that can support all these critically important tasks. 

There are two types of rankers: score-based and learned. In score-based ranking, a given set of candidates is sorted on a score, which is typically computed using a simple formula, such as a weighted sum of attribute values~\cite{10.1145/3533379}. In supervised learning-to-rank (LtR), a preference-enriched set of candidates is used to train a model that predicts rankings of unseen candidates~\cite{DBLP:series/synthesis/2014Li}. We motivate our work using score-based rankers and return to LtR later in the paper.

Score-based rankers are often seen as ``interpretable models''~\cite{rudin2019stop}: their scoring functions, such as $\queryScore{1} = 0.9 \times gpa + 0.1 \times essay$ in a college admissions setting, reflect a normative, a priori notion of merit. For instance, specifying $\queryScore{1}$ asserts that $gpa$ matters more than the essay, while $\queryScore{2} = 0.1 \times gpa + 0.9 \times essay$ asserts the opposite. Yet the apparent transparency---and sense of \emph{control over outcomes}---that such rankers afford is often misleading. Even with full knowledge of the formula, designers or decision-makers may struggle to anticipate or explain its output~\cite{miller2019explanation,molnar2020interpretable}. We illustrate this with an example.

\newcommand{\dataquerytab}{
    \small
	\begin{tabular}{|c||c|c|c||c|c|}
		\hline
		\rowcolor[HTML]{C0C0C0} 
		name  & gpa & sat & essay & $f$ & $g$  \\ \hline
		\val{Bob}  & 4 & 5 & 5  & 4.6 & 5 \\ \hline
		\val{Cal}  & 4 & 5 & 5  & 4.6 & 5 \\ \hline
		\val{Dia}  & 5 & 4 & 4  & 4.4 & 4 \\ \hline
		\val{Eli}  & 4 & 5 & 3  & 4.2 & 3 \\ \hline
		\val{Fay}  & 5 & 4 & 3  & 4.2 & 3 \\ \hline
		\val{Kat}  & 5 & 4 & 2  & 4.0 & 2 \\ \hline
		\val{Leo}  & 4 & 4 & 3  & 3.8 & 3 \\ \hline
		\val{Osi}  & 3 & 3 & 3  & 3.0 & 3 \\ \hline
	\end{tabular}
}
\newcommand{\rankscoretab}{
    \small
	\begin{tabular}{|c|}
		\hline
		\rowcolor[HTML]{C0C0C0} 
		$r_{\mathcal{D},f}$ \\ \hline
		\rowcolor[HTML]{EFEFEF} 
		\val{Bob}                    \\ \hline
		\rowcolor[HTML]{EFEFEF} 
		\val{Cal}                    \\ \hline
		\rowcolor[HTML]{EFEFEF} 
		\val{Dia}                    \\ \hline
		\rowcolor[HTML]{EFEFEF} 
		\val{Eli}                    \\ \hline
		\val{Fay}                    \\ \hline
		\val{Kat}                    \\ \hline
		\val{Leo}                    \\ \hline
		\val{Osi}                    \\ \hline
	\end{tabular}
}
\newcommand{\rankstemtab}{
    \small
	\begin{tabular}{|c|}
		\hline
		\rowcolor[HTML]{C0C0C0} 
		$r_{\mathcal{D},g}$ \\ \hline
		\rowcolor[HTML]{EFEFEF} 
		\val{Bob}                    \\ \hline
		\rowcolor[HTML]{EFEFEF} 
		\val{Cal}                    \\ \hline
		\rowcolor[HTML]{EFEFEF} 
		\val{Dia}                    \\ \hline
		\rowcolor[HTML]{EFEFEF} 
		\val{Eli}                    \\ \hline
		\val{Fay}                    \\ \hline
		\val{Leo}                    \\ \hline
		\val{Osi}                    \\ \hline
		\val{Kat}                    \\ \hline
	\end{tabular}
}
\newcommand{\ranknonstemtab}{
    \small
	\begin{tabular}{|c|}
		\hline
		\rowcolor[HTML]{C0C0C0} 
		$\btau_3$ \\ \hline
		\rowcolor[HTML]{EFEFEF} 
	     \val{Dia}       \\ \hline
		\rowcolor[HTML]{EFEFEF} 
         \val{Fay}       \\ \hline
		\rowcolor[HTML]{EFEFEF} 
	      \val{Bob}             \\ \hline
		\rowcolor[HTML]{EFEFEF} 
	       \val{Cal}           \\ \hline
	       \val{Kat}            \\ \hline
	        \val{Eli}          \\ \hline
	      \val{Leo}            \\ \hline
	      \val{Osi}            \\ \hline
	\end{tabular}
}

\newcommand{\rankracefair}{
    \small
	\begin{tabular}{|c|}
		\hline
		\rowcolor[HTML]{C0C0C0} 
		$\btau_3^{*}$ \\ \hline
		\rowcolor[HTML]{EFEFEF} 
	     \val{Dia}       \\ \hline
		\rowcolor[HTML]{EFEFEF} 
         \val{Bob}       \\ \hline
		\rowcolor[HTML]{EFEFEF} 
	      \val{Cal}             \\ \hline
		\rowcolor[HTML]{EFEFEF} 
	       \val{Kat}           \\ \hline
	       \val{Fay}            \\ \hline
	        \val{Eli}          \\ \hline
	      \val{Leo}            \\ \hline
	      \val{Osi}            \\ \hline
	\end{tabular}
}

\begin{figure}[t!]
    \centering
    \setlength{\tabcolsep}{0.3em}
	\subfloat[]
	{\dataquerytab
	 \label{fig:ad_example_queries:data}}
	\hfill
	\subfloat[]{
		\rankscoretab
		\label{fig:ad_example_queries:ranking_score}}
	\hfill
	\subfloat[]{
		\rankstemtab
		\label{fig:ad_example_queries:ranking_stem}
    }
    \caption{\small (a) Dataset  $\mathcal{D}$ of college applicants, scored on $gpa$, $sat$, and $essay$.~(b) Ranking $r_{\mathcal{D},f}$ of $\mathcal{D}$ on $f=0.4 \times gpa + 0.4 \times sat + 0.2 \times essay$; the highlighted top-4 candidates will be interviewed and potentially admitted.~(c) Ranking $r_{\mathcal{D},g}$ on $g=1.0 \times essay$; the top-$4$ coincides with that of $r_{\mathcal{D},f}$, signifying that $essay$ has the highest importance for $f$, despite carrying the lowest weight in the scoring function.}
    \label{fig:admissions}
\end{figure}

\begin{example}
Consider a dataset $\mathcal{D}$ of college applicants in Figure~\ref{fig:admissions}, with scoring features $gpa$, $sat$, and $essay$. Very different scoring functions $f=0.4 \times gpa + 0.4 \times sat + 0.2 \times essay$ and $g = 1.0 \times essay$ induce very similar rankings $r_{\mathcal{D},f}$ and $r_{\mathcal{D},g}$, with the same top-$4$ items appearing in the same order, apparently because $essay$ is the feature that is best able to discriminate between the top-$4$ and the rest, and that determines the relative order among the top-$4$.
\label{ex:different-rankers}
\end{example}

This example illustrates that ``intrinsically interpretable'' score-based rankers do not always yield explainable outcomes. Even when both the formula and the dataset are fully known, it may be difficult to accurately anticipate how individual features influence the final ranking~\cite{miller2019explanation,molnar2020interpretable}. This disconnect arises because a feature's weight in the scoring function does not necessarily correspond to its practical influence on the ranked outcome. For example, if $gpa$ and $sat$ scores are highly correlated, while $essay$ scores are more variable and less correlated with the others, the $essay$ component may exert disproportionate influence on rank positions \emph{despite having lower nominal weight}. Conversely, a heavily weighted feature might have little effect if its values are tightly clustered across candidates.

An additional nuance in ranking is that outcomes are inherently \emph{relative}, whereas feature values and computed scores are \emph{absolute}—an item’s score reveals little about its position \emph{relative to others}. The \emph{lack of independence between per-item outcomes} makes feature importance methods developed for classification and regression~\cite{DBLP:conf/nips/CovertLL20,DBLP:conf/sp/DattaSZ16,DBLP:journals/csur/GuidottiMRTGP19,lundberg2017unified,mohler2020learning,DBLP:conf/kdd/Ribeiro0G16,strumbelj2010efficient} inadequate for ranking. These methods evaluate how a feature affects an item's score, but a feature can shift the score without altering the rank. Consider an example.

\begin{example}
Consider Figure~\ref{fig:admissions} and suppose that Dia's essay score increases from 4 to 5, thus increasing the scores computed with both $f$ (4.4 to 4.6) and $g$ (4 to 5).  However, Dia's rank remains unchanged.
\end{example}

\begin{figure}[t!]
    \centering
    \setlength{\tabcolsep}{0.3em}
        \subfloat[Score QoI: Interdisciplinary is the most important feature, followed by AI and Systems, while Theory negatively impacts the score.]
	{\includegraphics[width=0.7\columnwidth]{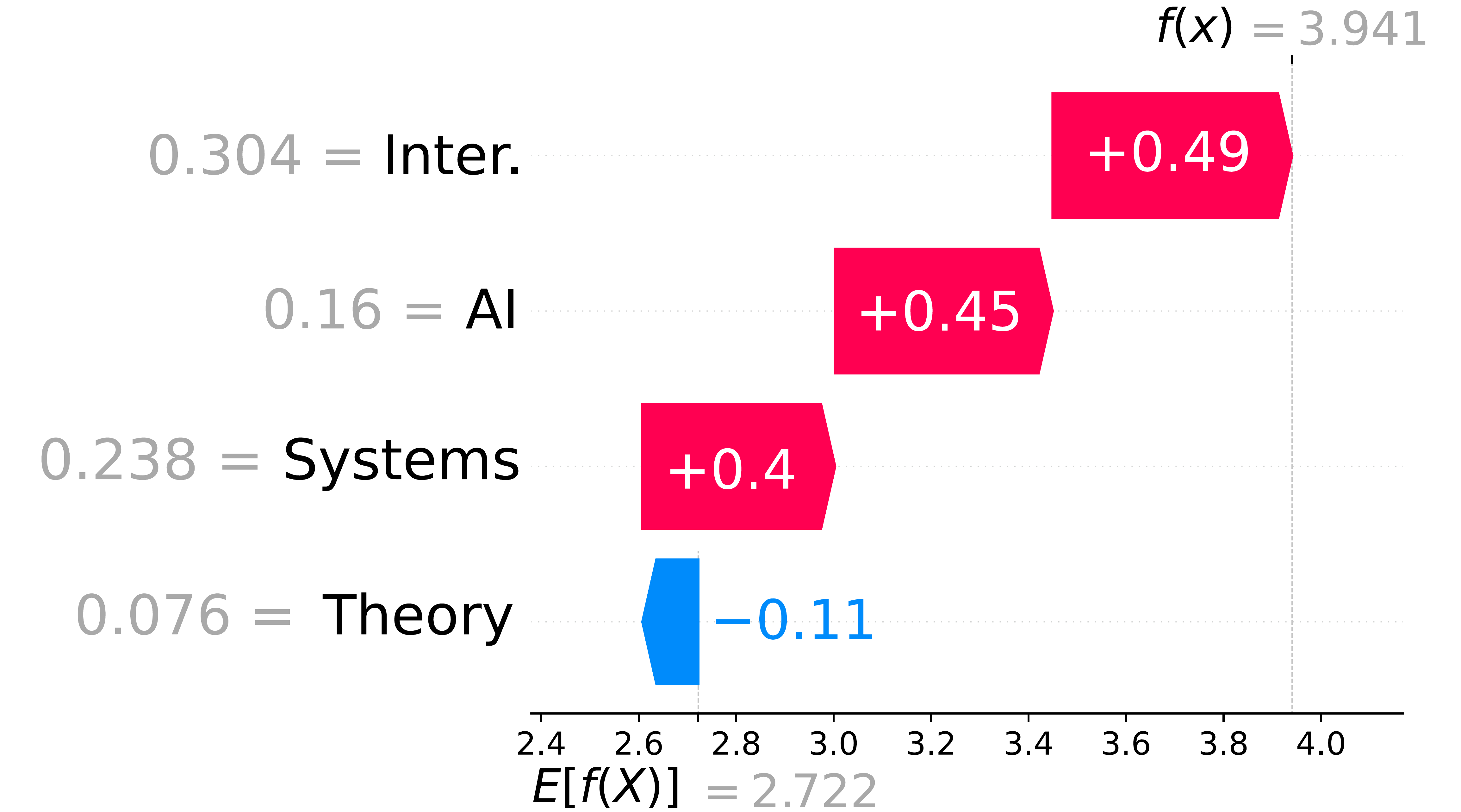}
	 \label{fig:csrankings/UT/score}}
	\hfill
	\subfloat[Rank QoI: Systems is the most important feature, followed by AI and Interdisciplinary. Theory is minimally but positively impacting the rank.]{
		\includegraphics[width=0.7\columnwidth]{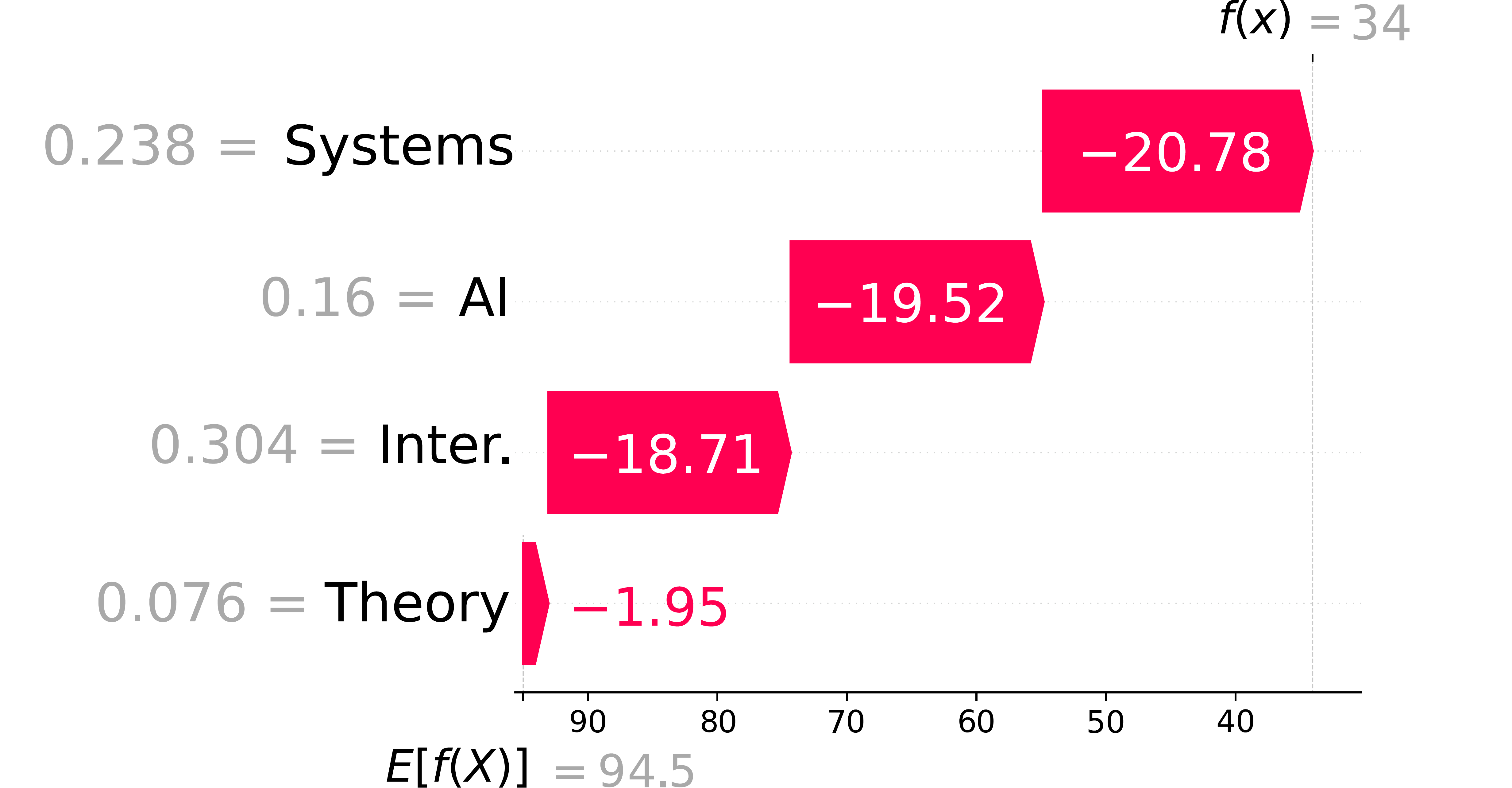}
		\label{fig:csrankings/UT/rank}}
    \caption{\small Feature importance for Texas A\&M in CS Rankings.} %The order of feature contributions and their relative magnitudes are different because score quantifies which features impact the score while rank quantifies which features impact the score to the degree that the rank changes. The sign of the contribution is determined compared to the average score and rank respectively. Score is heavily influenced by outliers while rank is not.}
    \vspace{-1.5em}
    \label{fig:CSR_waterfalls}
\end{figure}

Changes in score do not necessarily lead to changes in rank because, in selection and ranking, an item's outcome $\textbf{v}$ depends on the outcomes of other items in $\mathcal{D} \setminus \{\textbf{v}\}$. For example, only one item can occupy a given rank, and exactly $k$ items can appear in the top-$k$. Thus, any explainability method that measures score changes can only partially explain rank changes.  
This highlights that \textit{interpretability for ranking tasks requires measuring the features' impact on quantities beyond the score}, such as rank or top-$k$ presence. We preview these results for CS Rankings in Figure~\ref{fig:CSR_waterfalls}, where feature importance for score in~\ref{fig:csrankings/UT/score} and rank in~\ref{fig:csrankings/UT/rank} yield markedly different explanations. We discuss these findings in detail in Section~\ref{sec:qoi}.

\emph{In summary,} ranking differs fundamentally from classification and regression, as noted in learning-to-rank and fairness-in-ranking work~\cite{DBLP:series/synthesis/2014Li,10.1145/3533379,10.1145/3533380}. Interpretability methods must also be tailored to ranking, where scoring feature influence must account for the \emph{interdependence of item outcomes}. We formalize and build on this insight, making four contributions.

\emph{First,} we formalize several profit functions for computing Shapley values in ranking, capturing feature contributions to an item's score, rank, or top-$k$ presence. Building on the QII framework~\cite{DBLP:conf/sp/DattaSZ16}, which applies Shapley values~\cite{shapley1953} to classification, we adopt QII as a flexible foundation for defining ranked \emph{Quantities of Interest (QoIs)}.

\emph{Second,} we propose a Shapley-based method for explaining pairwise outcomes. Unlike prior methods that use a fixed baseline~\cite{lundberg2017unified,chen2023algorithms}, we adapt the baseline dynamically for each pair $u \prec v$, yielding explanations that reflect relative differences.

\emph{Third,} we release \sys—the first open-source library for explaining ranked outcomes over tabular data. \sys supports both score-based and learned rankers, includes exact and approximate QoI computation, and incorporates optimizations for scalability.

\emph{Fourth,} we evaluate ranking explainability methods through qualitative, quantitative, and usability studies. Using established metrics, we show that rank-aware QoIs provide complementary insights beyond score-based explanations. A large-scale evaluation confirms the scalability and effectiveness of our methods, while a CS Rankings usability study shows it helps users make sense of ranked outcomes.

\section{Related Work}
\label{sec:related}

\paragraph{Local feature-based explanations} 
\citet{DBLP:conf/kdd/Ribeiro0G16} introduced LIME, which explains classifiers using local interpretable models. \citet{lundberg2017unified} proposed SHAP, which uses Shapley values to explain predictions of classification and regression models. Both are implemented in software libraries and explain an item’s score---what we refer to as the score QoI.

\emph{Feature-based explanations for ranking.}
\citet{DBLP:conf/sigmod/YangSAHJM18} introduced a ``nutritional label'' for score-based rankers with two global explanation widgets: ``Recipe'' (scoring feature weights) and ``Ingredients'' (features with strongest rank-score correlation). They observed that a feature's weight often does not align with its correlation, highlighting the limits of global explanations. In contrast, we focus on local explanations for individual items or item pairs.

\citet{DBLP:journals/pvldb/GaleM20} proposed ``participation metrics'' for score-based rankers, notably ``weighted participation,'' which attributes an item’s presence in the top-$k$ to its features, weights, and  values. Their method aggregates over all top-$k$ items; ours provides per-item explanations using the top-$k$ QoI, which can be aggregated.

\citet{DBLP:conf/hilda/YuanD23} designed a sensitivity analysis tool for synthetic data with linear scoring, using mean-centered feature differences to approximate Shapley values. We re-implemented and extended their method to support arbitrary distributions, more features, and flexible scoring functions.

\citet{Anahideh2022} used local SHAP-based explanations for items near the one being explained, assuming rank stability across repeated competitions. While we also observe rank-stratum-specific feature effects, we show that small feature changes can cause large rank shifts, challenging their locality assumption.

\citet{Dexer2023} introduced DEXER to detect group disparities in top-$k$ inclusion and explained causes via SHAP on ranks fitted by linear regression. In contrast, \sys fully adapts Shapley values to rank-specific QoIs. We compare  with DEXER in Section~\ref{sec:exp:other}.

\citet{pastor2021identifying} used ranking-based profit functions to detect under- or overrepresented groups via attribute-level contributions, focusing on group fairness rather than individual explanations.

\citet{DBLP:conf/nips/HuCHS22} proposed PrefSHAP to explain pairwise preferences in learned rankers, transforming item pairs into artificial items and applying Shapley analysis. We share the motivation for ranking-specific QoIs but target preferences induced by score-based rankers or LtR, not kernel-based preference models as in PrefSHAP.

\emph{Shapley-based explanations in Information Retrieval (IR).} Concurrently with our work,~\citet{heuss_rankingshap} and~\citet{chowdhuryrankshap} proposed Shapley-based methods for explaining ranked outcomes in IR. Both compute feature contributions for the \textit{entire ranking} by perturbing all items simultaneously for each coalition. These methods are not applicable to settings that require explanations on a \textit{per-item basis} (\eg lending or hiring). In particular,~\citet{chowdhuryrankshap} define a profit function tied to query-specific rank-relevance, limiting generality. In contrast, our method supports per-item explanations while accounting for the interdependence of outcomes, using a general profit function that yields feature attributions analogous to SHAP in classification and regression.

Other recent work in IR explored the use of LIME to explain ranked outcomes~\cite{LIRME,singh2018EXS,Chowdhury_ranklime}, and introduced baseline document construction techniques to improve explanation quality~\cite{Fernando_2019}.

\emph{In summary,} we share motivation with these lines of work but take a leap by presenting the first comprehensive Shapley-value-based framework for explaining rankings and pairwise preferences.

\section{Preliminaries and Notation}
\label{sec:prelim}

\paragraph*{Ranking.} Let $\mathcal{A}$ denote an ordered collection of features (equiv. attributes), and let $\candidateSet{}$ denote a set of items (equiv. points or candidates).  An item $\textbf{v} = (v_1, \dots, v_d) \in \mathbb{R}^d$ assigns values to $| \mathcal{A} | = d$ features, and may additionally be associated with a score.  Score-based rankers use a scoring function $f(\textbf{v})$ to compute the score of \textbf{v}.  For example, using $f_1(\textbf{v})=0.4 \times gpa + 0.4 \times sat + 0.2 \times essay$, we compute $f(\val{Bob})=4.6$ and $f(\val{Leo})=3.8$. 

A \e{ranking} $r_\mathcal{D}$ is a permutation over the items in $\candidateSet{}$.  Letting $n = \left| \candidateSet{} \right|$, we denote by $r_\mathcal{D} = \ranking{\lst{\textbf{v}}}$ a ranking that places item $\textbf{v}_i$ at rank $i$.  We denote by $r_\mathcal{D}(i)$ the item at rank $i$, and by $r_\mathcal{D}^{-1}(\textbf{v})$ the rank of item \textbf{v} in $r_\mathcal{D}$.  In score-based ranking, we are interested in rankings induced by some scoring function $f$.  We denote these rankings $r_{\mathcal{D},f}$. For example, in Figure~\ref{fig:ad_example_queries:ranking_score},  $r_{\mathcal{D},f}(1)=\val{Bob}$, $r_{\mathcal{D},f}^{-1}(\val{Leo})=7$. We assume that $r_{\mathcal{D},f}^{-1}(\textbf{v}_1) < r_{\mathcal{D},f}^{-1}(\textbf{v}_2) < \dots < r_{\mathcal{D},f}^{-1}(\textbf{v}_n)$, where smaller rank means better position in the ranking.

We are often interested in a sub-ranking of $r_{\mathcal{D},f}$ containing its best-ranked $k$ items, for some integer $k \leq n$, called the top-$k$.  The top-$4$ of the ranking in Figure~\ref{fig:ad_example_queries:ranking_score} is $\angs{\val{Bob},\val{Cal},\val{Dia},\val{Eli}}$.

Our goal is to explain the importance of features  $\mathcal{A}$ to the ranking $r_{\mathcal{D},f}$.  We will do so using Shapley values~\cite{shapley1953}.

\paragraph{Shapley values} For a set $\mathcal{N}$ of $n$ players, and a value function $f$ that assigns a profit to any subset (or coalition) $\mathcal{S}$ of players, $f: 2^n \rightarrow \mathbb{R}$, where $f(\emptyset) = 0$, the Shapley value of player $i$ is:
\begin{equation}
\phi_i(f) = \sum_{\mathcal{S}} \frac{|\mathcal{S}|! (n-|\mathcal{S}|-1)!}{n!}(f(\mathcal{S}\cup\{i\})-f(\mathcal{S}))
\label{eq:shapley}
\end{equation}

We will use Shapley values to explain ranked outcomes using the set of features $\mathcal{A}$ as the players, and the outcome (or the quantity of interest, QoI) as the payoff function. 
In addition to the definition of players and the payoff function,  
Shapley values require the quantification of the payoff over a subset of the players. This, in turn, requires some way to estimate the payoff over a subset of the features. Consequently, for any Shapley value implementation, a method of feature removal or masking is required~\cite{Covert2021Removing,chen2023algorithms}.

A common method (\eg used in SHAP~\cite{lundberg2017unified, Covert2021Removing}), for a \textit{coalition} (subset of features) $\mathcal{S} \subseteq \mathcal{A}$, is to marginalize out the features not in the coalition $\mathcal{A} \setminus \mathcal{S}$ and draw values from the marginal distributions of the subset of features in $\mathcal{S}$ jointly, often referred to as the ``marginal'' approach. Another alternative (\eg used in QII~\cite{DBLP:conf/sp/DattaSZ16}) is to draw values of each feature in $\mathcal{S}$ independently from its marginal distribution, often referred to as the ``product of marginals'' approach. Another approach is called ``baseline'' and instead of sampling the features not in the coalition, they are replaced with the feature values of a specific fixed sample~\cite{lundberg2017unified}. Here, we choose the marginal approach for our implementation and take inspiration from the baseline approach for our pairwise method. In Section~\ref{sec:sys}, we show how both can be implemented using one algorithm.

Let $\textbf{v}_{\mathcal{S}}$ denote a projection of $\textbf{v}$ onto $\mathcal{S}$.  In the example in Figure~\ref{fig:admissions},  $(\val{Bob},\val{4},\val{5}, \val{5})_{\{name,gpa\}}=(\val{Bob},\val{4})$. We define a random variable $\textbf{U}$ that draws values from the marginal distributions of the subset of features in $\mathcal{S}$. Let \textbf{U} $= \langle \textbf{u}_1, \dots, \textbf{u}_m \rangle$ denote a vector of $m$ items sampled from  $\mathcal{D}$ using this method. For a subset of features $\mathcal{S} \in \mathcal{A}$, let $\textbf{v}_{\mathcal{A} \setminus \mathcal{S}}\textbf{U}_{\mathcal{S}}= \langle \textbf{v}_{\mathcal{A} \setminus \mathcal{S}}(\textbf{u}_1)_{\mathcal{S}}, \dots, \textbf{v}_{\mathcal{A} \setminus \mathcal{S}}(\textbf{u}_m)_{\mathcal{S}} \rangle$ denote a vector of items, in which each $\textbf{v}_{\mathcal{A} \setminus \mathcal{S}}(\textbf{u}_i)_{\mathcal{S}}$ takes on the values of the features in $\mathcal{S}$ from $\textbf{u}_i$, and the values of the remaining features $\mathcal{A} \setminus \mathcal{S}$ from $\textbf{v}$. We calculate Shapley values using this set of features $\textbf{v}_{\mathcal{A} \setminus \mathcal{S}}\textbf{U}_{\mathcal{S}}$, note that if $m=|\mathcal{D}-1|$ we use the entire dataset $\mathcal{D}\setminus \textbf{v}$ to calculate the \textit{exact} Shapley values.

Shapley values satisfy several natural axioms, including efficiency, symmetry, dummy, and additivity~\cite{shapley1953}, with additional useful properties, such as monotonicity, following from these axioms\cite{lundberg2017unified}.  Efficiency states that the sum of the contributions of all features for item $\textbf{v}$ equals the difference between the outcome $f(\textbf{v})$ and the average outcome: $\sum_{i\in \mathcal{A}}\varphi_i(f,\textbf{v}) = f(\textbf{v}) - \mathbb{E}_{\textbf{X}}[f(\textbf{X})]$~\cite{DBLP:conf/nips/CovertLL20,molnar2020interpretable}.  Using this property, explanation can be used to reconstruct the outcome. 
We will use the efficiency property to define the fidelity metric for comparing explanations (Section~\ref{sec:metrics}).

\section{Quantities of Interest for Ranking} 
\label{sec:qoi}
The first contribution of our work is that we define QoIs that are appropriate for ranked outcomes. In addition to the expected score, we introduce rank and top-$k$ QoI. We use the notation for the marginal feature removal approach in this section, but note that the QoIs we introduce can be used with any feature removal approach.

\paragraph{Score QoI}
The Shapley value function for the score QoI is:
\begin{equation}
QoI_{f,\textbf{v}}(S) = \mathop{\mathbb{E}}_{\textbf{U}_{\mathcal{S}}} [f(\textbf{v}_{\mathcal{A} \setminus \mathcal{S}}\textbf{U}_{\mathcal{S}})]
\label{eq:qoi-score}
\end{equation}

This QoI captures the impact of an item's features on its score. This is the QoI used by the popular feature-based explanation methods such as SHAP~\cite{lundberg2017unified} and LIME~\cite{ribeiro2016modelagnostic}. To get the contribution of a set of features $\mathcal{A} \setminus \mathcal{S}$, we take the expected value of the score over a random variable $\textbf{U}_{\mathcal{S}}$ that draws values from the marginal distributions of the set of features in $\mathcal{S}$.

\begin{figure*}[t!]
\includegraphics[width=1.7\columnwidth]{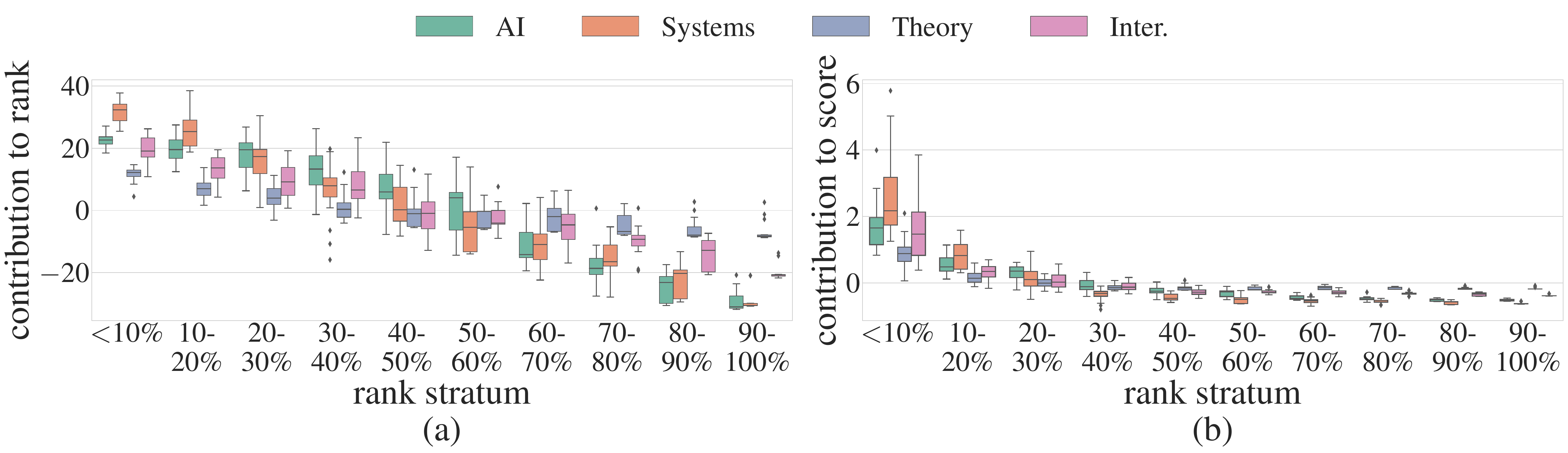}
\caption{Feature contributions to rank and score for the CSRankings dataset, aggregated over 10\% strata. In this ranking, 189 computer science departments are ranked based on a normalized publication count of the faculty across 4 research areas: AI (green), Systems (orange), Theory (purple), and Interdisciplinary (pink).  (a) Systems is the most important feature for an item's rank in the top-20\%, followed by AI. AI becomes more important for the rest of the ranking strata. ~(b) Feature contributions to score are less informative than to rank: both capture the same relative feature importance for the top 20\%; however, feature contributions become small and very similar as more items are tied for their score. (See rank vs. score plot on the top-right.) 
}
\label{fig:CSRanking}
\end{figure*}

\paragraph{Rank QoI} The Shapley value payoff function for the rank QoI is:

\begin{equation}
QoI_{f,\textbf{v},\mathcal{D}}(S) = \mathop{\mathbb{E}}_{\textbf{U}_{\mathcal{S}}} [r^{-1}_{\mathcal{D}',f}(\textbf{v}_{\mathcal{A} \setminus \mathcal{S}}\textbf{U}_{\mathcal{S}})]
\label{eq:qoi-rank}
\end{equation}

where $\mathcal{D}'$ is $\mathcal{D} \cup \{\textbf{v}_{\mathcal{A} \setminus \mathcal{S}}\textbf{U}_{\mathcal{S}}\} \setminus \textbf{v}$.
This QoI evaluates the impact of an item's features on its rank.
To get the contribution of a set of features $\mathcal{A} \setminus \mathcal{S}$, we take the expected value of the rank over a random variable $\textbf{U}_{\mathcal{S}}$ that draws values from the marginal distributions of the set of features in $\mathcal{S}$.

\paragraph{Top-$k$ QoI} The Shapley value payoff function to quantify the impact of an item's features on its presence or absence among the top-$k$ is stated similarly as rank QoI:

\begin{equation}
QoI_{f,\textbf{v}, \mathcal{D}}(S) = \mathop{\mathbb{E}}_{\textbf{U}_{\mathcal{S}}} [\mathds{1}_{r_{\mathcal{D}',f}(1\ldots k)}(\textbf{v}_{\mathcal{A} \setminus \mathcal{S}}\textbf{U}_{\mathcal{S}})]
\label{eq:qoi-top_k}
\end{equation}

where $\mathcal{D}'$ is $\mathcal{D} \cup \{\textbf{v}_{\mathcal{A} \setminus \mathcal{S}}\textbf{U}_{\mathcal{S}}\} \setminus \textbf{v}$.  The difference with rank QoI (Equation~\ref{eq:qoi-rank}) is that here we compute the expectation over the indicator function that returns 1 if an items' rank is at most $k$ and 0 otherwise. This QoI allows us to quantify how each feature contributed to getting the item into the top-$k$.

\paragraph{Shapley values for ranking} To compute Shapley values for the QoIs we defined, we need to apply Equation~\ref{eq:shapley} on the QoIs. Following the QII notation, we define the iota function $\iota$ as the difference between the QoI including feature $i$ and excluding it.

\begin{equation}
\iota_{f,\mathbf{v},\mathcal{D}}(i,\mathcal{S}) = \alpha (QoI_{f,\mathbf{v},\mathcal{D}}(\mathcal{S}\cup i) - QoI_{f,\mathbf{v},\mathcal{D}}(\mathcal{S}))
\label{eq:iota}
\end{equation}

Here, the QoI can be any defined earlier in this section, and $\alpha \in \{-1,1\}$ is a multiplier that adjusts the order of QoI terms. In this work, we consider QoIs beyond the score. For some, like rank, where smaller values are preferable, we set $\alpha = -1$ to adjust the $\iota$ function accordingly. 

Using this notation, we can define Shapley values for ShaRP:

\begin{equation}
\phi_i(f,\mathbf{v},\mathcal{D}) = \sum_{\mathcal{S}} \frac{|\mathcal{S}|! (n-|\mathcal{S}|-1)!}{n!}\iota_{f,\mathbf{v},\mathcal{D}}(i,\mathcal{S})
\label{eq:sharp-shapley}
\end{equation}

\paragraph{Case Study: QoIs for CSRankings}

We review local feature-based explanations generated by~\sys for CS Rankings, a real dataset ranking 189 U.S. Computer Science departments based on normalized faculty publication counts in four areas: AI, Systems, Theory, and Interdisciplinary~\cite{CSRankings}. See Appendix~\ref{sec:app:data} for dataset and ranker details. Our goal is to illustrate how \sys reveals meaningful insights about the data—and how those insights vary depending on the outcome being explained.

Figure~\ref{fig:CSRanking} shows feature contributions to the rank and score QoIs for CS Rankings, aggregated by 10\% rank strata. As shown in Figure~\ref{fig:CSRanking}a, Systems is the most important feature across all strata, followed by AI. Both contribute most positively in the top strata and most negatively in the bottom. \textit{Score-based explanations are less informative}: while they capture similar relative importance in the top 20\%, feature contributions flatten in lower strata, where many departments have near-tied scores, making comparisons difficult.

Figure~\ref{fig:csrankings/topk} presents aggregated feature contributions to the top-$k$ QoI, stratified by deciles. Systems again dominates in placing departments in the top-10, followed by AI. This trend is consistent with Figure~\ref{fig:CSRanking}b (score QoI), but more pronounced. Unlike the score QoI, the top-$k$ QoI also highlights Theory as impactful for top-$k$ inclusion. Notably, only the rank and top-$k$ QoIs capture a shift in relative importance between Systems and AI across strata.

Figure~\ref{fig:CSR_waterfalls}, previewed in the Introduction, shows a local explanation for Texas A\&M, ranked 34th with a score of 3.941. Waterfall plots in Figures~\ref{fig:csrankings/UT/score} (score QoI) and~\ref{fig:csrankings/UT/rank} (rank QoI) break down feature contributions relative to the mean outcome $\mathbb{E}[f(X)]$. In Figure~\ref{fig:csrankings/UT/score}, Interdisciplinary is the top contributor to Texas A\&M's score, followed by AI and Systems; Theory contributes negatively. For rank QoI, all features contribute positively, with Systems as the most impactful. This illustrates that different QoIs support different goals. To improve the score, Texas A\&M should focus on Interdisciplinary and AI. To improve rank, prioritizing Systems is more effective. The difference arises because \textit{increases in score do not always translate to changes in rank}—a score must exceed that of the next-highest item to affect position.

Another key aspect of these plots is the color of each feature, which indicates whether a feature contributes positively or negatively to the outcome. This is determined by the average feature value. Since the average score is influenced by outliers, while rank is not, the interpretation of contributions varies depending on the QoI.  For example, in CS Rankings, over 70\% of departments have scores below the mean. As a result, when using the score QoI, many or all of their features appear to contribute negatively. This highlights that the meaning of positive and negative contributions is dependent on the chosen QoI. See Figure~\ref{img:CSR-rankvsscore} and Appendix~\ref{sec:app:rank-vs-score} for the score vs. rank distribution for this dataset, and a more detailed comparison between the score-QoI-based and the rank-QoI-based explanations for CS Rankings.

\newcommand{\CSdataquerytab}{
    \small 
        \begin{tabular}{|c||c|c|c|c||c|}
		\hline
		\rowcolor[HTML]{C0C0C0} 
		Institution  & AI & Systems & Theory & Inter. & Rank\\ \hline
		\val{Georgia Tech}  & 28.5 & 7.8 & 6.9  & 10.2 & 5\\ \hline
		\val{Stanford}  & 36.7 & 5.4 & 13.3  & 11.5 & 6\\ \hline
		\val{UMich}  & 30.4 & 9.0 & 9.3  & 5.9 & 7\\ \hline
	\end{tabular}
}

\begin{figure}[t!]
    \centering
    \setlength{\tabcolsep}{0.3em}
        \subfloat[Feature contribution to the top-$k$ QoI, for $k=10\%$.  Systems is the most important feature, followed by Interdisciplinary and AI.]
	{\includegraphics[width=0.9\columnwidth]{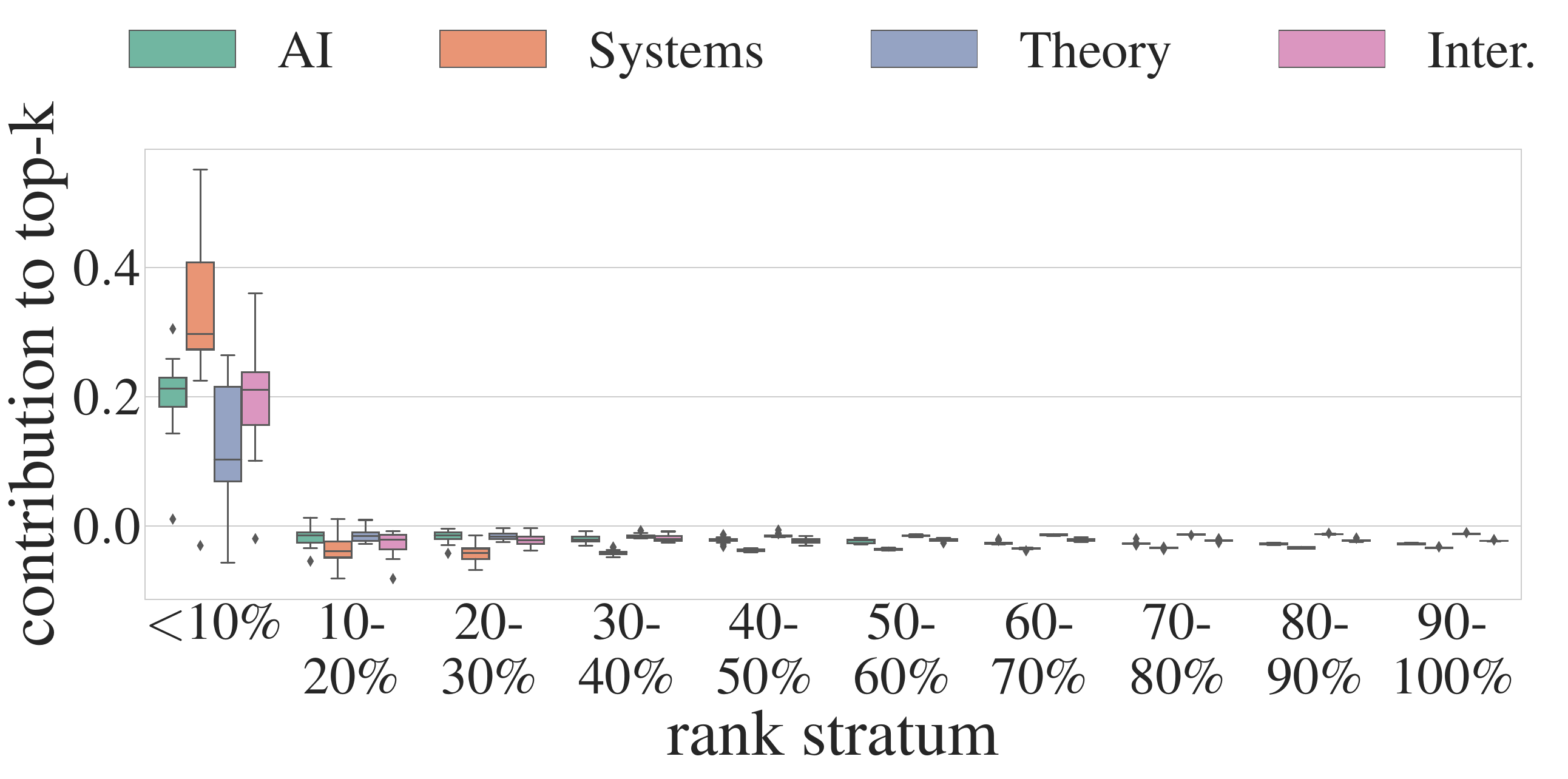}
	 \label{fig:csrankings/topk}}
	\hfill
	\subfloat[Feature values and rank of three highly ranked departments: Georgia Tech, Stanford, and UMich.]
	{\CSdataquerytab
	 \label{fig:CSRankings_data}}
	\hfill
	\subfloat[Pairwise QoI: Georgia Tech ranks higher than Stanford because of its relative strength in Systems.]{  
		\includegraphics[width=0.45\columnwidth]{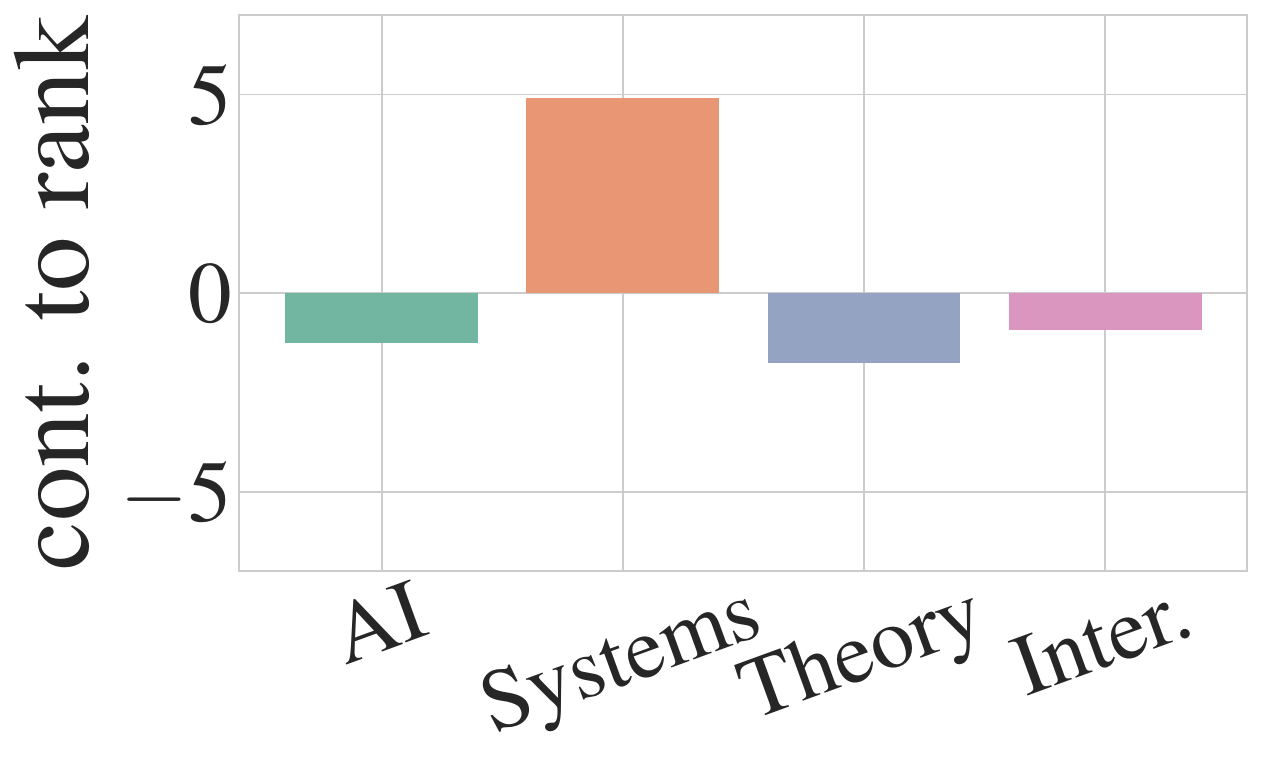}
		\label{fig:csrankings/pairwise}}
	\hfill
	\subfloat[Pairwise QoI: Stanford ranks higher than UMich despite Stanford's relative weakness in Systems.]{
		\includegraphics[width=0.45\columnwidth]{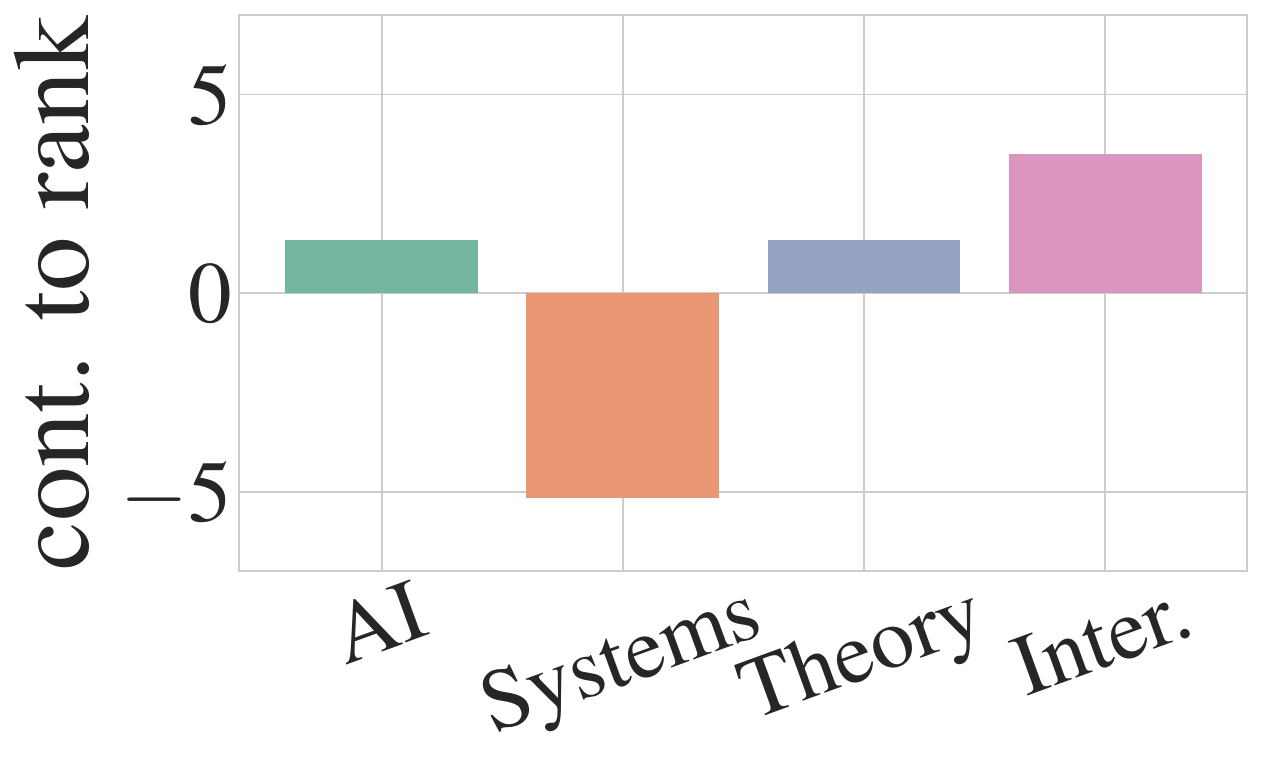}
		\label{fig:csrankings/pairwise2}
    }
    \caption{\small Feature importance for the top-$k$ QoI (\ie selection) for CS Rankings in~\ref{fig:csrankings/topk}, with further analysis of the relative orders among two pairs of departments in~\ref{fig:csrankings/pairwise} and~\ref{fig:csrankings/pairwise2}.}
    \label{fig:CSR_intro}
\end{figure}

\section{Pairwise Explanations}
\label{sec:pairwise}
We developed a method for computing feature importance for the relative order between a pair of items $\textbf{u}$ and $\textbf{v}$, to answer the question of why $\textbf{v}$ is ranked higher than $\textbf{u}$ (\ie  $\textbf{v}\succ\textbf{u}$). Our method is based on baseline Shapley value methods.

In Eq.~\ref{eq:shapley} we provided the definition of game-theoretic Shapley values. This equation uses a profit function defined over subsets $\mathcal{S}$ of the players. In the ML context, we use methods that take as input all features (players) - not a subset. Different Shapley value methods in ML take different approaches for addressing this problem, often referred to as the ``feature removal approach'' in the literature~\cite{chen2023algorithms}. One feature removal method is creating hybrid samples using the marginal distributions of the missing features and drawing values jointly. In Sec.~\ref{sec:prelim} we took this approach. We defined the items that we will be using in the Shapley value computations when using this marginal approach as $\textbf{v}_{\mathcal{A} \setminus \mathcal{S}}\textbf{U}_{\mathcal{S}}$ where \textbf{U} $= \langle \textbf{u}_1, \dots, \textbf{u}_m \rangle$ is a vector of $m$ items sampled from  $\mathcal{D}$.

For pairwise preferences, we will be using a different feature removal technique that uses a  ``baseline'' item to create hybrid items instead of the feature distributions. Baseline feature removal techniques select one item as the baseline item and then compare all other items to it. The benefit of these methods is that the exact feature contributions can be computed without any sampling. The disadvantage is that often it is hard to select the baseline sample because different baseline samples create different feature attributions and, in most contexts, it is hard to identify a ``neutral'' or ``average'' item. As an example, in related work, we mentioned ~\cite{singh2018posthoc} that attempts to identify a good baseline input document for DeepSHAP in IR. As another example, the baseline implementation of SHAP~\cite{lundberg2017unified} uses the all-zeroes item as the baseline sample. While selecting a baseline sample is not simple in most cases, we find that the baseline feature removal technique is a natural fit when we are explaining the difference in outcomes between two items $\textbf{v}$ and $\textbf{u}$.

When explaining the pairwise outcome of two items $\textbf{v}$ and $\textbf{u}$, we are going to generate an explanation for one item using the other as the baseline. In other words, for coalition $\mathcal{S}$, we will be creating the hybrid sample $\textbf{v}_{\mathcal{A} \setminus \mathcal{S}}\textbf{u}_{\mathcal{S}}$. Note that we do not need the feature distributions or any other parameters for this method. Additionally, note that we are not selecting a fixed item as the baseline, but we dynamically change it depending on the pair of items we want to compare.
This definition has a natural interpretation, the feature importance of a pairwise explanation amounts to the difference between the outcome of the two items. According to the property of efficiency (see Section~\ref{sec:prelim}) we have: $\sum_{i\in \mathcal{A}}\varphi_i(f,\textbf{v}) = f(\textbf{v}) - \mathbb{E}_{\textbf{X}}[f(\textbf{X})]=f(\textbf{v}) - \frac{1}{2}(f(\textbf{v}) + f(\textbf{u}))=\frac{1}{2}(f(\textbf{v}) - f(\textbf{u}))$.

The Shapley value of $\textbf{v}$ in comparison to $\textbf{u}$ is defined as:

\begin{equation}
\phi_i(f,\mathbf{v} \succ \textbf{u}) = \sum_{\mathcal{S}} \frac{|\mathcal{S}|! (n-|\mathcal{S}|-1)!}{n!}\iota_{f,\mathbf{v},\textbf{u}}(i,\mathcal{S})
\label{eq:sharp-shapley-pairwise}
\end{equation}

Note that Eq.~\ref{eq:sharp-shapley-pairwise} differs from Eq.~\ref{eq:sharp-shapley} in setting $\mathcal{D} = \{\textbf{u}\}$.  Note also that any QoI from Section~\ref{sec:qoi} can be used when calculating the pairwise explanation. Because pairwise preferences are of especial interest to ranking tasks, we will only be using rank as the QoI for the pairwise method in the rest of the paper.

\paragraph{Case Study: Explanations of Pairwise Outcomes in CS Rankings}
In Figure~\ref{fig:CSRankings_data}-~\ref{fig:csrankings/pairwise2} we continue our analysis of the top-$k$ and consider the relative ranking of three universities: Georgia Tech in rank 5, Stanford in rank 6, and UMich in rank 7. We wish to understand why Georgia Tech is ranked higher than Stanford (Figure~\ref{fig:csrankings/pairwise}), and why Stanford is ranked higher than UMich (Figure~\ref{fig:csrankings/pairwise2}). In both cases, Georgia Tech and UMich have lower values for all features except Systems. The Systems value of Georgia Tech is high enough to overcome the contributions of other features and rank it higher than Stanford. However, for UMich, we see that, while Systems is the most important feature in the top-10\% stratum, it is not important enough to move UMich above Stanford.

Pairwise Shapley explanations can clarify rank differences between two items. In Fig.~\ref{fig:csrankings/pairwise}, we explain the pairwise outcome for Georgia Tech vs. Stanford.  For $\textbf{v}_{\text{Georgia Tech}} = (28.5, 7.8, 6.9, 10.2)$, we use $\textbf{u}_{\text{Stanford}} = (36.7, 5.4, 13.3, 11.5)$ as the baseline. For coalition $\mathcal{S}=\{\text{AI},\text{Systems}\}$, we construct $\textbf{v}_{\mathcal{A} \setminus \mathcal{S}}\textbf{u}_{\mathcal{S}}= (36.7, 5.4, 6.9, 10.2)$, enabling a direct feature comparison.  The pairwise explanation from~\sys is intuitive: in the same figure, Systems improves Georgia Tech's rank by 5 compared to Stanford. Feature contributions sum to half the rank difference between these universities, aligning with Fig.~\ref{fig:csrankings/topk}, which highlights Systems as particularly influential for top-$k$ universities.

\section{Empirical Evaluation}
\label{sec:metrics}

Multiple metrics for evaluating explanation methods across key dimensions have been proposed~\cite{miller2019explanation, molnar2020interpretable}, including for ranking~\cite{singh2018posthoc,LIRME,bhatt_2021_evaluating,chowdhuryrankshap}. In this work, we use such metrics to compare explanation methods and adapt or define several others for evaluating feature importance in ranking. We aim to formulate these metrics as generally as possible to support broader applicability.

Our focus is on explanation methods that return a numerical vector of feature attributions explaining the outcome for a given item.  We can assess \emph{pair-wise explanation agreement} by comparing the feature vectors of a pair of explanations.  Furthermore, we can use an explanation to compute the outcome for the item being explained (e.g., its rank), and compare it to the actual observed outcome for that item.  This allows us to assess \emph{fidelity of an explanation}.  Below, we describe explanation agreement and fidelity metrics and also explain how these primitives can be aggregated to assess \emph{sensitivity} and \emph{fidelity} of an explanation method, and to quantify inter-method \emph{explanation agreement}.

\paragraph{Notation} In Section \ref{sec:prelim}, we have been using $\phi(\textbf{v})$ to represent the vector of feature weights, computed using Shapley values. We generalize our definition here to $g(\textbf{v})$ to represent the output of any feature-based explanation method $g$, regardless of whether it consists of Shapley values or of some other numerical quantification of feature importance. For all methods we consider, $g(\textbf{v})$ is a vector of numerical contributions of each feature towards the outcome for item $\textbf{v}$.

\subsection{Fidelity Metrics}
\label{sec:metrics:fidelity}

\paragraph{Explanation Fidelity}

A useful property of feature-based explanations is that the actual outcome can be computed from them. For Shapley-value-based explanations, this follows from the efficiency property of Shapley values, see Section~\ref{sec:prelim}. 
Fidelity measures how well the explanation $g(\textbf{v})$ matches the model prediction $f(\textbf{v})$ being explained, see~\cite{Chowdhury_ranklime,molnar2020interpretable}.  SHAP and LIME explanations can be used to compute an item's score (score QoI in our terminology)~\cite{lundberg2017unified,Chowdhury_ranklime}, with feature importance indicating the displacement due to that feature from the mean score, either positively or negatively. \sys explanations can be used to compute the outcome for all supported QoIs, including score, rank, and top-$k$, and for the pairwise method.  

For QoIs that concern a single item, namely, score, rank, and top-$k$, we compute fidelity of explanation $g$ for item \textbf{v} as:

\begin{equation}
F(g,\textbf{v}, \mathsf{QoI}()) = 1 - \frac{1}{Z} \left|{\mathsf{QoI}(\textbf{v}) - \sum_{i=1}^d g(i,\textbf{v})}\right|
\label{eq:fidelity}
\end{equation}

Here, $\mathsf{QoI}(\textbf{v})$ returns the value of the quantity of interest (i.e., the outcome being explained by $g$), such as \textbf{v}'s score, rank, or presence in the top-$k$, while $g(i,\textbf{v})$ is the contribution of the i-th feature of $v$.  Finally, $Z$ is the normalizer set to the maximum distance between a pair of outcomes for the given dataset $\mathcal{D}$ and ranker $f$ (omitted to simplify notation), and for the specified QoI.   Note that, for pairwise explanations, fidelity $F(g,\textbf{u} \succ \textbf{v}) = 1$ if $u$ is ranked higher than $u$ and if $g$ predicts that relative order among the items, and is $0$ otherwise.

\begin{example}
Consider, for example, the explanation of Texas A\&M University's $rank = 34$ in CS Rankings, presented as a waterfall plot in Figure~\ref{fig:csrankings/UT/rank}.
The sum of feature weights $- 20.78 - 19.52 - 18.71 - 1.95 = - 60.96$ captures the displacement of Texas A\&M University in the ranking relative to the middle of the ranked list (position 94.5 out of 189), up to rounding: $94.5 - 60.96 = 33.54$.  This explanation has near-perfect fidelity $1 - \frac{0.16}{189} = 0.998$. We use the length of the ranked list $Z=189$ as the normalizer for rank QoI.
\end{example}

\paragraph{Method Fidelity}
We aggregate per-item fidelity (per Equation~\ref{eq:fidelity}) to quantify the fidelity of an explanation method as:
\begin{equation}
F(g, \mathcal{D}) = \mathbb{E}_{\textbf{v} \in \mathcal{D}} F(g,\textbf{v})
\label{eq:fidelity:agg}
\end{equation}

For pairwise, we compute $F(g, \mathcal{D})$ as the expectation of $F(g,\textbf{u} \succ \textbf{v})$ over all pairs of distinct items $\textbf{u}, \textbf{v} \in \mathcal{D}$.

\subsection{Agreement Metrics}
\label{sec:metrics:agreement}

\paragraph{Explanation Agreement}
When comparing explanation methods, we may be interested in knowing how similar their explanations are for \emph{the same item}. Alternatively, when analyzing an explanation method, we may want to know how similar its explanations are for \emph{some pair of items} (\eg those that are similar in feature space, or that have similar outcomes, or both).  

We define explanation agreement, based on three distance metrics often used for comparing rankings~\cite{singh2018posthoc,chowdhuryrankshap}, (1) Kendall's tau distance, (2) Jaccard distance of the top-$2$ features, and (3) Euclidean distance between the explanation vectors. For each of these distance metrics, we normalize them to the $[0, 1]$  range and then transform their output so that 1 means full agreement (similarity) and 0 means full disagreement.
For dataset $\mathcal{D}$ and ranker $f$, we define explanation agreement as:

\begin{equation}
A(g, q, \textbf{u}, \textbf{v}, \mathsf{sim}() ) = \mathsf{sim}(g(\textbf{u}), q(\textbf{v}))
\label{eq:agreement}
\end{equation}

Here, $g$ and $q$ are explanation methods, \textbf{u} and  \textbf{v} are points being explained, and $\mathsf{sim}()$ is a function that computes the specified similarity metric over the explanations. Two important cases are: when $g=q$ and $\textbf{u} \neq \textbf{v}$, we are comparing explanations generated by the same method for different points.  Conversely, when $g \neq q$ and $\textbf{u} = \textbf{v}$, we are comparing explanations of the same point generated by different methods.

\begin{example}
For example, consider the explanations of Texas A\&M's score and rank, produced by \sys for score QoI~\ref{fig:csrankings/UT/score} and rank QoI~\ref{fig:csrankings/UT/rank}, respectively. These explanations are similar in the sense that they explain two related outcomes (score and rank) of the same item.  However, they are dissimilar in that the relative importance of Texas A\&M's features is different.  For rank QoI, the explanation ranks features as $\langle \text{Systems}, \text{AI}, \text{Inter}, \text{Theory} \rangle$. However, for score QoI, the explanation ranks features differently as $\langle \text{Inter}, \text{AI},\text{Systems}, \text{Theory} \rangle$. These lists are dissimilar in terms of the relative order of the features, with 3 out of 6 possible pairs appearing in the opposite relative order.  An explanation agreement metric that uses Kendall's tau distance as a sub-routine allows us to quantify this.
\end{example}

\paragraph{Method Agreement}
To compute agreement for a pair of explanation methods $g$ and $q$, for a dataset $\mathcal{D}$, we compute explanations for each item using each method, compute pair-wise explanation agreement per Eq.~\ref{eq:agreement}, and aggregate it across $\mathcal{D}$.

\begin{equation}
A(g, q, \mathcal{D}, \mathsf{sim}() ) = \mathbb{E}_{\textbf{v}\in \mathcal{D}} A(g, q, \textbf{v}, \textbf{v}, \mathsf{sim}() ) 
\label{eq:agreement:agg}
\end{equation}

\paragraph{Method Sensitivity}
The Sensitivity of an explanation method quantifies the similarity between explanations of similar items~\cite{bhatt_2021_evaluating}.  We will use $\mathsf{nbr}(\mathbf{v})$ (as in ``neighbor'') to refer to a function that retrieves items that are in some sense similar to $\mathbf{v}$, noting that this similarity may be based on items' features, their outcomes for some QoI, or both.  For each  \textbf{v}, we retrieve its neighbors $\mathsf{nbr}(\textbf{v})$, compute pair-wise explanation agreement between \textbf{v} and each of its neighbors per Eq.~\ref{eq:agreement}, and aggregate this value over $\mathcal{D}$:

\begin{equation}
S(g, \mathcal{D}, \mathsf{sim}, \mathsf{nbr}() ) = \mathbb{E}_{\textbf{v}\in \mathcal{D}, \textbf{u} \in \mathsf{nbr(\textbf{v}}) } A(g, g, \textbf{v}, \textbf{u}, \mathsf{sim}() ) 
\label{eq:agreement:agg-m}
\end{equation}
\section{The ~\sys library}
\label{sec:sys}
~\sys is implemented in Python,  follows an API structure similar to scikit-learn~\cite{sklearn_api}, and is parallelized.  The library can be used both to compute exact feature importance values and to approximate them to improve running times.

\paragraph{Implementation of QoIs for ranking.}  
We provide Algorithm~\ref{alg:sharp} to showcase the flexibility of \sys. Using this implementation, we can 1) easily switch between QoIs, 2) calculate both marginal and baseline Shapley values, and 3) approximate Shapley values for efficiency. The algorithm relies on black-box access to the model that generates the outcome (\ie specifying an input and observing the outcome used in the QoI). Specifically, Algorithm~\ref{alg:sharp} takes as input a dataset $\mathcal{D}$, a reference set $\mathcal{D}' \subseteq \mathcal{D}$ from which samples are drawn, an item $\textbf{v}$ for which the explanation is generated, the number of samples $m$, the maximum coalition size $c$, and the $\iota()$ function (Equation~\ref{eq:iota}) used to quantify feature importance.

To change the QoI, we modify the input $\iota()$ function. To switch to the pairwise baseline method, we set $\mathcal{D}' = \textbf{u}$ and $m=1$, where $\textbf{u}$ is the baseline item to compare against $\textbf{v}$. To approximate feature importance, we control the parameters $m$ and $c$. Passing in the full set of items as the reference set ($\mathcal{D}' = \mathcal{D}$), and setting $m = |\mathcal{D}| - 1$ and $c = |\mathcal{A}| - 1$, yields exact Shapley value computation—\ie each feature of $\textbf{v}$ is quantified against all other items in $\mathcal{D}$ using all possible coalitions of features except the one being evaluated.

Because we compute the rank of each item relative to the entire dataset $\mathcal{D}$, the dataset must be provided along with the reference set. We provide an empirical analysis of the impact of $m$ and $c$ on performance in Section~\ref{sec:exp:efficiency:approx}. 

We now describe the algorithm for marginal exact computation, which generalizes all cases discussed above. By definition, Shapley values compute feature importance using all possible coalitions of features and all items in the dataset—referred to here as the \emph{exact computation} of local feature-based explanations. For illustrative purposes, we explicitly include the construction of the random variable $\textbf{U}$ in lines 4–7 of Algorithm~\ref{alg:sharp}. 
For each feature $i \in \mathcal{A}$, the algorithm considers all coalitions $\mathcal{S} \subseteq \mathcal{A} \setminus \{i\}$. For each $\mathcal{S}$, it draws $m = |\mathcal{D}| - 1$ samples from $\mathcal{D}$. Two vectors of items are then constructed: $\textbf{U}_1$, where features in $\mathcal{S}$ vary as in $\textbf{U}$ and the rest are fixed to their values in $\textbf{v}$; and $\textbf{U}_2$, where features in $\mathcal{S} \cup \{i\}$ vary as in $\textbf{U}$, with the remaining features again fixed to $\textbf{v}$. 
The importance of coalition $\mathcal{S}$ for feature $i$, denoted $\phi_{i_\mathcal{S}}(\textbf{v})$, is computed using the QoI function $\iota()$, which measures the difference between $\textbf{U}_1$ and $\textbf{U}_2$. This quantity is then weighted by the number of coalitions of size $|\mathcal{S}|$—specifically, ${d-1 \choose |\mathcal{S}|}$—and accumulated into the final contribution $\phi_i(\textbf{v})$, normalized over all possible coalition sizes $d$.

In practice, one of the main bottlenecks in computing feature contributions, especially with complex black-box models, is inference time. To mitigate this, we cache inference results in a hash map, allowing repeated inputs to return cached outputs in constant time ($O(1)$). This significantly speeds up computation as more tuples are processed. Initially, the explainer experiences a ``cold start'' with no cached results, but performance improves to a ``warm start'' as the cache builds, reducing the need for repeated model inference.

\begin{algorithm}[tb]
\caption{Local feature importance using~\sys}
\label{alg:sharp}
\begin{algorithmic}[1]
\REQUIRE Dataset $\mathcal{D}'$, reference set $\mathcal{D}'$, item $\textbf{v}$, number of samples $m$, maximum coalition size c, $\iota()$
\STATE $\phi(\textbf{v}) = \langle 0, \ldots, 0 \rangle$
\FOR{$i \in \mathcal{A}$} 
\FOR{ $\mathcal{S} \subseteq \mathcal{A}\setminus \{ i \}$ and $|\mathcal{S}| \leq c$}
\STATE $\textbf{U} \sim \mathcal{D}' \setminus $\textbf{v}$, m$
\STATE $\textbf{U}_1 = \textbf{v}_{\mathcal{A}\setminus \mathcal{S}}\textbf{U}_{\mathcal{S}}$
\STATE $\textbf{U}_2 = \textbf{v}_{\mathcal{A}\setminus \{ \mathcal{S} \cup i \}}\textbf{U}_{\mathcal{S} \cup i}$
\STATE $\phi_{i_\mathcal{S}}(\textbf{v}) = \iota(\textbf{U}_1,\textbf{U}_2)$
\STATE $\phi_i(\textbf{v})= \phi_i(\textbf{v}) + \frac{1}{d}\frac{1}{{d-1 \choose |S|}} \phi_{i_\mathcal{S}}(\textbf{v})$
\ENDFOR
\ENDFOR
\RETURN $\phi(\textbf{v})$, the Shapley values $\textbf{v}$'s features
\end{algorithmic}
\end{algorithm}

Evaluating the $\iota()$ function, is straightforward for the \textit{score QoI} but not for the ranking-specific QoIs. Specifically, for the score QoI, using the definition in Section~\ref{sec:qoi}, we take the mean of the (per-element) difference of $f(\textbf{U}_1)$ and $f(\textbf{U}_2)$. However, this is not the case for ranking-specific QoIs. The rank of an item is computed with respect to all other items in the sample.  
This adds two steps to calculating the rank QoI compared to the score QoI. The item we are explaining needs to be removed from $\mathcal{D'}$, and the score of each item $\textbf{u}_i \in \textbf{U}_1$ (and equivalently $\textbf{u}_j \in \textbf{U}_2$) needs to be compared to the scores of all items in $\mathcal{D'}$. The computation of $\iota_{Rank}$ is summarized in Algorithm~\ref{alg:iota_rank}.

\begin{algorithm}[tb]
\caption{$\iota_{Rank}$}
\label{alg:iota_rank}
\begin{algorithmic}[1]
\REQUIRE Dataset $\mathcal{D}'$, scoring function $f$, item \textbf{v}, $\textbf{U}_1$, $\textbf{U}_2$, number of samples $m$
\STATE $\phi=0$
\FOR {$i \in \{1, \dots, m\}$}
\STATE $\textbf{u}_{1} = \textbf{U}_1(i)$
\STATE $\textbf{u}_{2} = \textbf{U}_2(i)$
\STATE $\mathcal{D}_1 = \mathcal{D} \setminus \{ \textbf{v} \} \cup \{ \textbf{u}_1 \}$
\STATE $\mathcal{D}_2 = \mathcal{D} \setminus \{ \textbf{v} \} \cup \{ \textbf{u}_2 \}$
\STATE $\phi  = \phi  + r^{-1}_{\mathcal{D}_2,f}(\textbf{u}_2) - r^{-1}_{\mathcal{D}_1,f}(\textbf{u}_1)$
\ENDFOR
\RETURN $\phi / |\textbf{U}_1|$
\end{algorithmic}
\end{algorithm}

To compute feature importance that explains whether an item appears at the top-$k$, for some given $k$, we use a similar method as for rank QoI. The difference is that, rather than computing the difference in rank positions for a given pair of items $\textbf{u}_1$ and $\textbf{u}_2$, we instead check whether one, both, or neither of them is at the top-$k$. 
As in Algorithm~\ref{alg:iota_rank}, we work with  $\mathcal{D}_1 = \mathcal{D} \setminus \{ \textbf{v} \} \cup \{ \textbf{u}_1 \}$ and $\mathcal{D}_2 = \mathcal{D} \setminus \{ \textbf{v} \} \cup \{ \textbf{u}_2 \}$ for each sample.  We increase the contribution to $\phi$ by 1 if only $\textbf{u}_1$ is in the top-$k$, and decrease it by 1 if only $\textbf{u}_2$ is in the top-$k$. We omit pseudocode due to space constraints.

\paragraph{Visualizing feature importance.} We use three visualization methods. First, waterfall plots (Figure~\ref{fig:CSR_waterfalls}) show feature importance for a single item, following~\cite{lundberg2017unified}. Second, box-and-whisker plots (Figures~\ref{fig:CSRanking},~\ref{fig:csrankings/topk},~\ref{fig:moving-company},~\ref{fig:distr-dependence}) aggregate local importance across 10\%-width ranking strata, showing median and variance per feature. Third, bar charts (Figures~\ref{fig:csrankings/pairwise},~\ref{fig:csrankings/pairwise2}) display pairwise contributions from the perspective of the first item in each pair.

\section{Experimental Evaluation of~\sys}
\label{sec:exp}

We ran extensive experiments on real and synthetic datasets with score-based ranking tasks to demonstrate the utility and performance of \sys. Section~\ref{sec:exp:efficiency} presents efficiency results, Section~\ref{sec:exp:scored} provides a qualitative evaluation, and Section~\ref{sec:exp:other} compares \sys to other methods using the metrics from Section~\ref{sec:metrics}. All experiments were run on a 14-core Intel Xeon Platinum 8268 (2.90GHz) machine with 128GB RAM.
We evaluate the performance of~\sys and compare it to other local feature importance methods, using several real and synthetic datasets, with the corresponding ranking tasks.  Dataset properties, along with ranker type (score-based or learned) are summarized in Table~\ref{table:datasets}, see Appendix~\ref{sec:app:data} for details.

\begin{table}[b!]
\caption{Datasets, sorted by \# tuples. S stands for score-based ranked task and LtR for learning-to-rank.}
\centering
\small 
\begin{tabular}{ l c r c c} 
 name & source & \# tuples & \# features & task \\ 
 \midrule
    Tennis (\texttt{ATP}) & \cite{ATP_Tennis} & 86  & 6 & S \\
    CS Rankings (\texttt{CSR}) & \cite{CSRankings} & 189 & 5  & S\\
    Times Higher Education (\texttt{THE}) & \cite{Times_Higher_Ed} & 1,397 & 5 & S  \\
    Synthetic (\texttt{SYN}) & here & 2,000 & 2 or 3 & S \\
    ACS Income - Alaska (\texttt{ACS-AK}) & \cite{ding_retiring_2022} & 3,546 & 10 & LtR \\
    Moving company (\texttt{MOV}) & \cite{DBLP:conf/forc/YangLS21} & 4,000 & 3 & LtR \\
    ACS Income - Texas (\texttt{ACS-TX}) & \cite{ding_retiring_2022} & 135,924 & 10 & LtR \\
 \bottomrule
\end{tabular}
\label{table:datasets}
\end{table}

\subsection{Qualitative Analysis}
\label{sec:exp:scored}
We already presented a detailed case study of CS Rankings presented as an example across the previous sections. To evaluate \sys across different settings, we conducted two additional experiments. First, we analyzed a set of simple synthetic datasets coming from multiple different distributions and studied how each distribution affects the ranking. Secondly, we compared the explanations resulting from two different LtR rankers for the Moving Company dataset.

\subsubsection{Score-based Ranking with Synthetic Data}
\label{sec:rank-synth-data}
In this set of experiments (see Appendix~\ref{sec:app:score-based} for details), we use simple two-feature datasets to study how feature distributions and scoring functions interact with ranking. We consider two settings: (1) fixed scoring function with varying distributions, and (2) fixed distributions with varying scoring functions.

When the scoring function is fixed, \textit{feature importance depends on both distribution and stratum}. Features with higher variance dominate at the top, while in the middle, either feature may prevail, increasing variability. For negatively correlated features, the pattern holds with opposite contribution signs. Discrete features (\eg Bernoulli) split the ranking into segments, with the second feature determining order within each. When distributions are fixed and scoring functions vary, \textit{importance varies by stratum}, depending on both weight and variance. A low-variance feature can dominate if its weight is high. Finally, we show that under certain distributions, \textit{low-ranked items can jump to the top-$k$}, contradicting the locality assumption in~\citet{Anahideh2022}. Even items in the top-50\% can move into the top-10\% with specific value changes.

\begin{figure}
    \centering
        \subfloat[XGB (original data)]{
        \includegraphics[width=0.48\columnwidth]{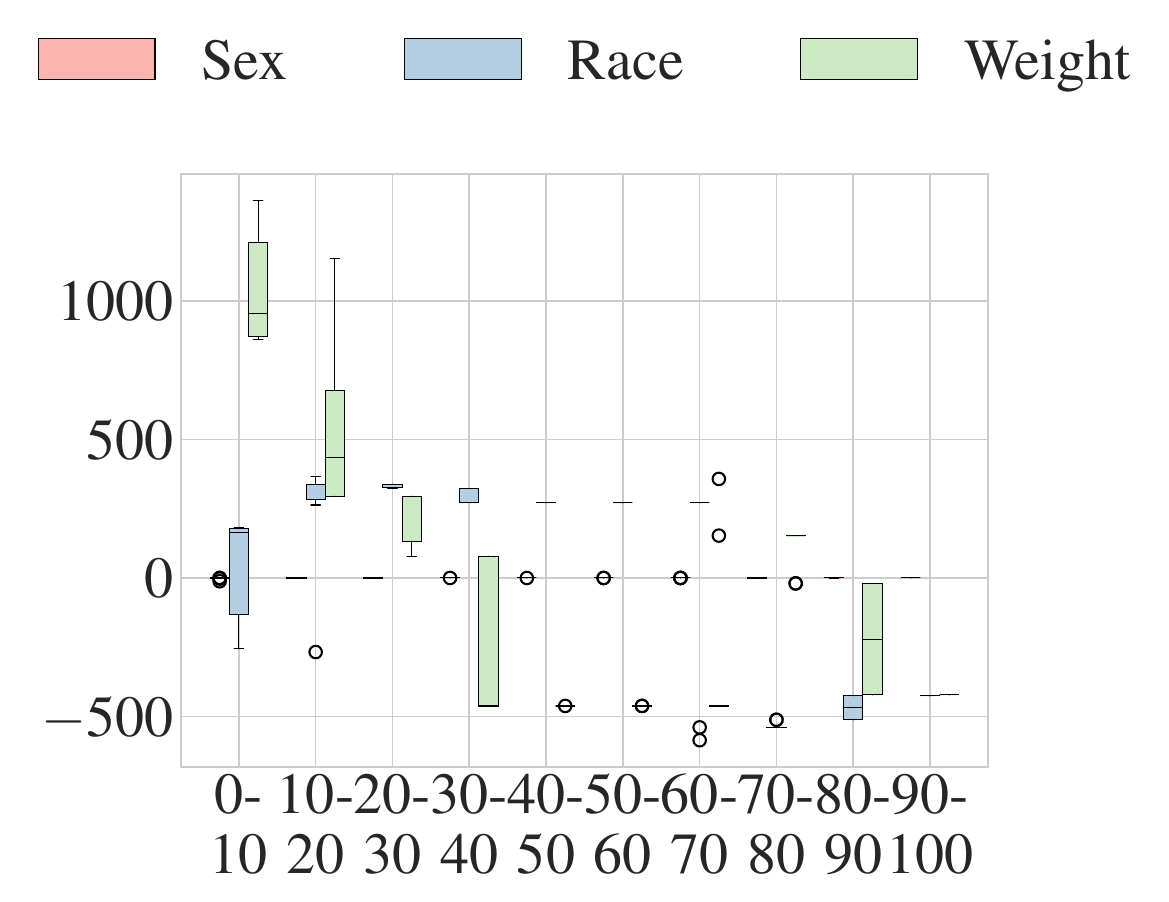}
            \label{img:xgb-biased}}
        \hfill
        \subfloat[XGB (fairness interv.)]{
        \includegraphics[width=0.48\columnwidth]{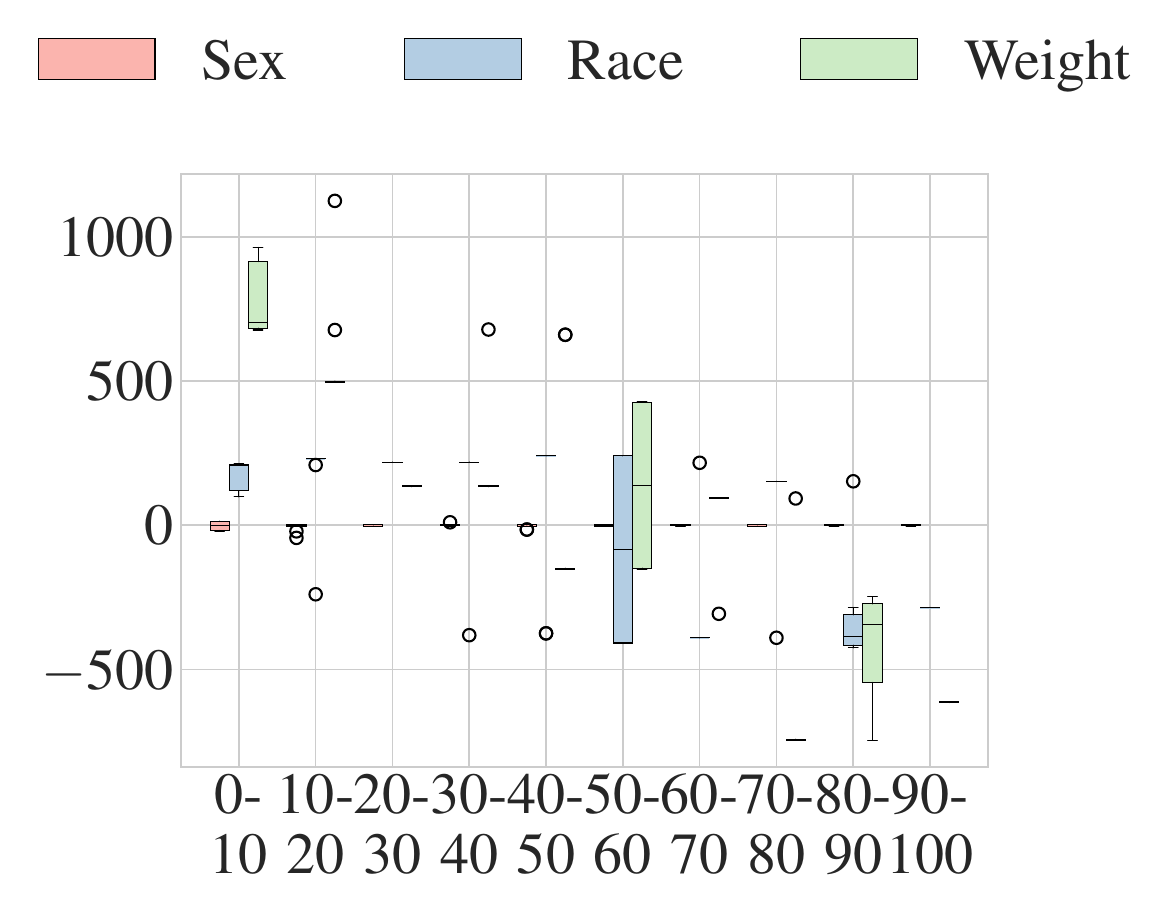}
            \label{img:xgb-fair}}
        \hfill
        \subfloat[LGB (original data)]{
        \includegraphics[width=0.48\columnwidth]{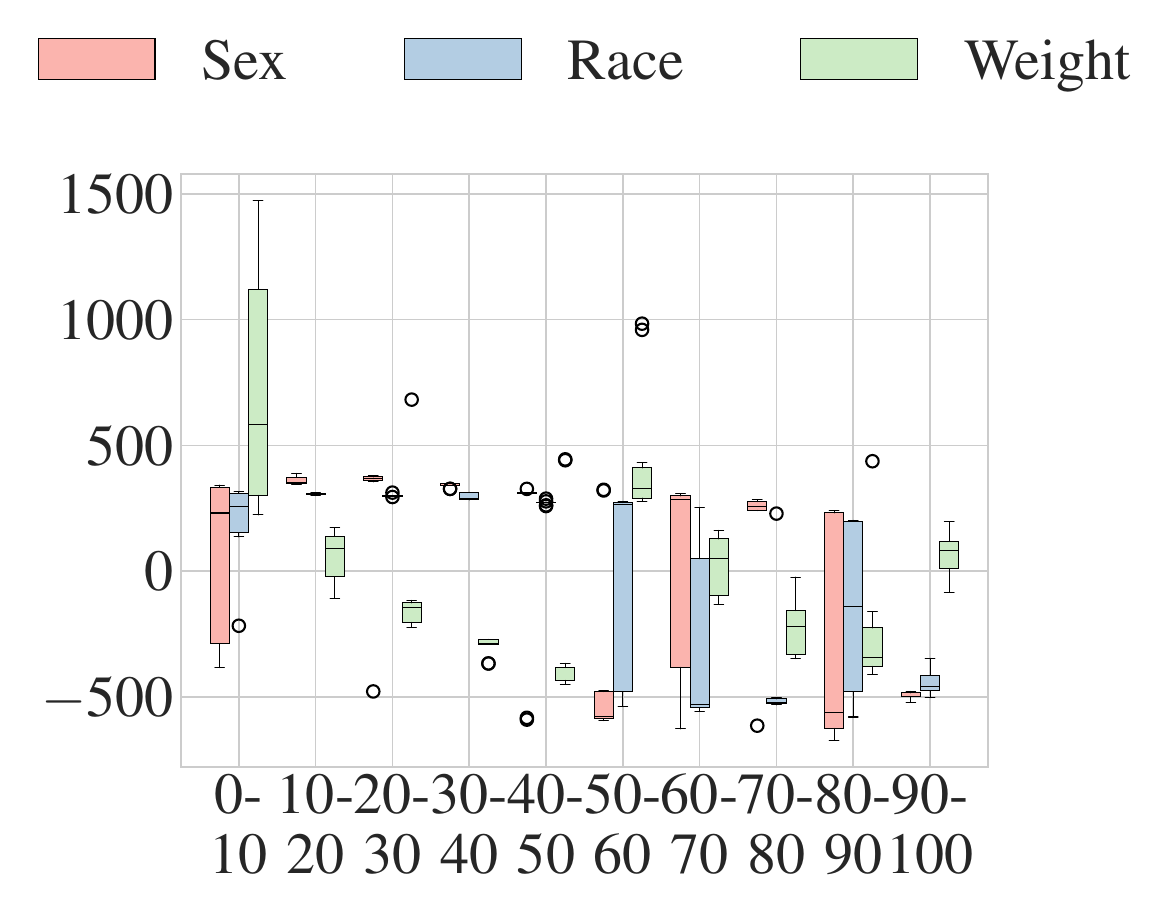}
            \label{img:lgb-biased}}
        \hfill
        \subfloat[LGB (fairness interv.)]{
        \includegraphics[width=0.48\columnwidth]{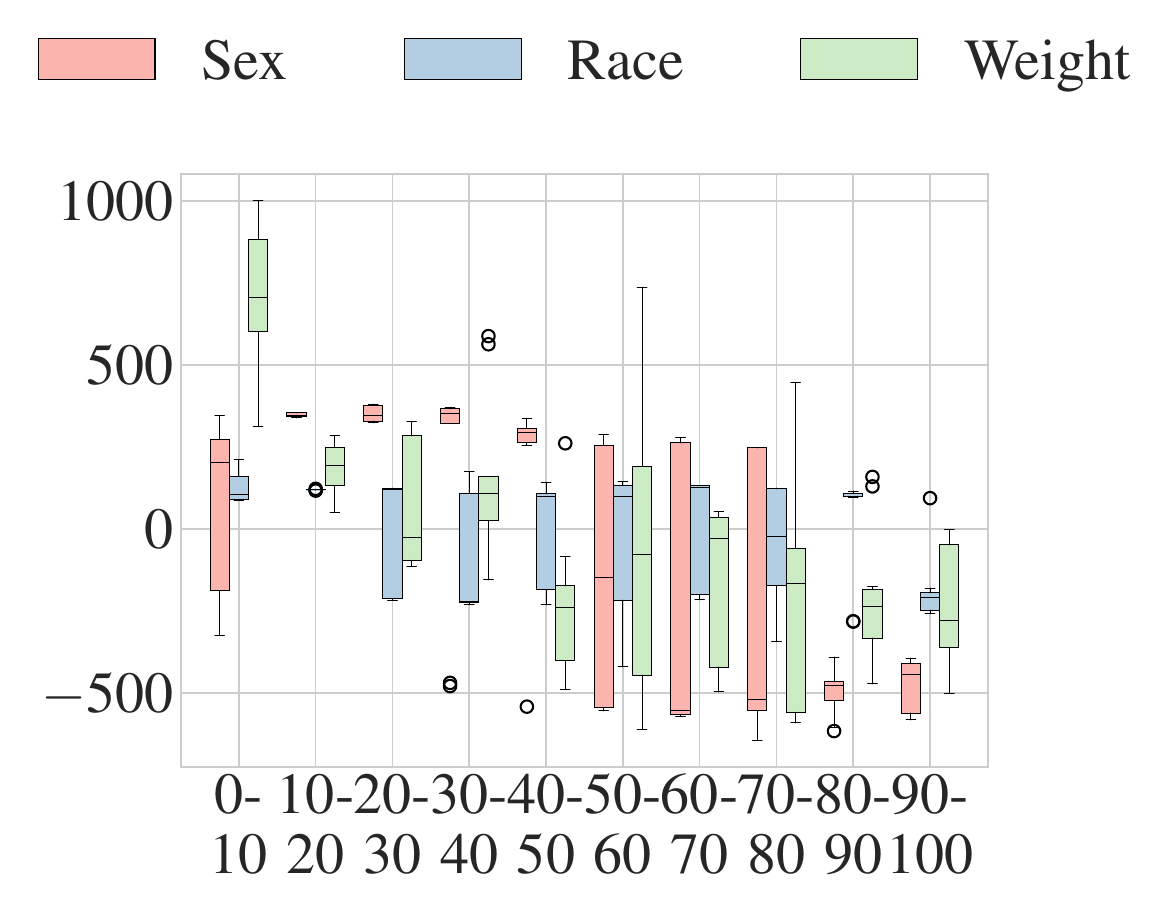}
            \label{img:lgb-fair}}
      \caption{Feature contribution to the rank QoI for (a) XGB over the original moving company dataset, (b) XGB over the unbiased version, (c) LGB over the original moving company dataset and (d) LGB over the unbiased version.}
    \label{fig:moving-company}
\end{figure}

\subsubsection{Learning to Rank}
\label{sec:ltr}
We now showcase how \sys can be used to audit black-box rankers and understanding their decision process. We use an XGB ranker with a pairwise ranking objective and an LGB ranker with a LambdaRank objective. Both are trained on training sets and evaluated on test sets of 2,000 tuples each.
We use ~\sys to explain 100 items (10 per stratum) of each test set, with no approximations and the rank QoI.
In Figure \ref{fig:moving-company}, we observe that the two LtR models behave significantly differently.

XGB rankers do not appear to rely on the Sex feature, regardless of whether the de-biasing intervention from \cite{DBLP:conf/forc/YangLS21} is applied. However, Race remains influential; in Figure~\ref{img:xgb-biased}, it boosts applicants' rankings by roughly 400 positions up to the 70th percentile. This is notable given that Weight Lifting contributes positively in the 70–80th percentile range but negatively in the 60–70th range. Ideally, its impact should be more monotonic, as partially achieved in Figure~\ref{img:xgb-fair}. Although Race shows slightly reduced influence after the intervention, it remains an important feature.

In contrast, LGB rankers tend to rely on all features. In the original model (Figure~\ref{img:lgb-biased}), Sex and Race are highly influential across all strata, often ranking as the top features for applicants in the lower percentiles (50th and below). Analysis of the 10–20th, 60–70th, and 90–100th percentiles shows that Weight Lifting has minimal impact on decisions, with Sex and Race largely determining rank. The fairness intervention reduces this effect somewhat (Figure~\ref{img:lgb-fair}) by increasing the influence of Weight Lifting, but Race and Sex remain dominant features, occasionally outweighing Weight Lifting.

In summary, results indicate that XGB relies more on Race, while LGB emphasizes Sex. Bias mitigation is effective up to the 10th percentile but fails to correct bias across the remaining strata.

\subsubsection{ACS Income}
\label{sec:acs-income}
\begin{figure*}
    \centering
        \subfloat[ACS strata]{
        \includegraphics[width=0.72\linewidth]{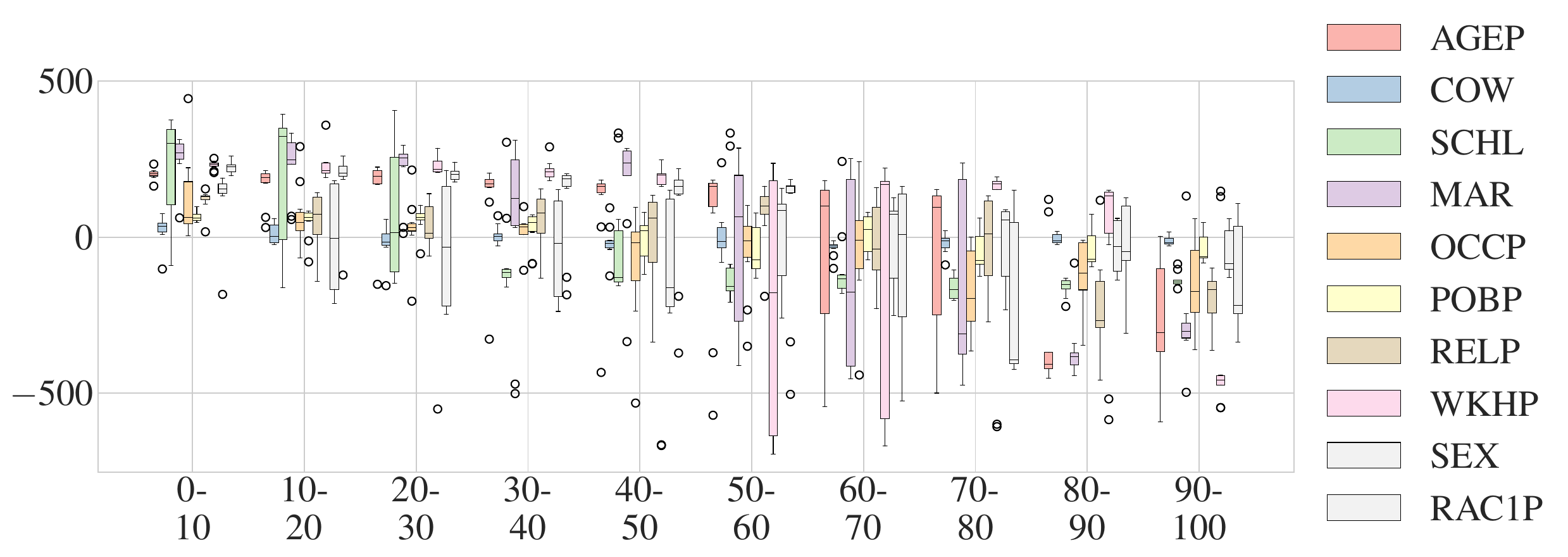}
            \label{img:acs-strata}}
        \hfill
        \subfloat[Overall]{
        \includegraphics[width=0.24\linewidth]{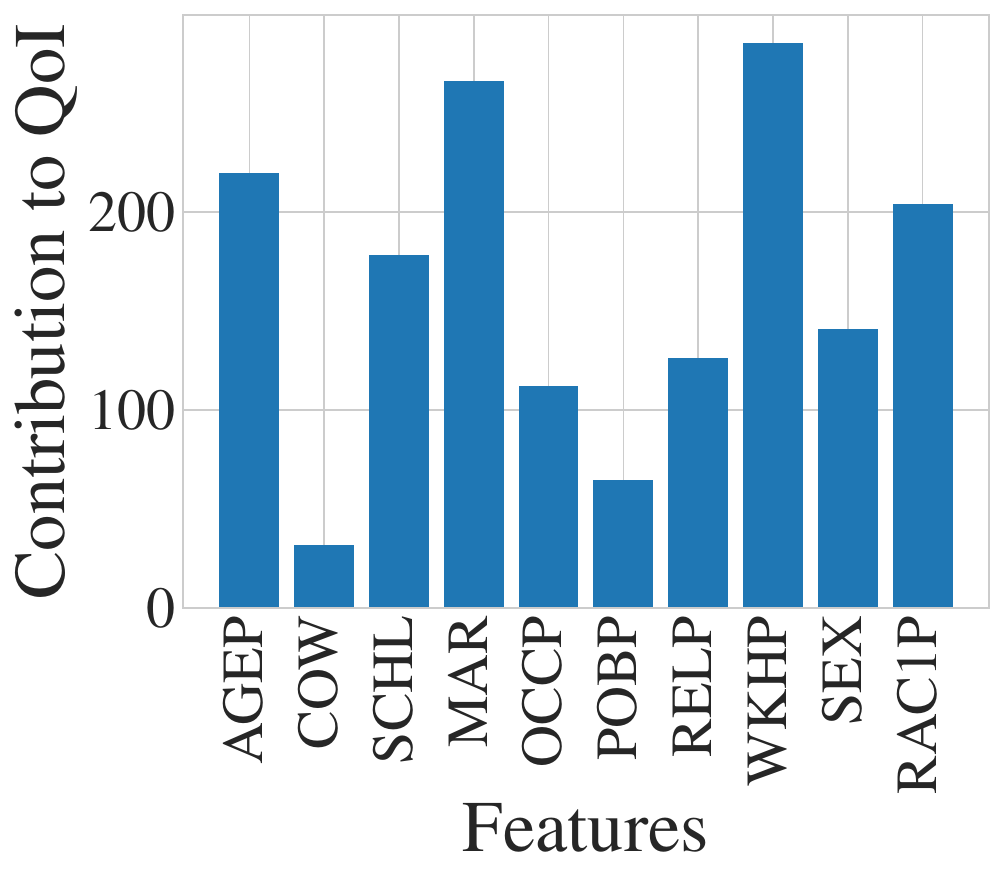}
            \label{img:acs-overall}}
      \caption{Feature contribution on  ACS Income (Alaska) to the rank QoI (a) across strata and (b) overall.}
    \label{fig:acs-overall-alaska}
\end{figure*}

We use the 2018 ACS Income dataset (10 features, 6 categorical) from Alaska (3,546 records) and Texas (135,924 records) as a secondary case study. The task is to predict whether an individual's income exceeds \$50,000, using a pipeline with one-hot encoding and a Random Forest Classifier (RFC). Unlike other methods, \sys can generate explanations at any pipeline stage, including over raw features. Individuals are ranked by classification score, with explanations shown in Figures~\ref{fig:acs-overall-alaska} and~\ref{fig:acs-overall-texas} (Appendix~\ref{app:exp:scored}).

Figure~\ref{img:acs-overall} shows overall feature importance in Alaska. Hours worked (WKHP), marital status (MAR), age (AGEP), and race (RAC1P) are most influential, followed by education (SCHL), which only matters in the top 20\%. Marital status impacts rank across all strata, while race, marital status, and sex dominate in the top 60\%, 50\%, and 10\% respectively. The top 10\% are mostly white, married, and male; in contrast, education and hours worked vary more but are less important. Feature importance shifts notably in Texas. Education becomes key—especially in the top 10\% and bottom 30\%. Age plays a smaller role, marital status remains influential at both extremes, race has limited impact, and sex is relevant but rarely dominant.

This experiment shows the effectiveness of \sys on higher-dimensional data and highlights nuanced differences in feature importance across data subsets.

\subsection{Comparison to Other Methods}
\label{sec:exp:other}

In this section, we compare explainability methods using the metrics from Section~\ref{sec:metrics}, focusing on a subset from Section~\ref{sec:related}. Since ShaRP and Shapley values target individual explanations, we exclude global methods such as those by Yang et al. and Gale and Marian~\cite{DBLP:conf/sigmod/YangSAHJM18, DBLP:journals/pvldb/GaleM20}. To compare with HIL~\cite{DBLP:conf/hilda/YuanD23}, we adapt their code to support real data and arbitrary score-based rankers (see Appendix~\ref{sec:app:hil}), and focus on their weight-based methods, as their Shapley approximation is already covered by SHAP. We exclude PrefShap~\cite{DBLP:conf/nips/HuCHS22}, which is restricted to pairwise data with a specialized kernel model.

We compare to HRE~\cite{Anahideh2022} but use only four of their internal methods as provided by their public code base (Decision Trees (DT), Linear Regression (LR), Ordinary Least Squares (OLS), and Partial Least Squares (PLS)) and their default neighborhood settings (5-10 consecutive positions above and below the item being explained). We compare to DEXER which fits a linear regression model to the ranks and explains this model using the score-based SHAP instead of the original blackbox, treating rank as a score. Finally, we compare to SHAP~\cite{lundberg2017unified} and LIME~\cite{DBLP:conf/kdd/Ribeiro0G16}, due to their wide use and availability, even though they are not designed for ranking.

\subsubsection{Sensitivity}
\label{sec:exp:sensitivity}
\begin{figure}[b!]
    \centering
        \subfloat[HRE LR]{
        \includegraphics[width=0.5\columnwidth]
        {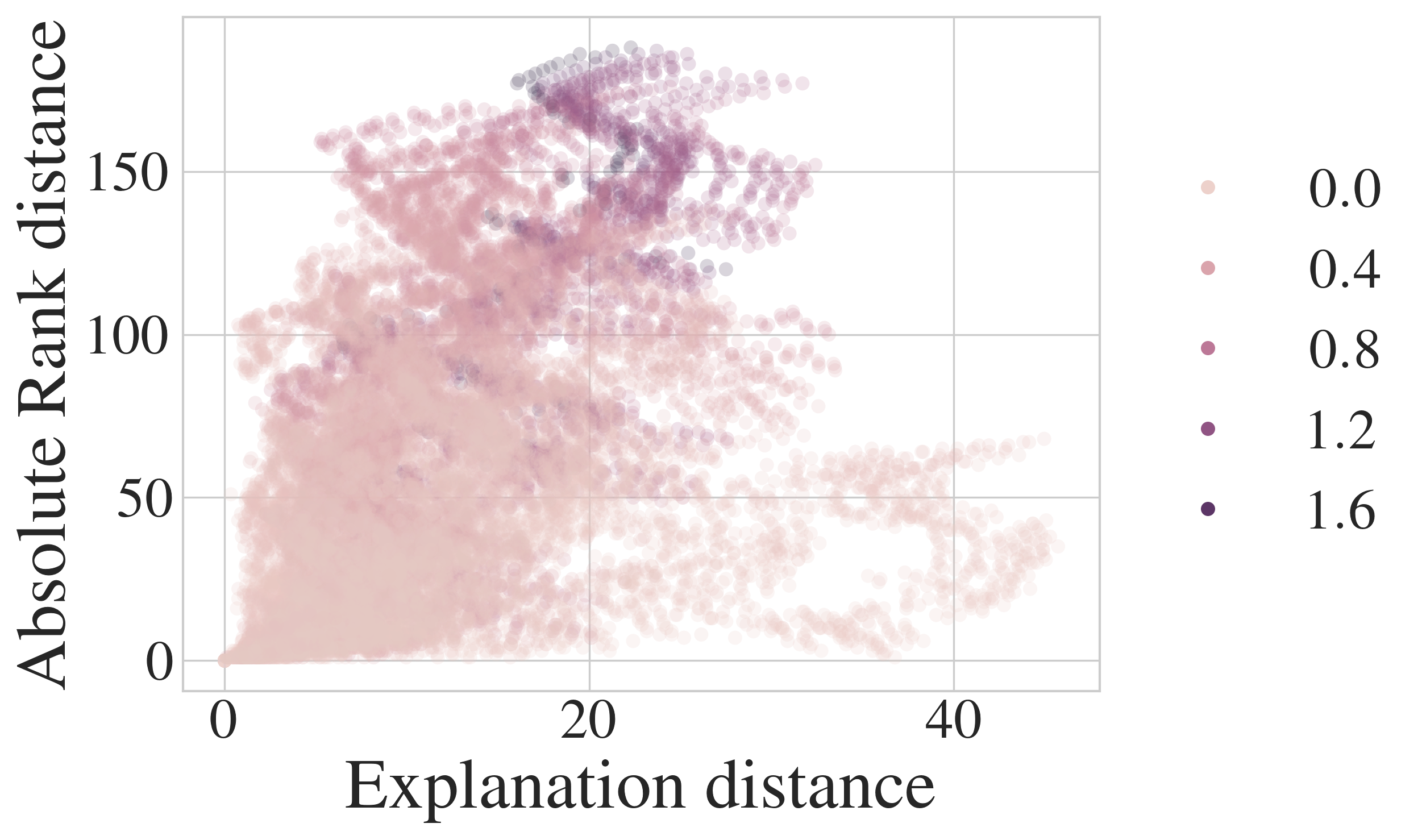}
            \label{img:HRE-LR-CSRank-partial}}
        \subfloat[DEXER]{
        \includegraphics[width=0.5\columnwidth]
        {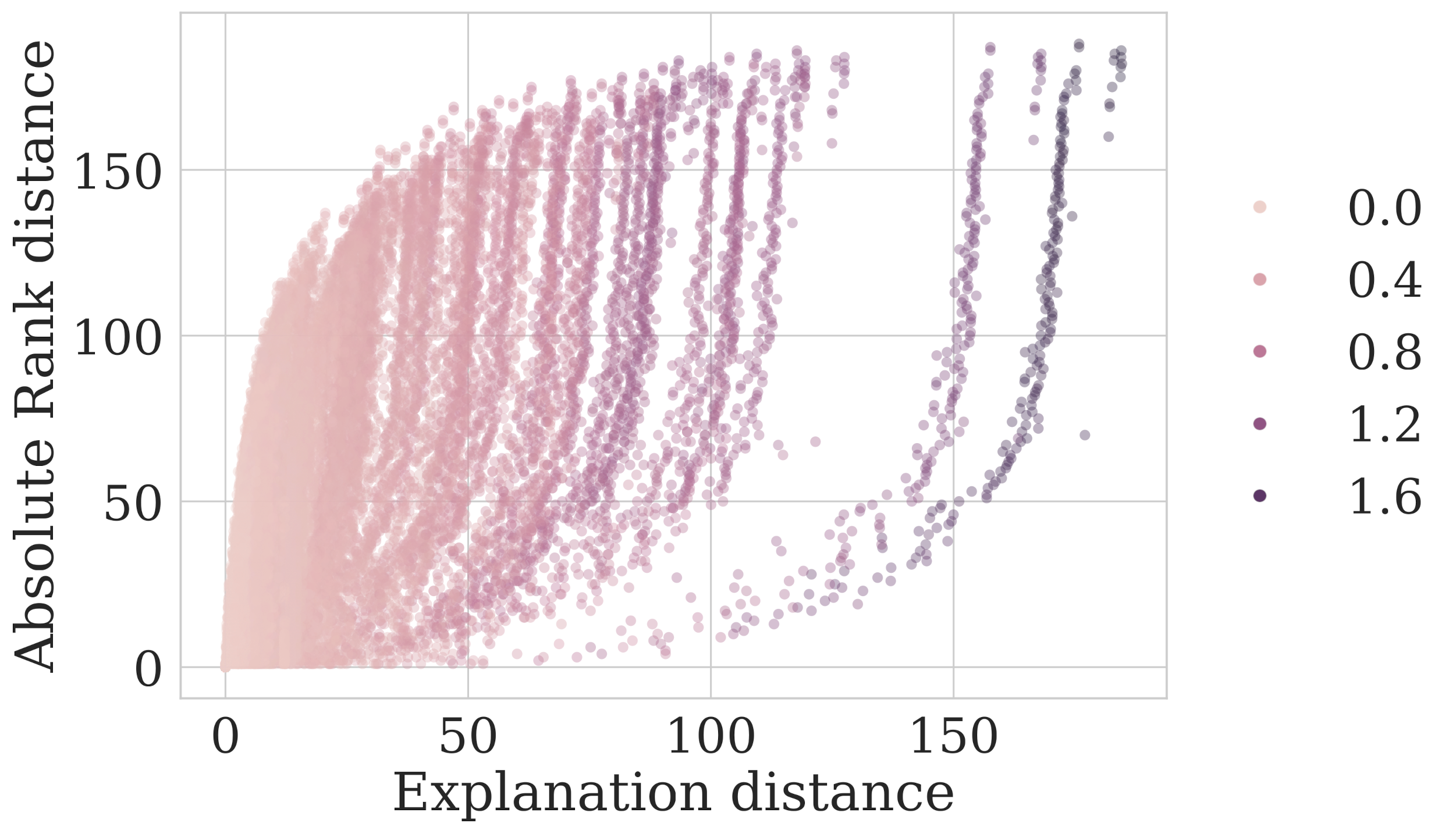}
            \label{img:DEXER-CSRank-partial}}
        \hfill
        \subfloat[LIME]{
		\includegraphics[width=0.5\columnwidth]
        {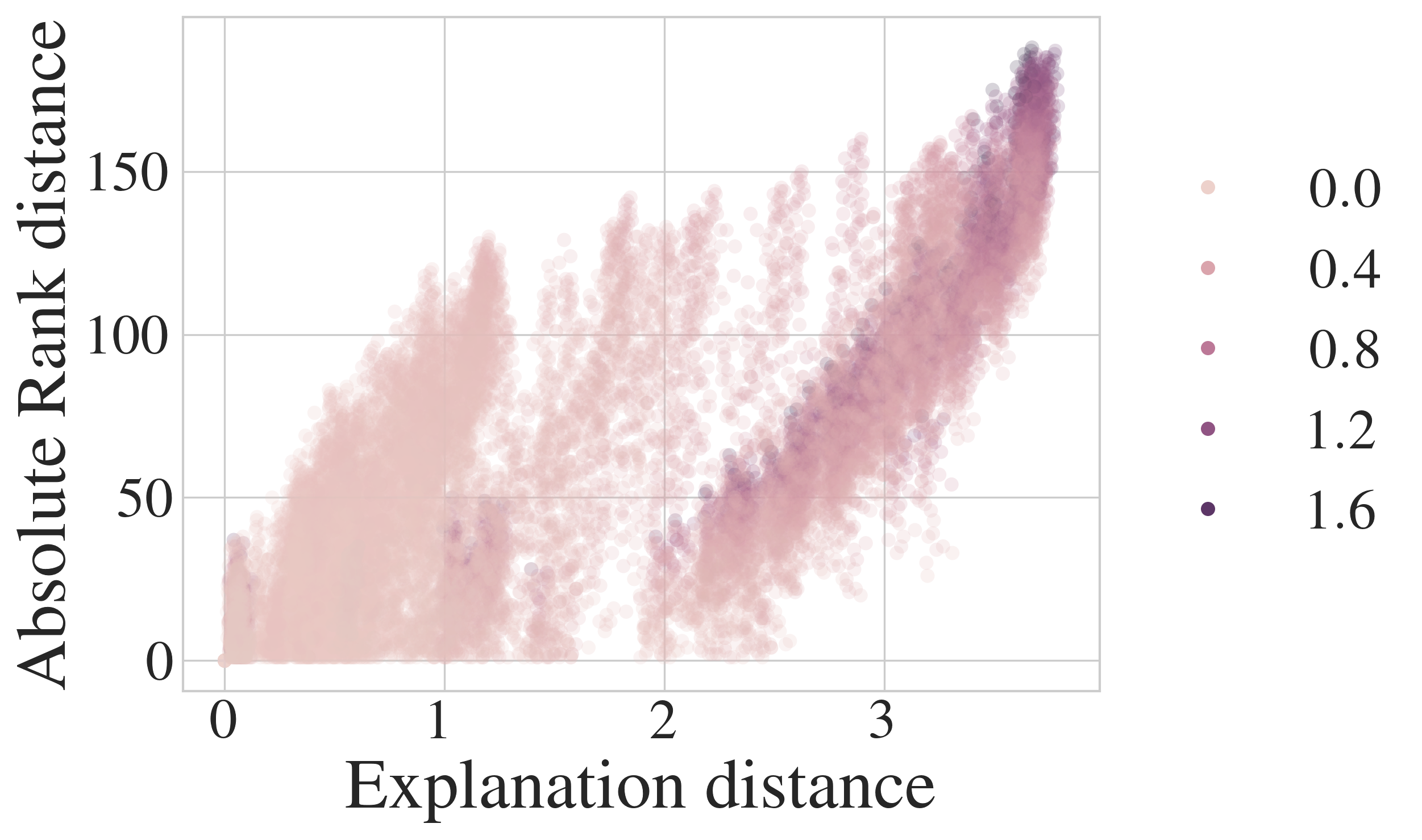}
		\label{img:LIME-CSRank-partial}}
        \subfloat[SHAP]{
		\includegraphics[width=0.5\columnwidth]
        {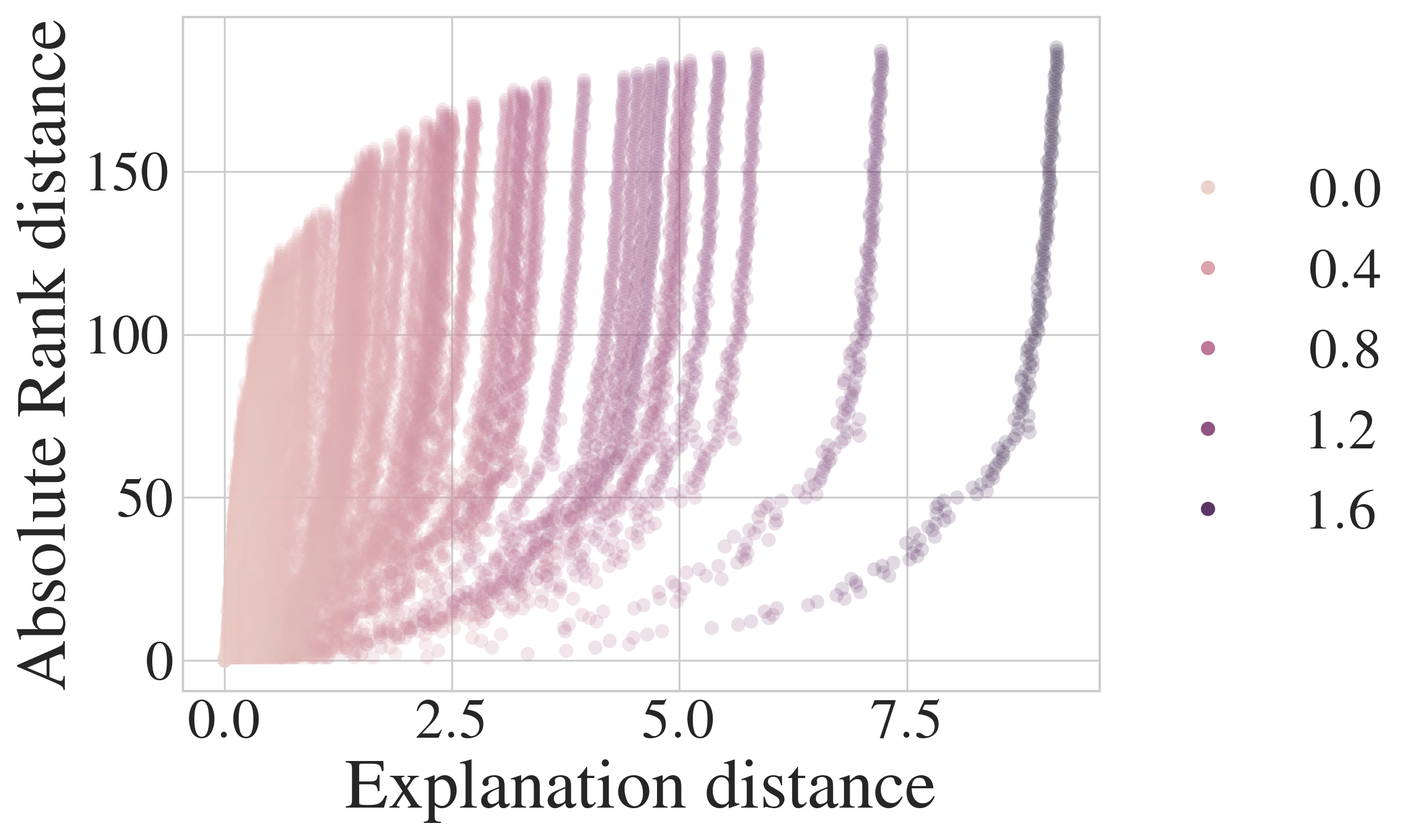}
		\label{img:SHAP-CSRank-partial}}
        \hfill
        \subfloat[ShaRP Rank]{
	    \includegraphics[width=0.5\columnwidth]
        {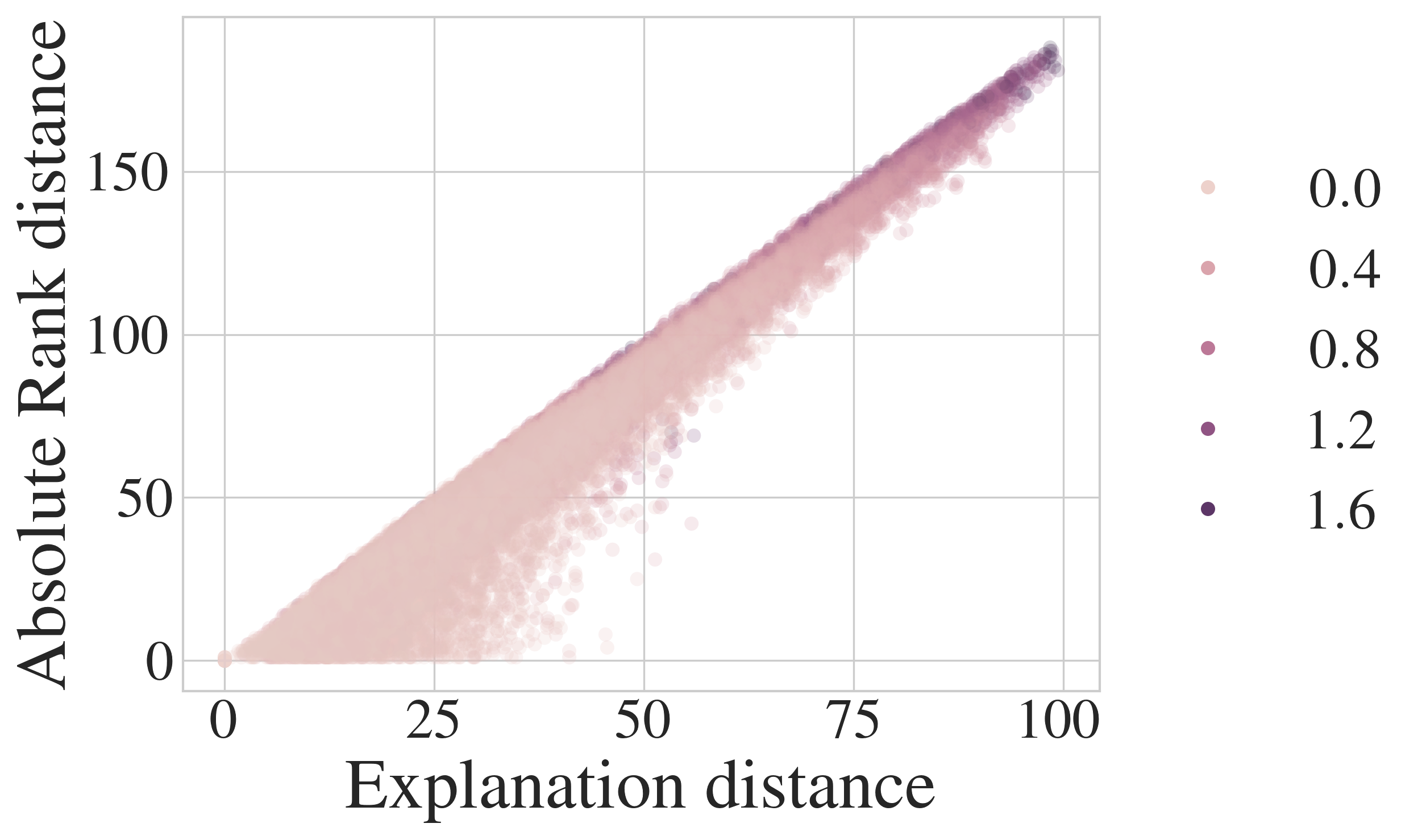}
		\label{img:ShaRP_RANK-CSRank-partial}}
        \subfloat[HIL Std Rank]{
        \includegraphics[width=0.5\columnwidth]
        {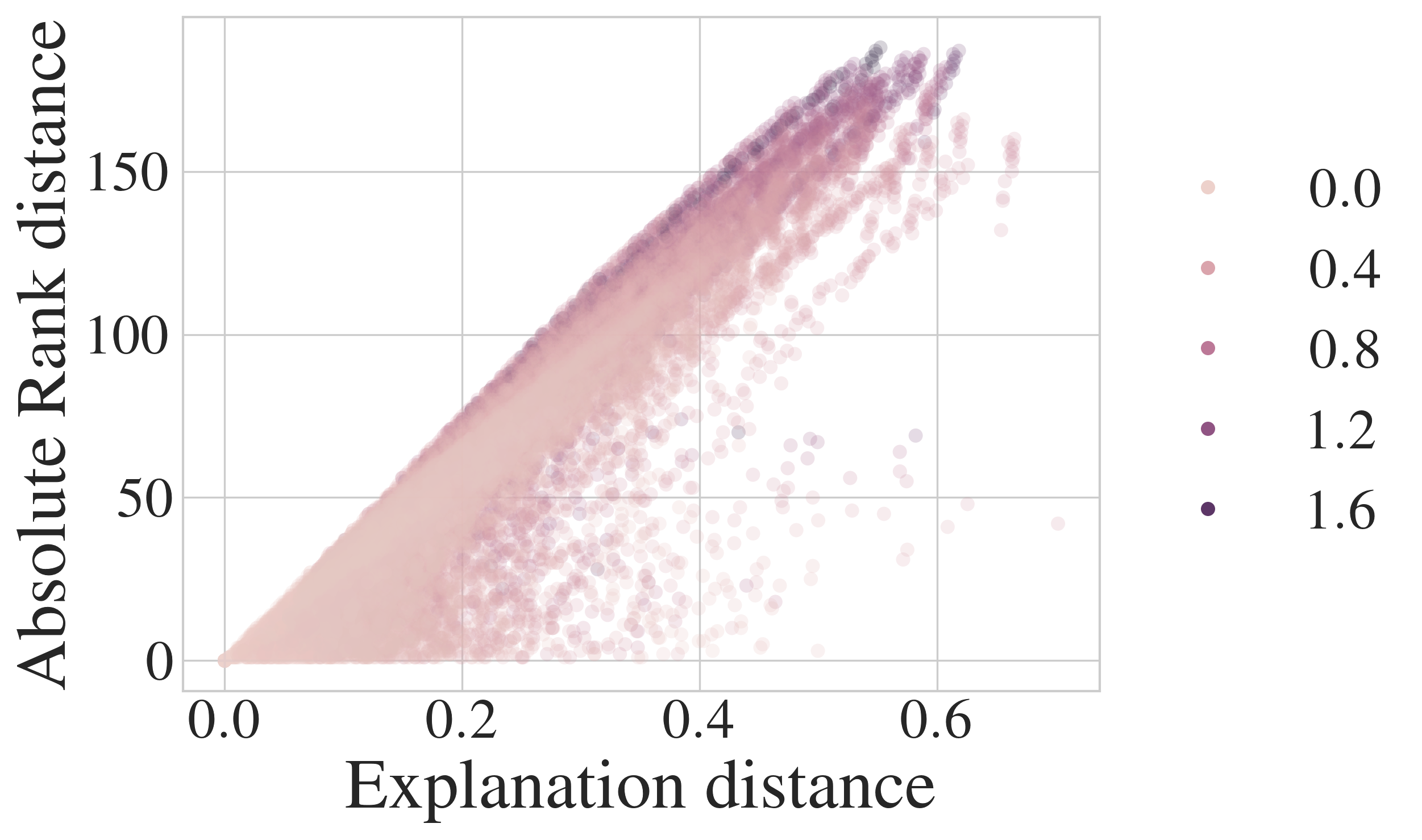}
            \label{img:HIL-Rank-CSRank-partial}}
    \caption{Sensitivity results for CS Rankings. Each dot represents a neighbor of the reference item; the x-axis shows Euclidean explanation distance, the y-axis rank difference, and hue indicates feature similarity. Methods using rank as the profit function (\sys and HIL Std rank) perform best, with \sys leading. These are the only methods that consistently produce similar explanations for items with similar features and outcomes.}
    \label{fig:sensitivity-CSRank-partial}
\end{figure}

Figure~\ref{fig:sensitivity-CSRank-partial} compares the sensitivity of all methods by evaluating explanation similarity for pairs of similar items. For each pair, we compute: (1) Euclidean distance between explanations (x-axis), (2) rank difference (y-axis), and (3) feature distance (hue; lighter means more similar). Each plot centers the reference item at (0,0), with scatter points showing neighbors' distances. Results are overlaid across all items, each used in turn as the reference point.

Intuitively, items with similar features and close rankings should have similar explanations—points should lie near the diagonal $y = x$, with hue darkening as distance grows. In practice, this often fails: a dominant feature may decouple feature and explanation similarity, and dissimilar items can yield similar outcomes. Ideally, explanations should vary for closely ranked items with distinct features and differ significantly for distant ranks, filling the space below $y = x$ with hue darkening outward.

In Figure~\ref{fig:sensitivity-CSRank-partial}, only the rank QoI methods produce the expected shape. Both~\sys (Figure~\ref{img:ShaRP_RANK-CSRank-partial}) and HIL-Std-Rank (Figure~\ref{img:HIL-Rank-CSRank-partial} in our implementation) generate similar explanations for similarly ranked, feature-similar items, with \sys forming slightly denser clusters. In contrast, SHAP (Figure~\ref{img:SHAP-CSRank-partial}), a score-based method, reflects primarily feature distance: its plot shows darkening bands away from the origin, but assigns nearly identical explanations to items with similar features even when their ranks differ substantially.

LIME (Figure~\ref{img:LIME-CSRank-partial}), another score-based method, reflects both feature and rank distance but fails to distinguish explanations as clearly as rank-based methods. Score-based methods generally struggle to capture the nonlinear relationship between score and rank. DEXER (Figure~\ref{img:DEXER-CSRank-partial}), which uses linear regression to predict rank and SHAP for explanations, performs similarly to other score-based approaches. While non-linear models might better approximate rank, our approach directly integrates rank into the Shapley value utility. HRE (Figure~\ref{img:HRE-LR-CSRank-partial}) shows no clear pattern with respect to rank or feature distance; similar and dissimilar explanations appear across all ranks and hues. This is expected, as HRE depends on local neighbors, which can vary widely in features and outcomes.

In Appendix~\ref{sec:app:comparisons}, we provide additional sensitivity results, comparing \sys with rank QoI to HIL Std rank and analyzing a score-based task. We show that \sys outperforms HIL across datasets and that ShaRP with score QoI aligns well with the diagonal in score-based tasks—underscoring the importance of choosing a QoI aligned with the explanation goal.

In summary, explanations for the rank QoI, which we are introducing in this paper, are able to more accurately explain ranking tasks compared to other local feature-based explanation methods.

We also quantified agreement between explanations produced by different methods.  We show these results in Appendix~\ref{app:exp:agreement}.

\subsubsection{Fidelity}
\label{sec:exp:fidelity}
It is possible to calculate Fidelity for SHAP, LIME, ShaRP, and the HIL-score. It is impossible to compute Fidelity for HIL-rank and all the HRE methods. All methods except HIL-score perform very well.
We compute the Fidelity averaged across all items in all datasets.  All methods are executed using their recommended settings to compute explanations for score QoI. Additionally, we compute fidelity for~\sys for the rank QoI.  Recall that~\sys is the only method that can compute an explanation for this QoI.
LIME, SHAP and~\sys are all achieving high explanation fidelity, on average ranging from 0.94-0.98, 0.97-1.00 and 1.00 correspondingly.  HIL has reasonable fidelity for CSR (0.85) but does not perform consistently on other datasets ranging from 0.14-0.64. See Table~\ref{table:fidelity} in Appendix~\ref{sec:app:comparisons} for details.

\begin{figure*}[t!]
    \centering
        \subfloat[Speedup vs. sample size, ACS]{
        \includegraphics[width=0.46\columnwidth]{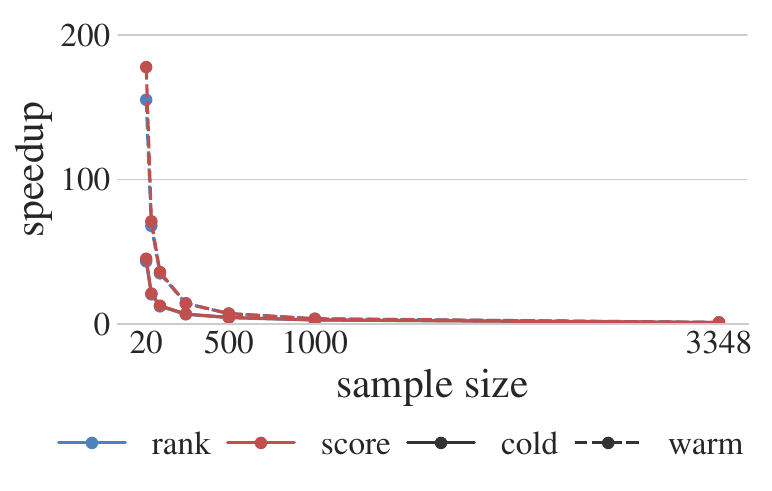}
            \label{img:ACS-sample-speedup}}
        \subfloat[Fidelity vs. sample size, ACS]{
        \includegraphics[width=0.5\columnwidth]{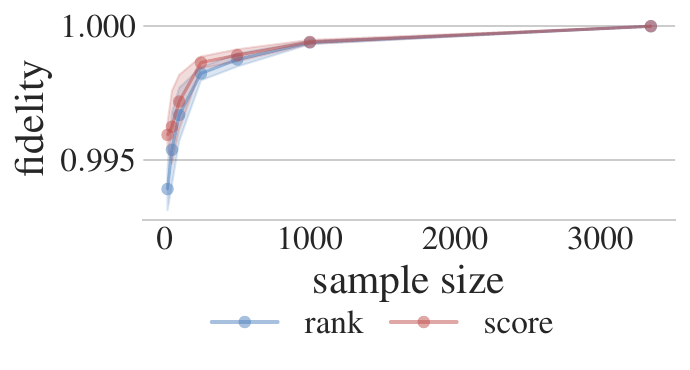}
            \label{img:ACS-sample-fidelity}}
        \subfloat[Speedup vs. max coalition size, ACS]{
        \includegraphics[width=0.58\columnwidth]{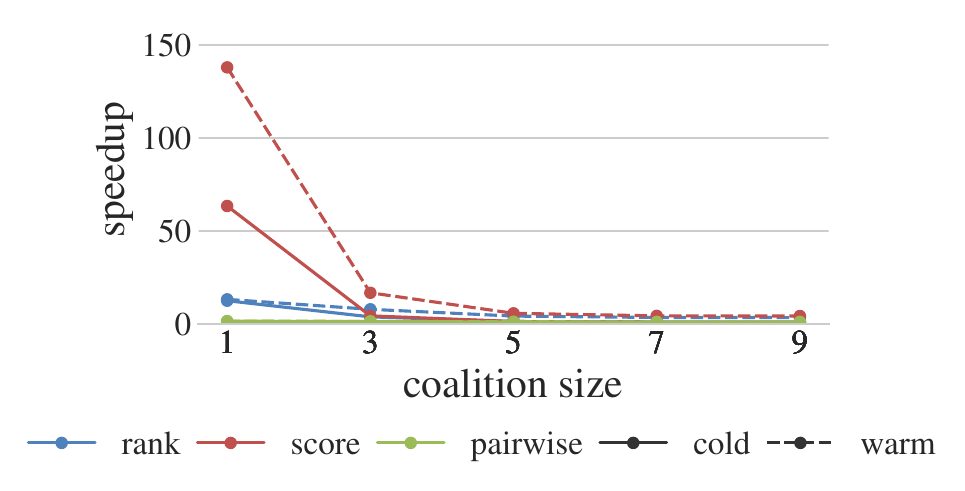}
            \label{img:ACS-coalition-speedup}}
        \subfloat[Fidelity vs. max coalition size, ACS]{
        \includegraphics[width=0.48\columnwidth]{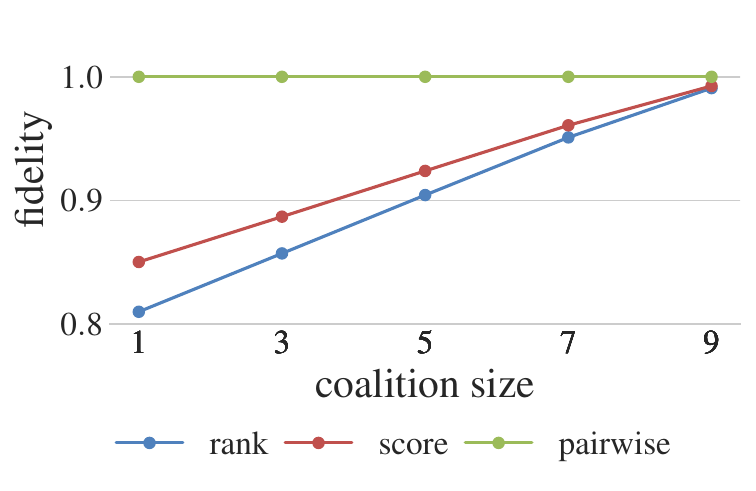}
            \label{img:ACS-coalition-fidelity}}
      \caption{Running time of approximation for ACS Income (AK).  In (a) and (b), max coalition size is 9; in (c) and (d), sample size is 100. Speedup is computed vs. to exact times in Table~\ref{tab:time:exact}, see Table~\ref{tbl:acs-ak-times-full} in the Appendix for additional information. Due to a slight difference in the tie breaking method, the dataset's size (and maximum sample size) was set to 3,348.}
    \label{fig:time-approx}
\end{figure*}

\subsection{Efficiency and Approximation}
\label{sec:exp:efficiency}

\begin{table}[b!]
\caption{Running time of exact computation, cold start.}
\centering
\small 
\begin{tabular}{ l r c r r r} 
  &  &  & \multicolumn{3}{c}{avg. time (sec)}  \\ 
 dataset & \# tuples & \# features & score & rank & pair \\
 \midrule
 \texttt{ATP} & 86  & 6  & 0.004 & 0.026 & 0.004 \\
 \texttt{CSR} & 189 & 5  & 0.002 & 0.022 & 0.003 \\
 \texttt{THE} & 1,397 & 5 & 0.011 & 0.423 & 0.007 \\
 \texttt{SYN} & 2,000 & 3 & 0.002 & 0.126 & 0.003 \\
\texttt{ACS-AK} & 3,546  & 10 & 1,960.7 & 1,956.8 & 2.53 \\
 \bottomrule
\end{tabular}
\label{tab:time:exact}
\end{table}

\begin{table*}[t!]
\caption{Running time of optimized computation. Running times are reported per data point, in seconds. Parameter optimization was performed separately for each dataset. The optimal RFC for ACS (AK) used 100 estimators, compared to 10 for ACS (TX), resulting in faster cold-start inference per tuple for ACS (TX).}
\centering
\small 
\begingroup
\begin{tabular}{ l r c | c c c | r r r | l l l } 
  &  &  & & & & \multicolumn{3}{c}{avg. time (sec)} & \multicolumn{3}{c}{fidelity} \\ 
 dataset & \# tuples & \# features & start & max coal. size & sample size & score & rank & pair & score & rank & pair \\
 \midrule
 \texttt{ACS (AK)} & 3,348  & 10 & cold & 9 & 100 & 143.54  & 151.28 & 1.97 & 0.997 & 0.997 & 1.0 \\
 \texttt{ACS (AK)} & 3,348  & 10 & warm & 9 & 100 & 40.42  & 41.56 & 1.64 & 0.997 & 0.997 & 1.0 \\
 \texttt{ACS (AK)} & 3,348 & 10 & warm & 9 & 20   & 8.09  & 9.45  & 1.64 & 0.996 & 0.994 & 1.0 \\
 \texttt{ACS (AK)} & 3,348 & 10 & warm & 7 & 20   & 7.95  & 9.28  & 1.64 & 0.960 & 0.951 & 1.0 \\
 \texttt{ACS (AK)} & 3,348 & 10 & warm & 5 & 20   & 6.07  & 7.37  & 1.56 & 0.923 & 0.904 & 0.9 \\
 \texttt{ACS (AK)} & 3,348 & 10 & warm & 3 & 20   & 2.08  & 3.39  & 1.35 & 0.886 & 0.856 & 0.9 \\
 \texttt{ACS (AK)} & 3,348 & 10 & warm & 2 & 20   & 0.74  & 2.05  & 1.27 & 0.868 & 0.833 & 0.9 \\
 \midrule
 \texttt{ACS (TX)} & 135,924  & 10 & cold & 9 & 100 & 126.39 & 139.69 & 7.69 & 0.998 & 0.997 & 1.0 \\
 \texttt{ACS (TX)} & 135,924  & 10 & warm & 9 & 100 & 40.42 & 48.79 & 7.65 & 0.998 & 0.997 & 1.0 \\
 \texttt{ACS (TX)} & 135,924  & 10 & warm & 9 & 20  & 8.07 & 16.28 & 7.59 & 0.992 & 0.989 & 1.0 \\
 \texttt{ACS (TX)} & 135,924  & 10 & warm & 7 & 20  & 7.95 & 16.35 & 7.69 & 0.973 & 0.959 & 0.9 \\
 \texttt{ACS (TX)} & 135,924  & 10 & warm & 5 & 20  & 6.27 & 14.33 & 7.50 & 0.944 & 0.913 & 0.9 \\
 \texttt{ACS (TX)} & 135,924  & 10 & warm & 3 & 20  & 2.52 & 12.46 & 7.84 & 0.911 & 0.864 & 0.8 \\
 \texttt{ACS (TX)} & 135,924  & 10 & warm & 2 & 20  & 0.93 & 10.98 & 7.13 & 0.894 & 0.839 & 0.8 \\
 \bottomrule
\end{tabular}
\endgroup 
\label{tab:time:opt}
\end{table*}

\subsubsection{Running time of exact computation} 
\label{sec:exp:efficiency:exact}

In our first experiment, we measure the exact computation time for the rank and score QoIs, and the pairwise method with rank QoI, on three real and one synthetic dataset from Table~\ref{table:datasets}. We include only one synthetic dataset, as all have the same size ($m=2{,}000$) and at most three features; differences in correlation structure do not affect runtime. We omit the top-$k$ QoI, as its implementation mirrors the rank QoI, resulting in indistinguishable runtimes.

Table~\ref{tab:time:exact} presents the results, reporting the time to generate an explanation per point, averaged over 100 points for CSR, THE, SYN, and ACS-AK, and over 83 points (dataset size) for ATP. Runtime for rank and score QoIs increases with both the number of items ($m$ in Algorithm~\ref{alg:sharp}) and features ($d$), as exact computation scales linearly with $m$ and exponentially with $d$ ($2^d - 1$ coalitions). Pairwise methods involve only two items, so their runtime is independent of $m$ but remains exponential in $d$. Our pairwise method for rank QoI also requires recomputing ranks after each intervention (line 7, Algorithm~\ref{alg:iota_rank}), which scales linearly with $m$ in our implementation. This explains why pairwise QoI for THE ($m=1{,}397$, $d=5$) runs slower than for ATP ($m=86$, $d=6$). Exact computation is particularly challenging for ACS-AK due to its higher feature count. We next demonstrate how approximations can mitigate this cost.

\subsubsection{Running time and quality of approximation}
\label{sec:exp:efficiency:approx}
To reduce runtime, we implement two approximation methods: limiting the number of samples and bounding coalition size. We report running time and fidelity (Eq.~\ref{eq:fidelity}) to assess approximation quality. Figure~\ref{fig:time-approx} shows results for ACS-AK, see Appendix~\ref{sec:app:approx} for ATP and CS Rankings.

Figure~\ref{img:ACS-sample-speedup} shows the speed-up achieved by reducing the number of samples $m$. Lowering $m$ from $1,348$ (exact) to 20, while maintaining a maximum of size 9 coalitions (the largest possible for 10 features), accelerates rank QoI by a factor of 79, reducing runtime from 1956 sec to 45 sec.  Crucially, this performance gain does not compromise fidelity, which remains above 0.99 (out of 1) across all sample sizes in all experiments.
Figures~\ref{img:ACS-coalition-speedup} and~\ref{img:ACS-coalition-fidelity} show speed-up and fidelity when bounding coalition size. The largest speed-up occurs for coalition size 1, though fidelity is lower: at least 0.81 for rank and 0.85 for score (fidelity is 1 for pairwise). Fidelity improves with coalition size 3, reaching 0.86 for rank and 0.89 for score.

Table~\ref{tab:time:opt} shows per-tuple explanation times across different maximum coalition and sample sizes, highlighting the trade-off between runtime and fidelity. For large datasets, approximate methods yield substantial speedups with minimal fidelity loss. In ACS (AK), for example, a ranking can be explained in 9.45 seconds (vs. 1,956 seconds for exact computation). Warm start is typically 3 times faster than cold start, and pairwise explanations are the fastest overall. Figures~\ref{img:ACS-sample-fidelity} and~\ref{img:ACS-sample-speedup} illustrate how fidelity and runtime vary with sample size. As shown in Table~\ref{tbl:acs-ak-times-full}, runtime grows linearly with sample size, while fidelity decreases gradually, reflecting a favorable accuracy–efficiency trade-off.

In summary, reducing the number of samples and bounding coalition size improves runtime while maintaining high explanation fidelity. Computing Shapley values is exponential in the number of features, and it is common to develop model-specific approximations for explainers like SHAP~\cite{lundberg2017unified}. Designing more sophisticated custom optimizations for our QIIs is in our immediate plans.

\section{User Study}
\label{sec:focus_group}

We conducted an IRB-approved study (NYU IRB-FY2025-9983) to explore how users interpret rank-based vs. score-based explanations, using CS Rankings.  We summarize the study protocol and the results, see Appendix~\ref{app:study_protocol} and ~\ref{app:study_materials} for details.

\emph{Participant recruitment and study protocol.} Through our institution, we recruited 13 participants: 6 PhD students, 3 postdocs, 2 professors, and 2 research staff. All completed forms detailing their academic backgrounds and familiarity with explainability and the dataset. Students and postdocs, all from CS, reported moderate to high familiarity with explainability. Professors and staff, with social science backgrounds applied to AI, showed varied familiarity with explainability. CS Rankings familiarity ranged from high to low, independent of seniority.

Participants were divided into \grpRank (7 people) and \grpScore (6 people). Both groups received an introductory document corresponding to their group, completed a range of tasks that included either rank-based or score-based explanations, and then participated in a discussion. Each participants answered 22 questions, divided into 3 categories: understanding the rank of a specific department (3 departments $\times$ 4 questions), understanding why one department is ranked higher than another (3 department pairs $\times$ 2 questions), and understanding feature importance trends across the ranking (2 sets of 6 departments $\times$ 2 questions).

\emph{Results.} \grpRank outperformed  \grpScore in terms of accurately answering questions (73\% vs. 67\%), and also reported higher confidence (4.15 vs. 3.90 on a 5-point Likert scale), see Table~\ref{tab:study_results} in Appendix~\ref{app:study_protocol}). Notably, \grpScore expressed greater distrust in the ranking and the dataset, echoing findings from~\cite{aechtner2022comparing}, for example: \emph{``Maybe my mind started looking for some kind of [...] preconceived biases and wondering? [...] There was one figure [...] towards the end. The difference was almost imperceptible, and I kept thinking, why is one ranked few points higher than the other?''}

Several \grpScore participants noted needing multiple explanations to understand the ranking, as score-based explanations lack rank context. For example: \emph{``At first [for the items at the top of the ranking], the differences were so big that [the answer] was very clear, and then at the end, you know which one is better 1.05 or 1.08 [...]? So it makes you want to go back to the earlier questions and makes you question your initial impression and understanding of [the ranking].''}.

While further study is needed to understand the sources of mistrust and validate findings with more participants, our results provide preliminary evidence that rank-based explanations better support understanding and trust in ranking tasks as compared to score-based explanations. Most importantly, several participants underscored that they found feature-based explanations useful.  For example: \emph{``I thought that the experience is successful on raising awareness and provoking critical thinking about using rankings.''}

\section{Conclusions}
\label{sec:conc}

We introduced a comprehensive framework for quantifying feature importance in selection and ranking. Given the impact of rankers on individuals, organizations, and populations, understanding their decisions is crucial for \emph{auditing and compliance} (ensuring legal adherence), \emph{recourse} (helping individuals improve outcomes), and \emph{design} (optimizing ranking procedures). Our work addresses the interpretability needs of these tasks.

We demonstrated the effectiveness of~\sys~through a qualitative analysis of an impactful real-world task---the ranking of Computer Science departments. This was complemented by an evaluation on real and synthetic datasets, revealing that our defined profit functions provide valuable and complementary insights beyond simple score-rank relationships.  We showed that feature importance varies with data distribution \emph{even when} the scoring function is fixed and exhibits locality. Finally, we compared \sys~to other local feature-based explanation methods, showing it performs favorably. ~\sys is an open-source Python library, and is the only available library for explaining ranked outcomes in tabular data.

\section{Acknowledgments}
\label{sec:ack}

This research was supported in part by NSF Awards No.~2326193 and 2312930. Ivan Shevchenko and Kateryna Akhynko conducted this work through the \href{https://r-ai.co/ukraine}{RAI for Ukraine} program of the NYU Center for Responsible AI and supported in part by the Simons Foundation (SFARI Award \#1280457, JS). We thank Tilun Wang for contributions to the earlier code base, and Lucius Bynum and Falaah Arif Khan for helpful discussions.

\balance
\bibliographystyle{ACM-Reference-Format}
\bibliography{main}

\newpage
\appendix
\label{appendix}
\section{Datasets}
\label{sec:app:data}

\begin{figure}[b!]
    \centering
        \subfloat[Tennis (ATP)]{
        \includegraphics[width=0.45\columnwidth]
        {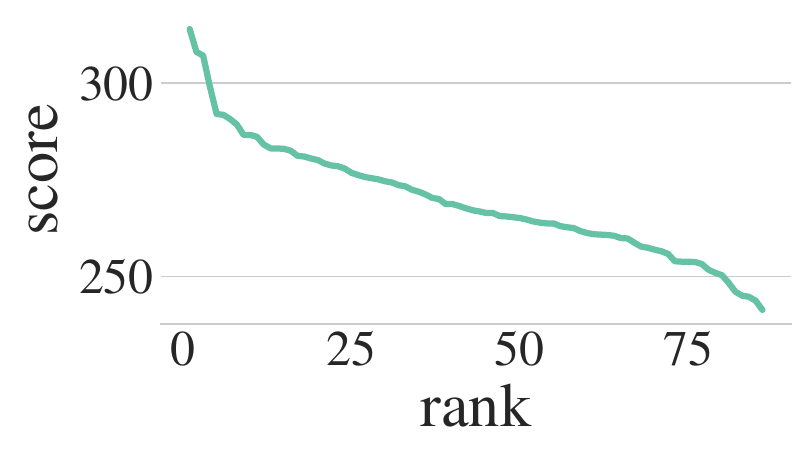}
            \label{img:ATP-rankvsscore}}
        \subfloat[\csr (CSR)]{
		\includegraphics[width=0.45\columnwidth]
        {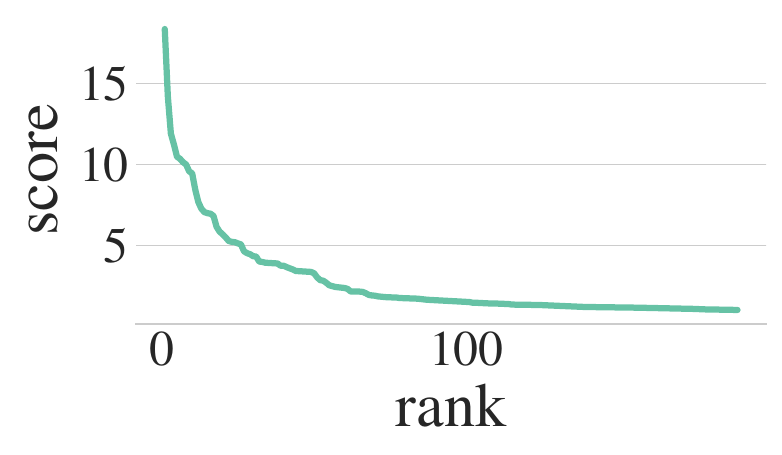}
		\label{img:CSR-rankvsscore}}
        \hfill
	\subfloat[Times Higher Education (THE)]{
		\includegraphics[width=0.45\columnwidth]
        {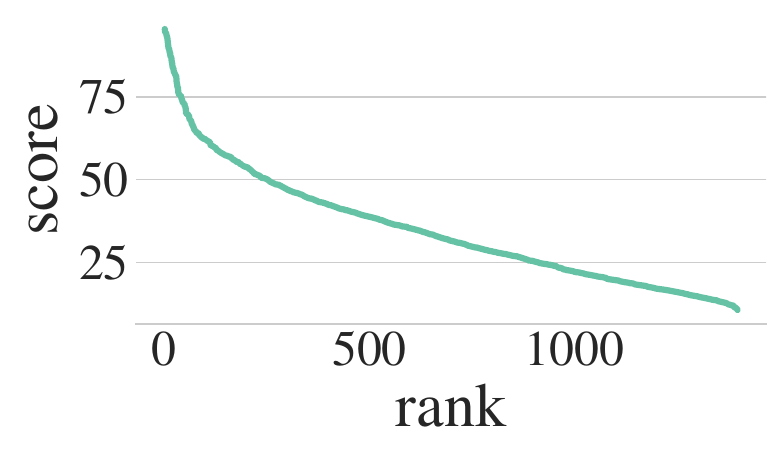}
		\label{img:THE-rankvsscore}}
        \subfloat[a representative synthetic dataset]{
        \includegraphics[width=0.45\columnwidth]
        {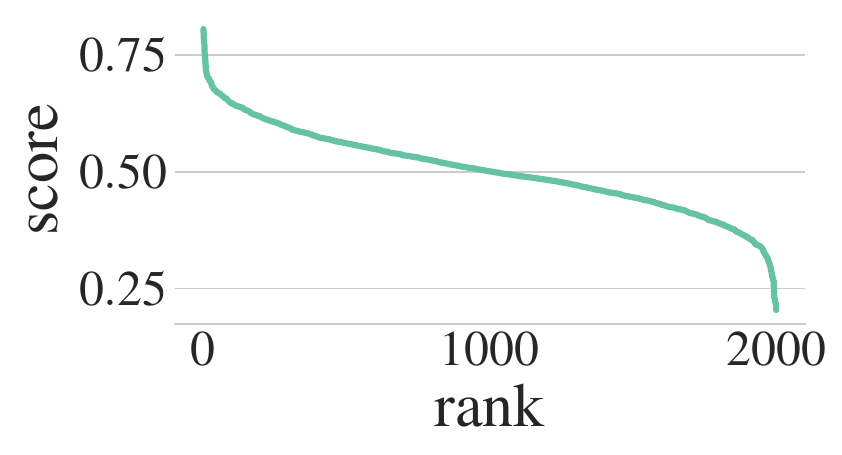}
            \label{img:Syn0-rankvsscore}}
        \hfill
        \subfloat[Moving Company - train]{
        \includegraphics[width=0.45\columnwidth]
        {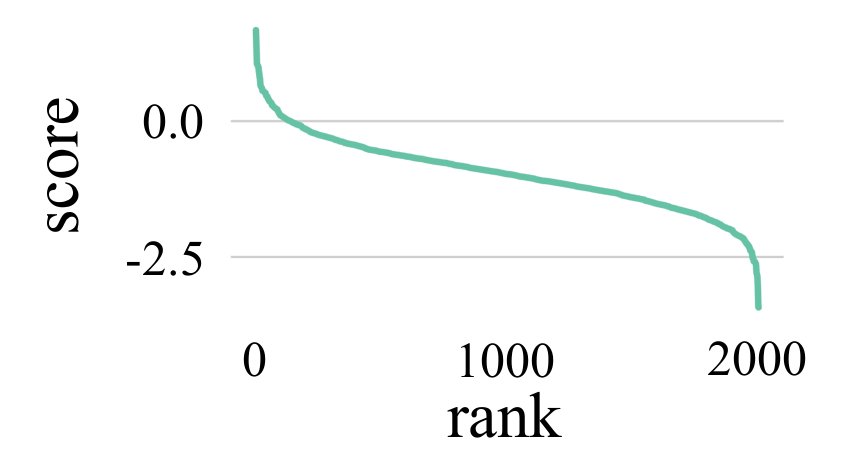}
            \label{img:movers}}
        \subfloat[Moving Company - test]{
        \includegraphics[width=0.45\columnwidth]
        {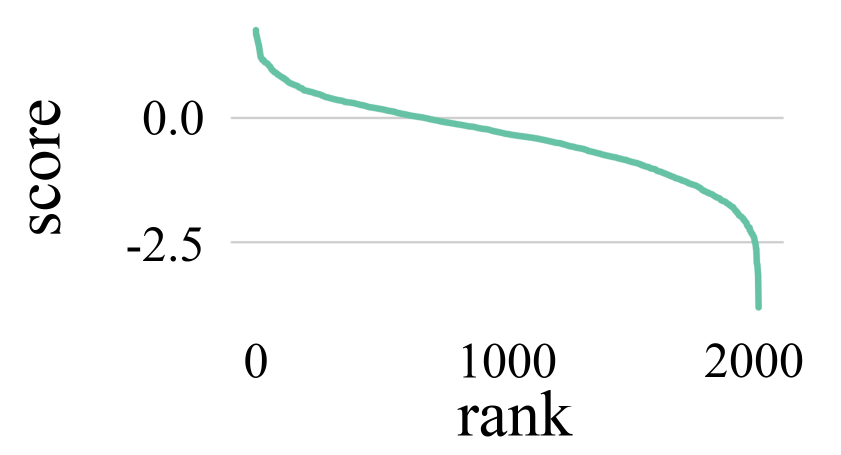}
            \label{img:movers-test}}
        \hfill
        \subfloat[ACS Income (Alaska)]{
        \includegraphics[width=0.45\columnwidth]
        {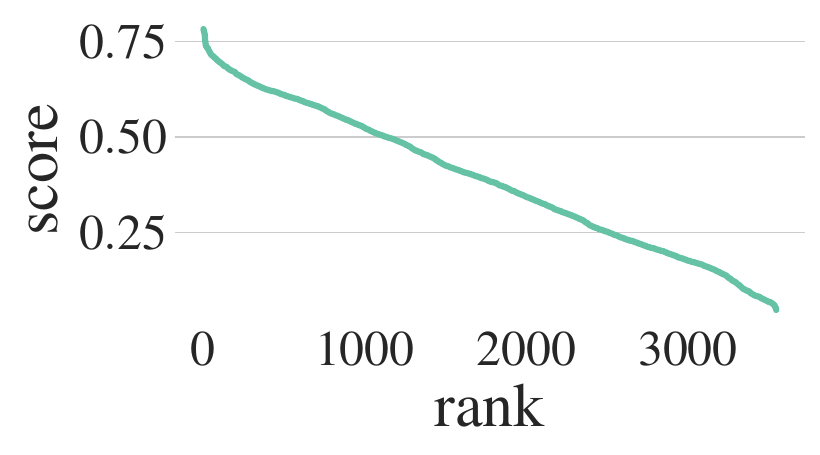}
            \label{img:acs-income-alaska}}
        \subfloat[ACS Income (Texas)]{
        \includegraphics[width=0.45\columnwidth]
        {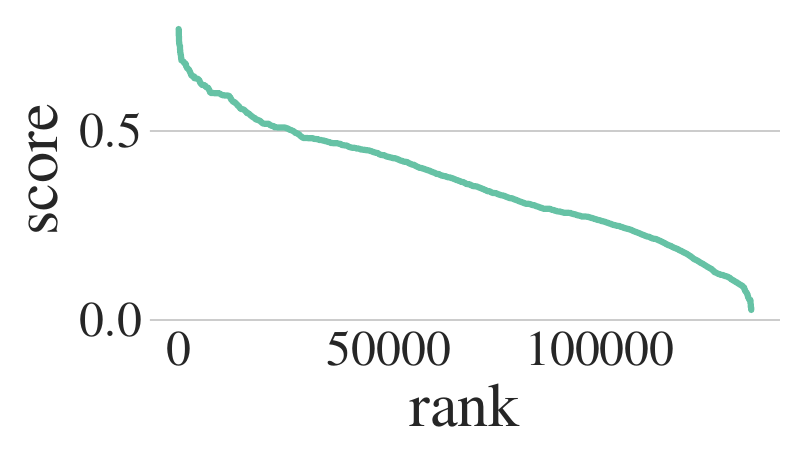}
            \label{img:acs-income-texas}}

    \caption{The relationship between an item's score ($y$-axis) and its rank ($x$-axis) for four score-based tasks.}
    \label{fig:rank-vs-score-all}
\end{figure}

We evaluate the performance of~\sys and compare it to other local feature importance methods, using several real and synthetic datasets, with the corresponding ranking tasks.  Dataset properties, along with ranker type (score-based or learned), are summarized in Table~\ref{table:datasets} and described below.  We show the relationship between score and rank for score-based ranking tasks in Figure~\ref{fig:rank-vs-score-all}.

\emph{\csr (CSR)} ranks 189 Computer Science departments in the US based on a normalized publication count of the faculty across 4 research areas: AI, Systems (Sys), Theory (Th), and Interdisciplinary (Int)~\cite{CSRankings_2023}.  We use publication data for 2013-2023, with the scoring function provided by \csr, a geometric mean of the adjusted counts per area, with \# of sub-areas as exponent:
\begin{equation*}
    f = \sqrt[27]{(AC_{AI}^5 + 1)(AC_{Sys}^{12} + 1)(AC_{Th}^3 + 1)(AC_{Int}^7 + 1)}
\end{equation*}

\emph{ATP Tennis (ATP)} is based on publicly available 2020-2023 performance data of tennis players from the Association of Tennis Professionals (ATP)~\cite{ATP_Tennis}. 
We use 2022 data that includes 5 performance-related attributes of 86 players.  We select 2022 because this is the year in which data for all 5 attributes is available for the highest number of players.
We use the following scoring function that we recovered from the ATP site using the scores:
\begin{equation*}
\begin{aligned}
    f = 100 \times (\mbox{\% 1st Serve}) + 100 \times (\mbox{\%1st Serve Points Won}) &+\\
    100 \times (\mbox{\%2nd Serve Points Won}) + 100 \times (\mbox{\% Service Points Won}) &+\\ 100 \times (\mbox{Avg Aces/Match}) - 100 \times (\mbox{Avg Double Faults/Match})
\end{aligned}
\end{equation*}

\emph{Times Higher Education (THE)} is a dataset of worldwide university rankings~\cite{Times_Higher_Ed}.
It contains the university name, country, and the scores assigned to the university by Times Higher Education for teaching (TEA), research (RES), citations (CIT), income (INC), and international students (INT).
We use 2020 data, for consistency with Anahideh and Mohabbati-Kalejahi~\cite{Anahideh2022} who also used it in their paper, with the scoring function provided by THE:
\begin{equation*}
    f = 0.3 \times TEA + 0.3 \times RES + 0.3 \times CIT + 0.025 \times INC + 0.075 \times INT
\end{equation*}

\paragraph*{Moving Company}
The moving company scenario \cite{DBLP:conf/forc/YangLS21} simulates a hiring process where job applicants are ranked based on their \emph{qualification score}, computed as a function of their weight lifting ability, sex, and race. We train two different rankers, over two scenarios:
\begin{enumerate}
    \item Using the original data from a previous hiring process from that company, where female applicants generally display lower weight-lifting ability than male applicants and a lower qualification score. In addition, black applicants have a lower qualification score compared to white applicants, but similar weight-lifting ability. Hence, black females face greater discrimination compared to the rest of the applicants.
    \item After applying the intersectional fairness intervention proposed in the same paper to the data.
\end{enumerate}

All versions of this dataset (both scenarios and train/test sets) contain 2000 tuples.

We use an eXtreme Gradient Boosting (XGB) and a Light Gradient Boosting (LGB) Machine to model the rankings of the applicants in the training set, and infer and compute the feature contributions of the applicants in the test set, i.e., all results reported in this section correspond to the test set. The XGB ranker was defined with the pairwise ranking objective, while the LGB ranker uses the lambdarank objective.

\paragraph*{ACSIncome} ACSIncome contains income-related data from adults in the US. It consists of 10 features: age, class of worker, educational attainment, marital status, occupation, place of birth, relationship to the reference person, work hours per week, sex, and race. The task is to predict whether the yearly income is over \$50,000.

For this task, we use a Random Forest Classifier (RFC) and rank the items based on the predicted probability of positive class membership.

\paragraph*{Synthetic datasets} We also use numerous synthetic datasets to showcase specific quantitative and qualitative aspects of local feature-based explanations and metrics, and to study specific aspects of performance. These datasets contain 2,000 tuples.
In five of them, items have 2 features, $x_1$ and $x_2$, distributed according to the uniform, Gaussian, or Bernoulli distributions, with varying parameters. We experiment with both independent and correlated features.  Each synthetic dataset consists of $2,000$ items.  We use three linear scoring functions: $f_1 = 0.8 \times x_1 + 0.2 \times x_2$, $f_2 = 0.5 \times x_1 + 0.5 \times x_2$, and $f_3 = 0.2 \times x_1 + 0.8 \times x_2$.

To explore correlations further, we create three datasets that have three Normal features $x_1$, $x_2$, and $x_3$, and 2,000 items. In the first dataset, all features are independent. In the second, we draw $x_1$ and $x_2$ from the 2D Gaussian, and they are negatively correlated with a correlation of -0.8. The third feature $x_3$ is independent. For the third dataset, we draw the features from the 3D Gaussian. $x_1$ and $x_2$ are negatively correlated with correlation -0.8, $x_1$ and $x_3$ are positively correlated with correlation 0.6, and $x_2$ and $x_3$ are negatively correlated with correlation -0.2. For all three datasets, we use the same scoring function $f_4 = 0.33 \times x_1 + 0.33 \times x_2 + 0.34 \times x_3$.

\section{Distributional Analysis for Ranking}
\label{sec:app:score-based}
\begin{figure}
\includegraphics[width=0.9\columnwidth]{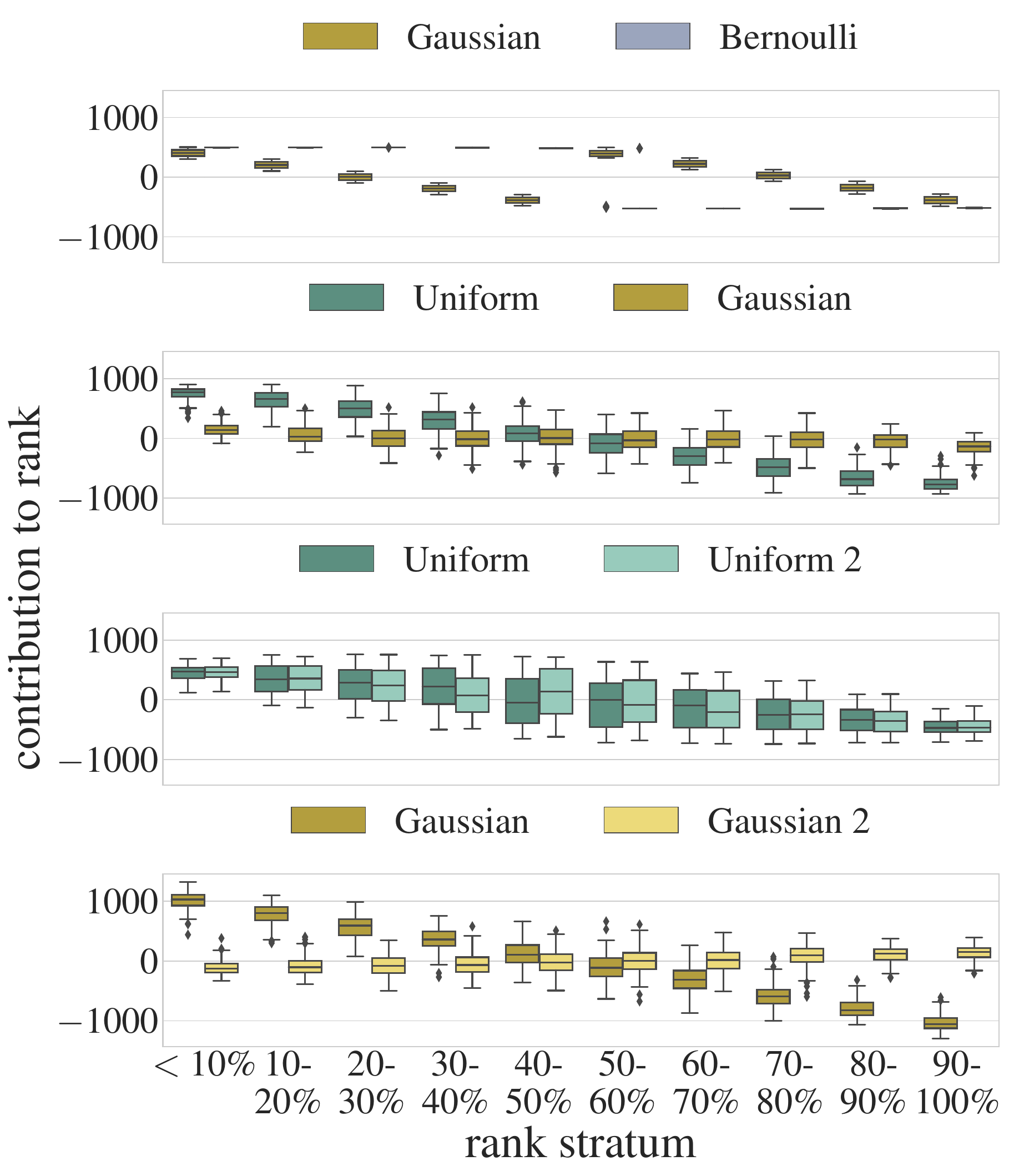}
\caption{The rank QoI using $f_2 = 0.5 \times x_1 + 0.5 \times x_2$ for four datasets; $D_1$: $x_1 \sim  N(0.5, 0.1)$, $x_2 \sim Bern(0.5)$; $D_2$: $x_1 \sim  [0,1]$, $x_2 \sim  N(0.5, 0.1)$; $D_3$: $x_1 \sim  [0,1]$,  $x_2 \sim [0,1]$; $D_4$: $x_1 \sim  N(0.5, 0.05)$, $x_2 \sim  N(0.75, 0.016)$, with -0.8 correlation. Feature contributions are different per rank stratum and data distribution.}
\label{fig:distr-dependence}
\end{figure}

\paragraph*{Fixed scoring function, varying data distribution.} In this experiment, we illustrate that feature importance is impacted by the data distribution of the scoring features to a much greater extent than by the feature weights in the scoring function.  Further, we show that feature importance varies by rank stratum. In Figure~\ref{fig:distr-dependence}, we show rank QoI for 4 synthetic datasets with the same scoring function $f_2$.   

We observe that, while the features have equal scoring function weights, their contributions to rank QoI differ for most datasets. In $D_1$, the Bernoulli-distributed $x_2$ determines whether the item is in the top or the bottom half of the ranking, while the Gaussian-distributed $x_1$ is responsible for the ranking inside each half. For $D_2$, the uniform $x_1$ has higher importance because it often takes on larger values than the Gaussian $x_2$. In $D_4$, $x_1$ and $x_2$ are negatively correlated, so when one contributes positively, the other contributes negatively. Only for $D_3$, with two uniform identically distributed features, the median contributions of both features are approximately the same within each stratum.  

Additionally, we see that feature contributions differ per rank stratum. For example, for $D_3$, the medians show a downward trajectory across strata. 
This is because they quantify the expected change (positive or negative) in the number of rank positions to which the current feature values contribute.  Also for $D_3$, feature contributions have higher variance in the middle of the range, because a 40-60\% rank corresponds to many feature value combinations.

\begin{figure}
    \centering
    \includegraphics[width=0.9\columnwidth]{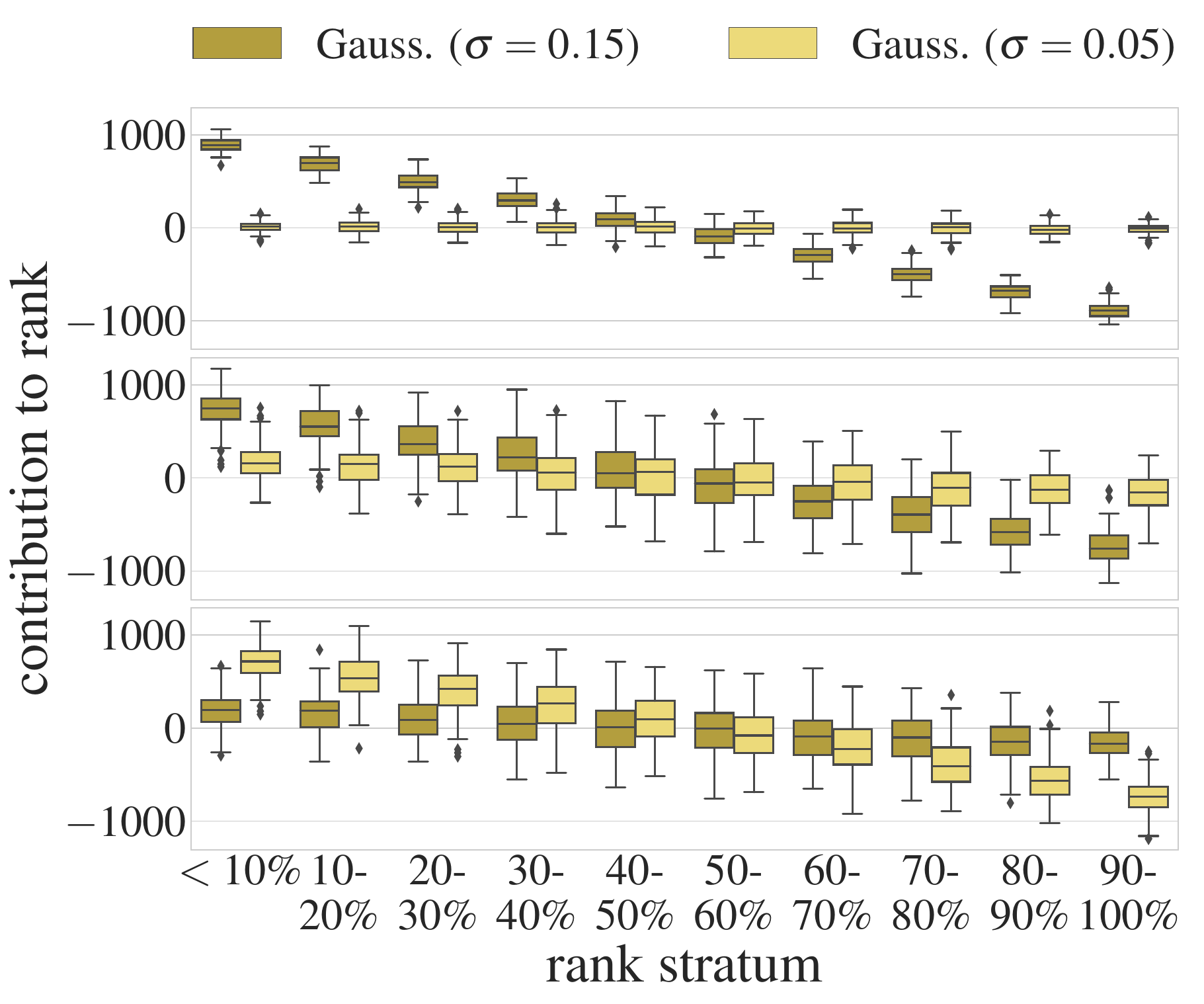}
    \caption{Rank QoI for $D_5$: $x_1 \sim  N(0.5, 0.1)$, $x_2 \sim  N(0.5, 0.05)$. Subplots correspond to different scoring functions: $f_1 = 0.8 \times x_1 + 0.2 \times x_2$ (top), $f_2 = 0.5 \times x_1 + 0.5 \times x_2$ (middle), $f_3 = 0.2 \times x_1 + 0.8 \times x_2$ (bottom).}
    \label{img:df3-rank}
\end{figure}

\paragraph*{Fixed data distribution, varying scoring function.} In this experiment, we investigate the impact of the scoring function on rank and top-$k$ QoI for two datasets.  In Figure \ref{img:df3-rank}, we use $D_3$  and see that the contributions to rank QoI vary depending on the scoring function. For $f_1$, $x_1$ is the only important feature (although it carries 0.8 --- and not 1.0 --- of the weight).  This can be explained by the compounding effect of the higher scoring function weight and the higher variance of the distribution from which $x_1$ is drawn. Between $f_2$ and $f_3$, features $x_1$ and $x_2$ switch positions in terms of importance, and show a similar trend, despite being associated with different scoring function weights (0.5 \& 0.5 vs. 0.2 \& 0.8).  This, again, can be explained by the higher variance of $x_1$, hence, $x_2$ needs a higher scoring function weight to compensate for lower variance and achieve similar importance.

\begin{figure}
    \centering
    \includegraphics[width=0.9\columnwidth]{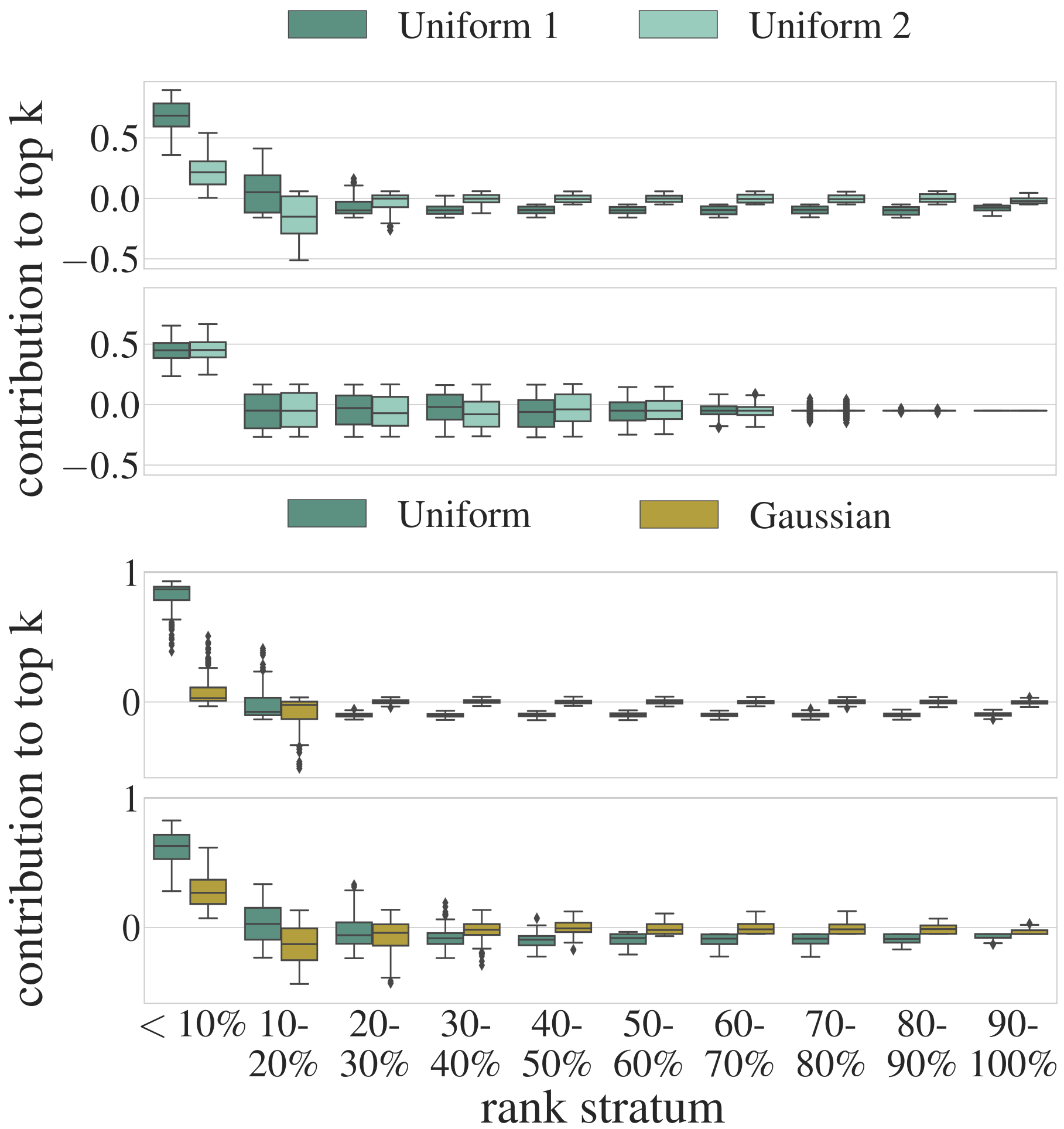}
    \caption{Top-$k$ QoI for $k=10\%$, $D_2$: $x_1 \sim  [0,1]$, $x_2 \sim  N(0.5, 0.1)$, and $D_3$: $x_1 \sim  [0,1]$,  $x_2 \sim [0,1]$. Subplots correspond to different scoring functions: $f_1 = 0.8 \times x_1 + 0.2 \times x_2$ (top), $f_2 = 0.5 \times x_1 + 0.5 \times x_2$ (bottom).}
    \label{img:df1-topK}
    \end{figure}

\paragraph*{Top-$k$ access.} Access to the top-$k$ is determined by the interaction between the scoring feature weights and the distributions of these features.
The top-k QoI tells us how important each feature is when we consider only access to the top-$k$. A positive feature contribution signifies that changing the feature's value will result in decreased chances of getting to the top-$k$. A very high (or very low) value shows that the changes are significant. Figure~\ref{img:df1-topK} illustrates this for datasets $D_2$ and $D_3$. When we consider two identical uniform features that have equal weights ($D_2$ under $f_2$), we first notice that their control of top-$k$ access is identical, as expected. Additionally, we see that for the top-10, changing either feature would reduce access to the top-k (the values are both very positive). However, for each stratum up to the top-70\%, changing either feature can contribute either positively or negatively.

When we consider two identical uniform features ($D_1$), one of which has a higher weight ($f_1$) or dataset $D_2$ (under either $f_1$ or $f_2$), we see again that for the top-10, changing either feature would reduce access to the top-k. Also, we see that how high the contributions are depends on the distributions. However, we see that for the top 10\%-20\%, changing the second, less important feature would increase the chances of getting into the top-$k$. For the rest of the strata, with some variations depending on the dataset and function, changing the most important feature provides a non-zero probability of moving to the top-$k$, and interestingly, this persists even for the lower strata. Evidence that items from lower strata can move to the top-$k$ under some scoring functions and feature distributions counters the assumption of~\citet{Anahideh2022} that changes in rank are localized.

\begin{figure*}[t!]
    \centering
        \subfloat[Contribution to rank]{
        \includegraphics[width=0.8\textwidth]{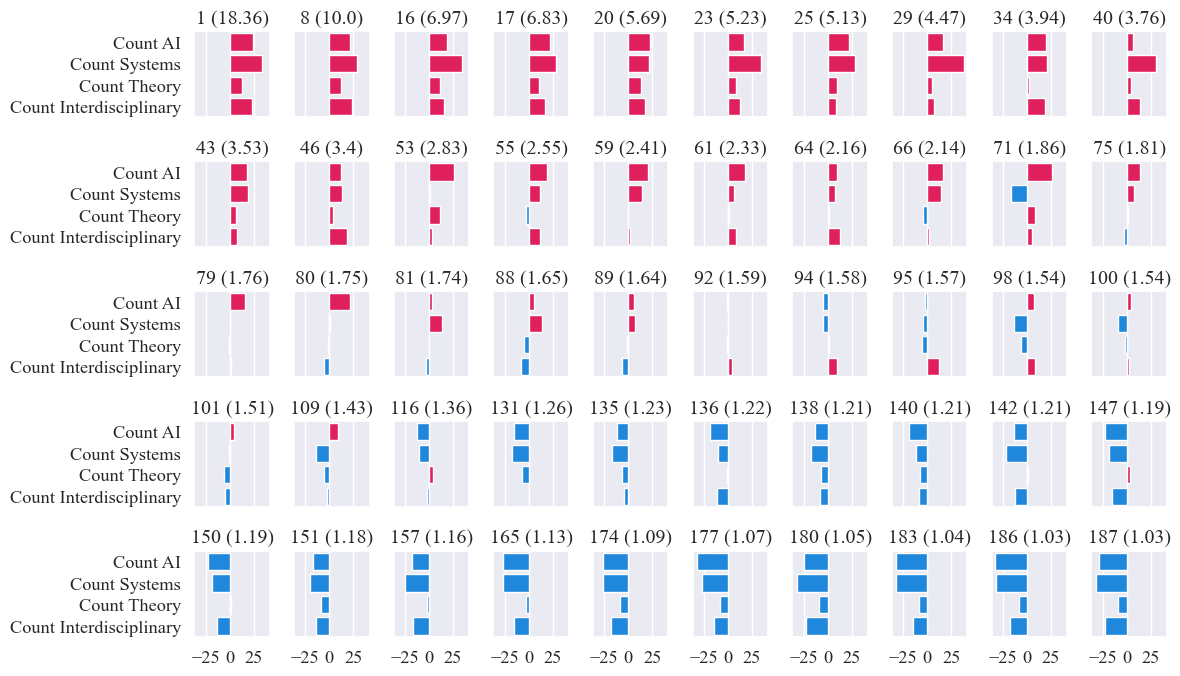}
            \label{img:rank-local-many}}
        \hfill
        \subfloat[Contribution to score]{
        \includegraphics[width=0.8\textwidth]{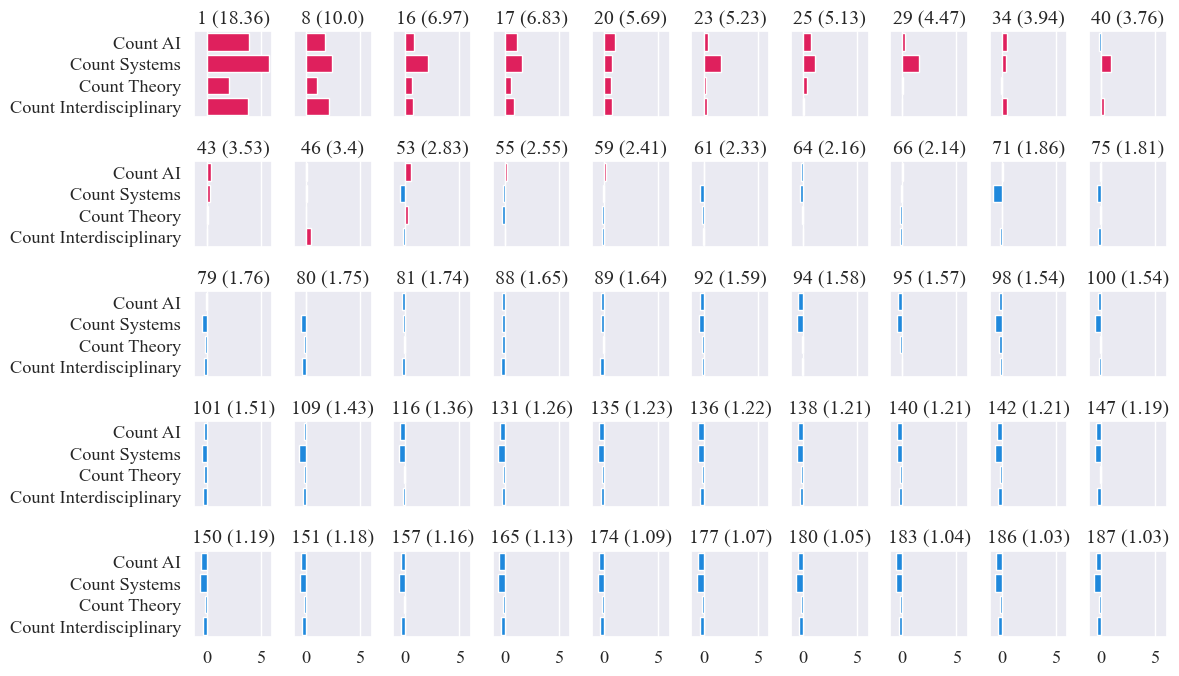}
            \label{img:score-local-many}}
      \caption{Shapley value explanations for fifty random universities for the rank QoI and the score QoI. The title of each subplot indicates the rank of each university and also contains its score in parentheses. The exponentially decreasing score-to-rank relationship and the dependence of Shapley values on the mean make score explanations indistinguishable and negative for most of the items.}
    \label{fig:score-vs-rank-local-many}
\end{figure*}
\section{Rank-QoI-based and Score-QoI-based explanations for \csr}
\label{sec:app:rank-vs-score}
In Section~\ref{sec:qoi} we discuss the differences between the rank QoI and the score QoI for the \csr dataset. In this section, we provide additional details for this comparison. Specifically, we demonstrate that considering different outcomes as profit functions has a profound impact on the explanations for the entire range of the ranking.

In Figure~\ref{fig:score-vs-rank-local-many}, we provide local Shapley value explanations for fifty universities from the CSR dataset for both the rank (Fig.~\ref{img:rank-local-many}) and the score QoI (Fig.~\ref{img:score-local-many}). These universities are randomly chosen; they are approximately 25\% of the dataset and span the entire ranking. Each subplot in each subfigure shows one explanation for one university, and its title shows each university's rank and score (the score is in parentheses). The universities are the same across both subfigures.

Looking at this collection of explanations, we can see how the rank and the score QoI behave significantly differently. Matching what we showed in Figure~\ref{fig:CSRanking}, the score QoI explanations become indistinguishable as we move down the ranking. Additionally, the contributions of all features become negative around rank 61 for the score QoI as opposed to 131 for the rank QoI. Finally, for the score QoI, the contributions are very small for almost all universities, as opposed to the rank QoI, where the contributions are small for the middle of the ranks.

There are two main reasons why the behavior between the rank and the score QoI based explanations is so different. The first is that the \textit{score-to-rank relationship is exponentially decreasing for this dataset} (see Fig.\ref{img:CSR-rankvsscore}). This means that the top of the ranking has very high scores, and the scores quickly reach a plateau. The second is that \textit{Shapley values explain the contribution of each feature to the distance of the outcome from the mean outcome}.
Indeed, the mean score for this dataset is 2.72, and its range is 18.36-1.03, while the mean rank is 95, and its range is 1-198. Together, these two facts mean that for the score QoI, for most items, the distance between its score and the mean score is very small. Because the score-based explanation explains the difference from the mean score, and those differences are very small for most items, the contributions are \textit{very low} for most items. Additionally, the mean score is very influenced by the outliers at the top, so most items have \textit{negative} contributions for all their features, even when ranked in the top 30\% (e.g., the university ranked at position 61).

As discussed in multiple works, for instance ~\cite{molnar2020interpretable,bhatt_2021_evaluating} explanations should differ when the outcomes and the items are different. In these figures, we can see that this is not the case for the score QoI based explanations. Items ranked in the middle of the ranking (e.g., item ranked in position 92) have similar explanations to items ranked at the bottom of the ranking (e.g., item ranked in position 183).

This behavior of the local explanations, coupled with the fact that the score QoI is not able to know when the rank changes (see Section~\ref{sec:intro}), argues for using the rank QoI when explaining rankings.

\section{Implementation of HIL}
\label{sec:app:hil}
HIL ~\cite{DBLP:conf/hilda/YuanD23} is the only other method that recommends the usage of ranks as a profit function for individual explanations in ranking. While this method is not general, we are interested in comparing it with our rank QoI. This was not straightforward because the method is available as a web app that works only for linear weight scoring functions and datasets of two Gaussian features. To compare the rank-relevance contributions introduced in that paper to the rank QoI, we adapted their method using their definitions and code. This implementation is available alongside our own.  
Further, we extended their method to work with the specific non-linear scoring function used by CS Ranking, by changing the way that Std rank and Std score (discussed below) are computed.

More specifically, because HIL ~\cite{DBLP:conf/hilda/YuanD23} works only with linear weight scoring functions, they do not provide a full Shapley values implementation but use the linear weights to approximate Shapley values assuming feature independence, see Corollary 1 in~\cite{lundberg2017unified} and also~\cite{DBLP:journals/kais/StrumbeljK14}. This is a well-established method to compute Shapley values for linear weights, also implemented by SHAP, so we do not compare with this part of the method. In addition, HIL defines two methods to acquire feature contributions: ``standardized Shapley values'' and ``rank relevance Shapley values,'' which we will call Std score and Std rank, respectively. Those are not calculated using the linear weight method described above, but rather directly from the weights, and without using the mean score or rank. For an item $\textbf{v}$, each feature $i$ contribution for Std score is $\phi_i = \frac{\beta_i \textbf{v}_i}{\sum_{\textbf{u}\in \mathcal{D}}f(u)}$, where $\beta_i$ is the weight for feature $i$. In other words, the contribution of each feature for each item is the score contribution of this feature over the sum of all scores for all items. Similarly, for Std rank, the contribution of feature $i$ for an item $\textbf{v}$ is $\phi_i =\beta_i \textbf{v}_i \alpha_{\textbf{v}}$, where $\alpha_{\textbf{v}}$ is a scaling factor used to transform the score of the specific item to the rank of the specific item calculated as $\alpha_{\textbf{v}} = \frac{(\max_{r \in r_{\mathcal{D}}}(r) - r^{-1}_{\mathcal{D}(\textbf{v})})\sum_{\textbf{u}\in \mathcal{D}}f(u)}{\max_{r \in r_{\mathcal{D}}}(r) f(\textbf{v})}$. Note that neither of the two formulas is computing Shapley values; rather, they assign a contribution to the features based on the linear weights and the score and rank. This implies that our rank QoI is the only rank QoI for Shapley values.

\section{Additional Details on Method Comparisons}
\label{sec:app:comparisons}
\subsection{Fidelity}
We provide more details on the Fidelity results discussed in Section~\ref{sec:exp:fidelity}. We compute the Fidelity of all the methods that have that property across all datasets. We use SHAP and LIME out of the box, so their performance is not perfect (although extremely good). We make this choice to highlight the importance of using exact Shapley values when computing local explanations, where the error in each separate explanation is important, as each explanation impacts a separate person.

\begin{table}[h!]
\caption{Fidelity across all methods across all datasets.}
\begin{tabular}{ c c c c c c c}
 & LIME  & SHAP & \multicolumn{2}{ c }{ShaRP} & HIL\\
 dataset  &  score & score  & score & rank & score \\
 \midrule
 ATP &  0.98 &  1.00 & 1.00 & 1.00 & 0.14 \\
 CSR &  0.95 &  0.99 & 1.00 & 1.00 & 0.85 \\
 THE &  0.94 &  0.97 & 1.00 & 1.00 & 0.64 \\
 Syn 0 & 0.95 & 0.99	& 1.00 & 1.00 & 0.37 \\
 Syn 1 & 0.95 & 0.99 & 1.00 & 1.00 & 0.29 \\
 Syn 2 & 0.95 & 0.99 & 1.00 & 1.00 & 0.35 \\
 \bottomrule
\end{tabular}
\label{table:fidelity}
\end{table}

\begin{figure}
    \centering
    \includegraphics[width=0.65\columnwidth]
        {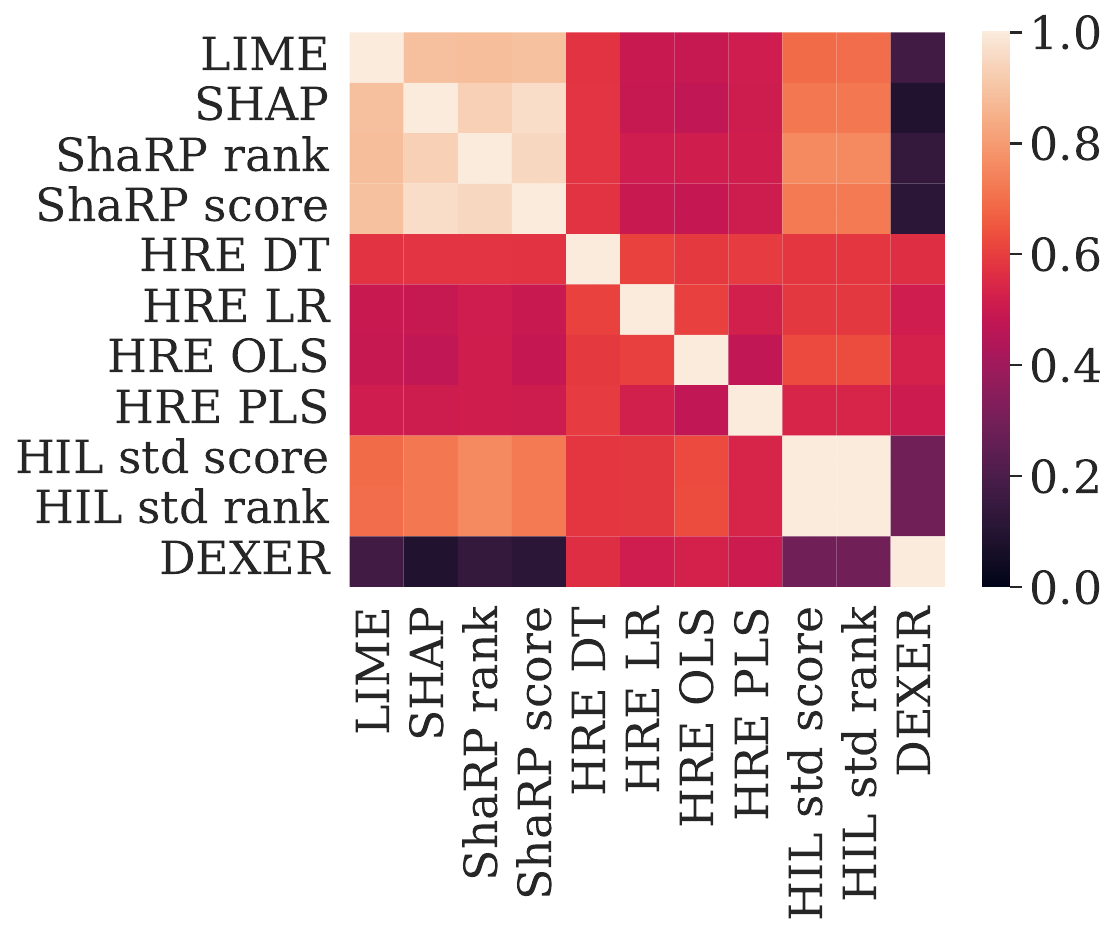}
    \caption{Method agreement averaged across all datasets}
    \label{fig:agreement-sharp}
\end{figure}

\begin{figure}
    \centering
        \subfloat[Method Agreement between Sharp using the rank QoI and all methods across all datasets]{
        \includegraphics[width=0.65\columnwidth]
        {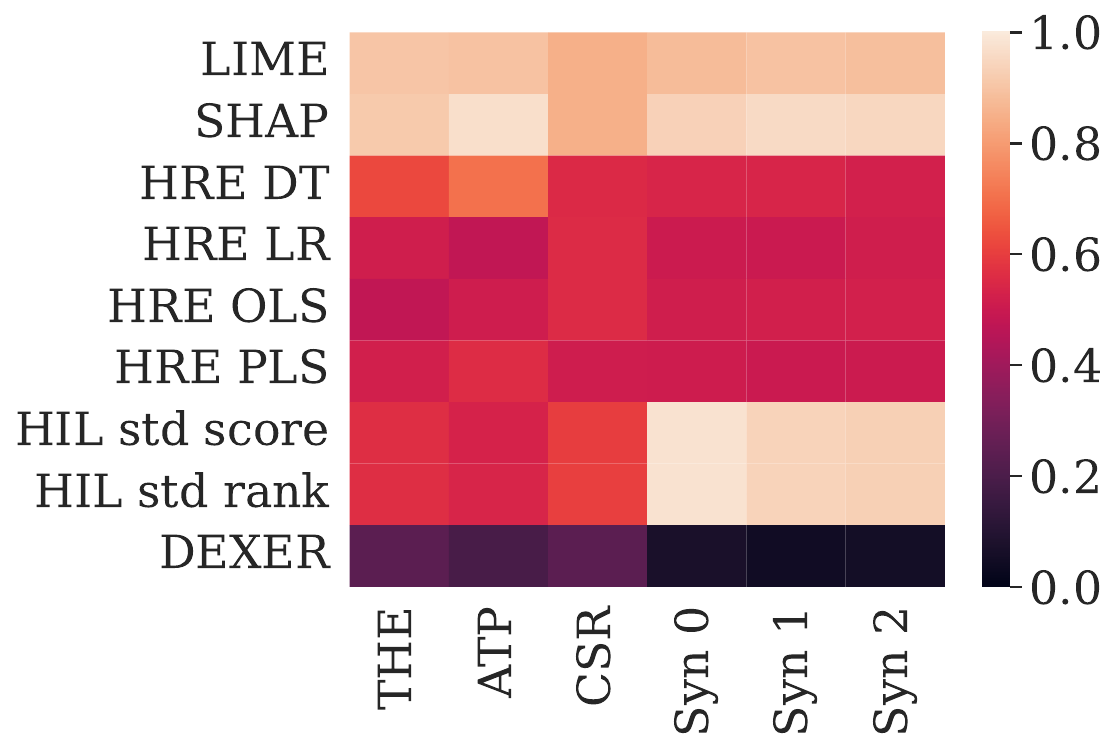}
            \label{img:rank-heatmap}}
        \hfill
        \subfloat[Method Agreement between Sharp using the score QoI and all methods across all datasets]{
		\includegraphics[width=0.65\columnwidth]
        {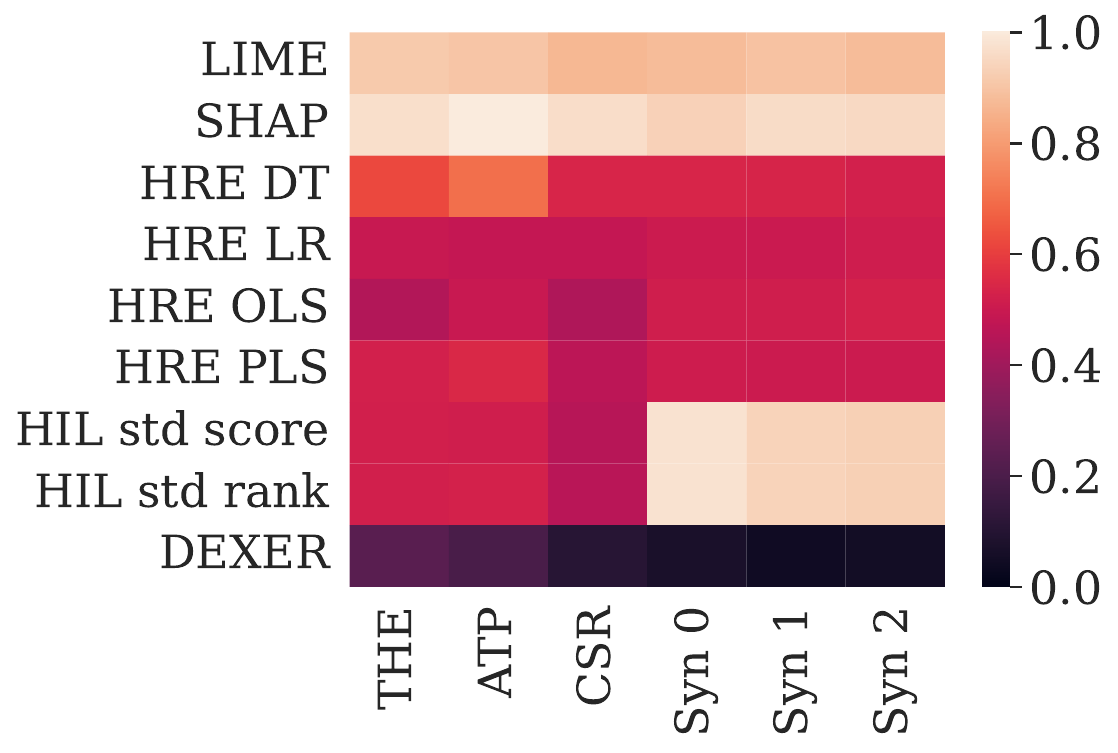}
		\label{img:score-heatmap}
    }
    \caption{Method Agreement}
    \label{fig:agreement-sharp-2}
\end{figure}

\subsection{Agreement between Explanations}
\label{app:exp:agreement}

Figure~\ref{fig:agreement-sharp} presents agreement between~\sys and all other methods averaged across all datasets. We use rank and score QoIs for this comparison, as they match those used by the methods we evaluate. Kendall's tau distance is computed to enable cross-method comparisons. We observe that explanations vary significantly by method, regardless of the QoI. ShaRP aligns most closely with LIME and SHAP across both rank and score QoIs. HRE, which relies on localized information, naturally differs. However, even among HRE variants, explanations remain inconsistent.  The two HIL methods and the two ShaRP methods produce similar explanations despite using different QoIs, suggesting that explanation consistency depends more on the method than the QoI. In contrast, DEXER, which fits a linear regression to the ranking output and applies SHAP, differs greatly from all methods, indicating that rank cannot be effectively explained without a rank QoI.

Figure~\ref{fig:agreement-sharp-2} provides a per-dataset visualization of the agreement between the explanations of the methods in Section~\ref{sec:exp:other}.

In Fig. \ref{img:rank-heatmap}, we visualize Kendall's tau explanation distance correlation of ShaRP using the rank QoI with all other methods across every dataset. In Fig.~\ref{img:score-heatmap}, we plot the same result for ShaRP using the score QoI. As in the aggregated method agreement plot (Fig.~\ref{fig:agreement-sharp}), ShaRP is very similar to SHAP and LIME for both QoIs. As expected, it is more similar to the score QoI but not identical, which is perhaps because we used SHAP out of the box, which applies some approximation parameters for running time optimization. Similarly, ShaRP behaves similarly to what we discussed in Fig.~\ref{fig:agreement-sharp} to all methods across the datasets except the HIL methods for the Synthetic datasets. We hypothesize that this is because the HIL methods are able to perform better for those datasets due to their Synthetic nature.

\subsection{Sensitivity}
\label{sec:app:sensitivity}
\begin{figure}
    \centering
        \subfloat[HRE DT]{
        \includegraphics[width=0.5\columnwidth]
        {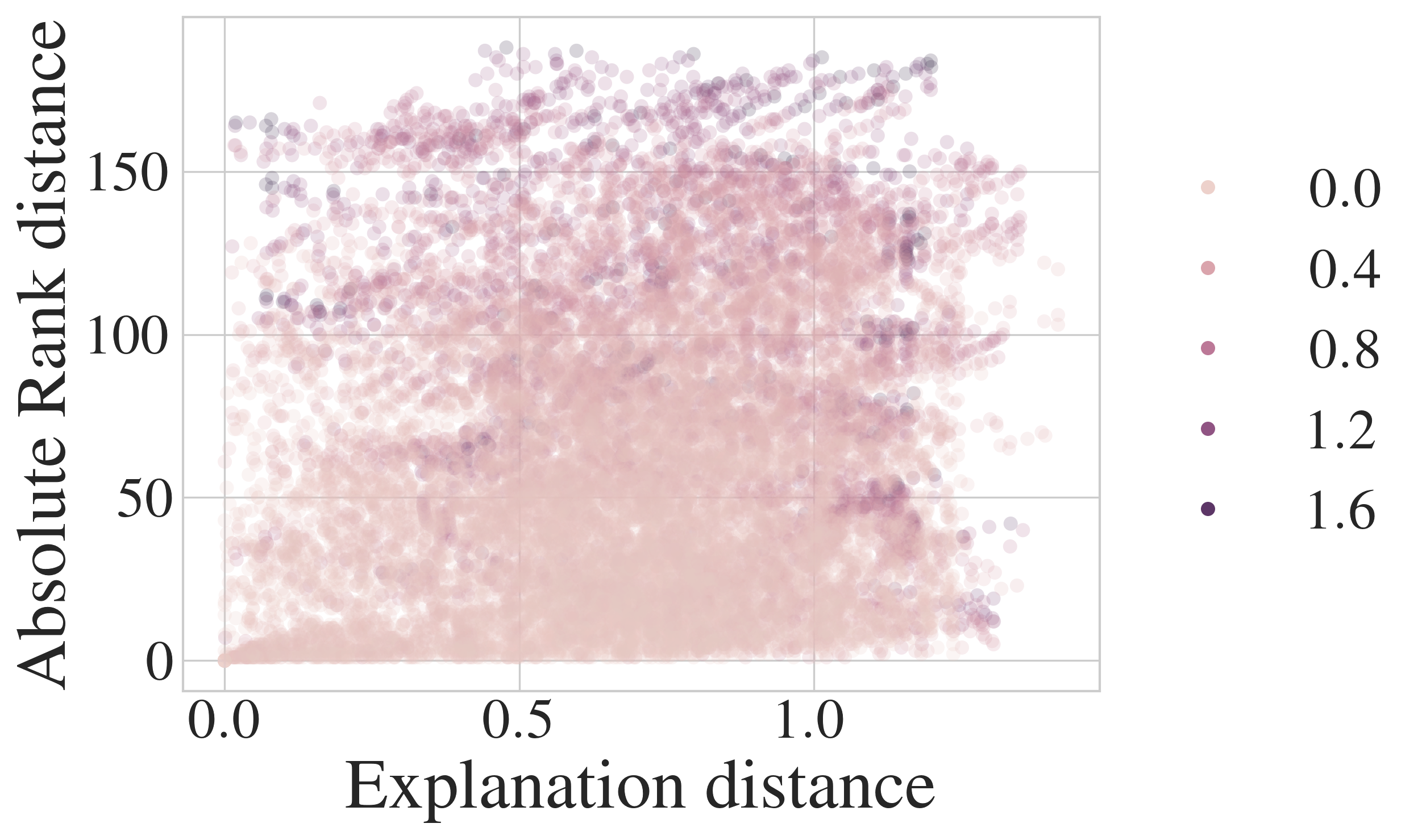}
            \label{img:HRE-DT-CSRank}}
        \subfloat[HRE LR]{
        \includegraphics[width=0.5\columnwidth]
        {images/sensitivity-scatterplot-euclidean-CSRank-HRE_LR.png}
            \label{img:HRE-LR-CSRank}}
        \hfill
        \subfloat[HRE OLS]{
        \includegraphics[width=0.5\columnwidth]
        {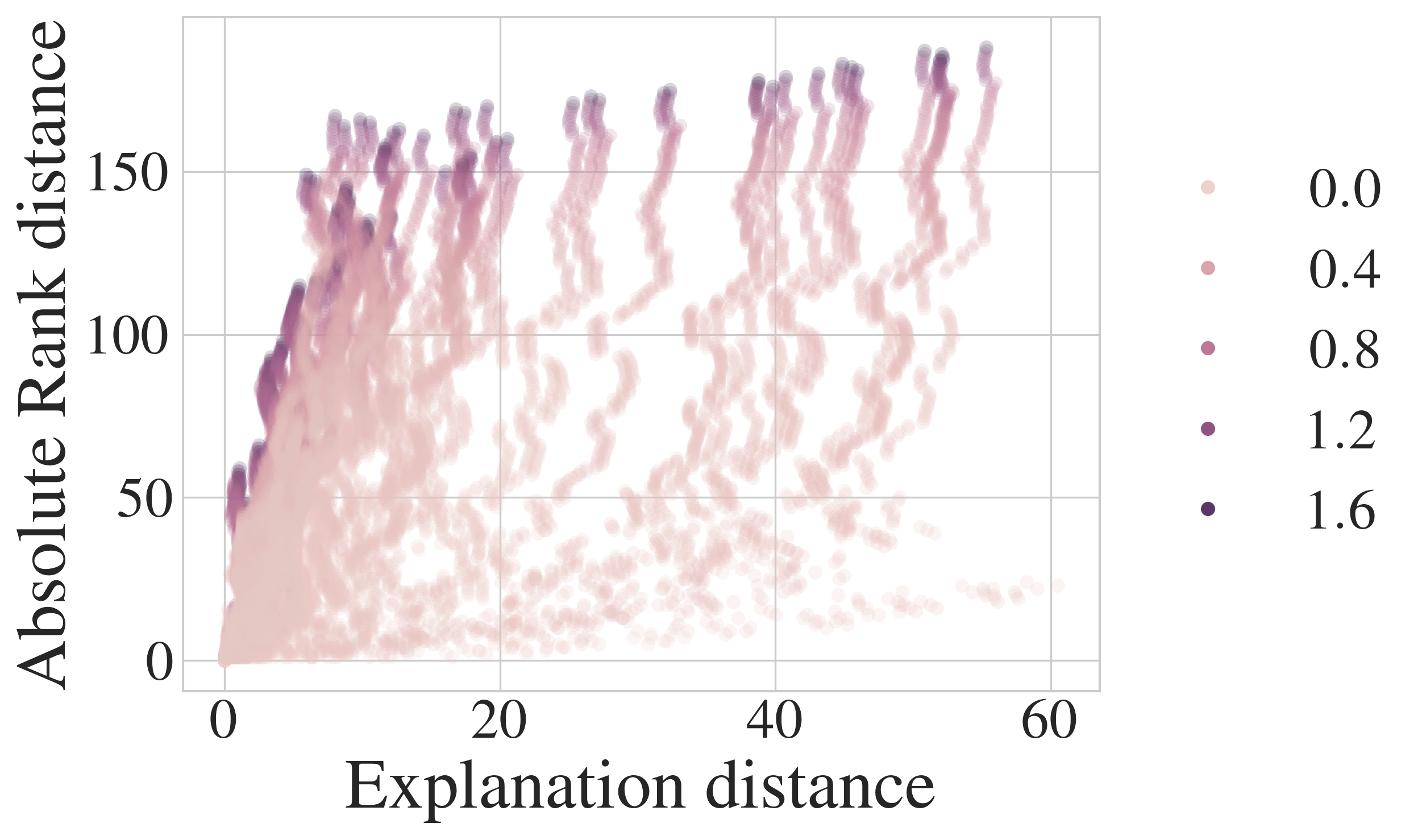}
            \label{img:HRE-OLS-CSRank}}
        \subfloat[HRE PLS]{
        \includegraphics[width=0.5\columnwidth]
        {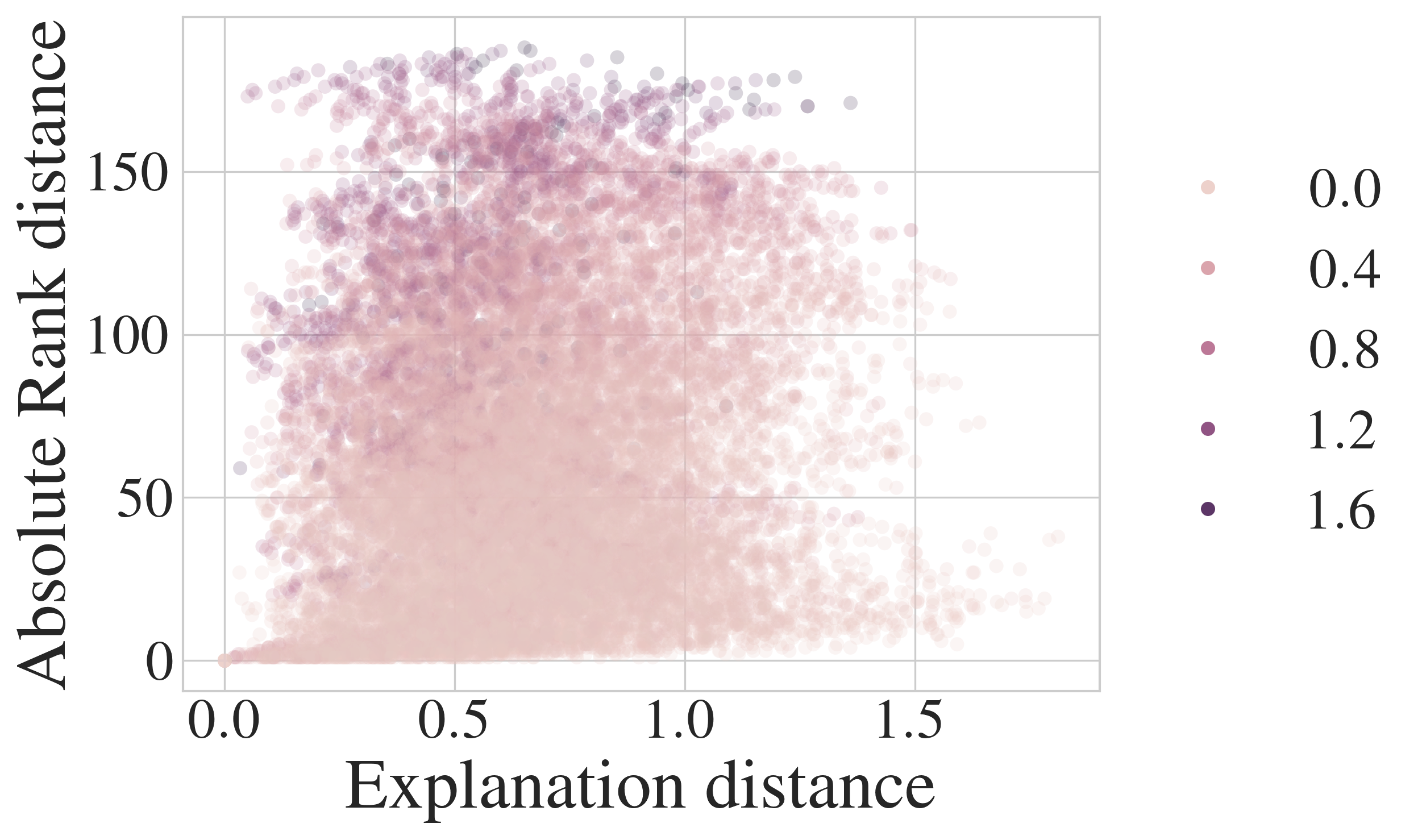}
            \label{img:HRE-PLS-CSRank}}
        \hfill
        \subfloat[LIME]{
		\includegraphics[width=0.5\columnwidth]
        {images/sensitivity-scatterplot-euclidean-CSRank-LIME.png}
		\label{img:LIME-CSRank}}
        \subfloat[SHAP]{
		\includegraphics[width=0.5\columnwidth]
        {images/sensitivity-scatterplot-euclidean-CSRank-SHAP.png}
		\label{img:SHAP-CSRank}}
        \hfill
        \subfloat[ShaRP Score]{
		\includegraphics[width=0.5\columnwidth]
        {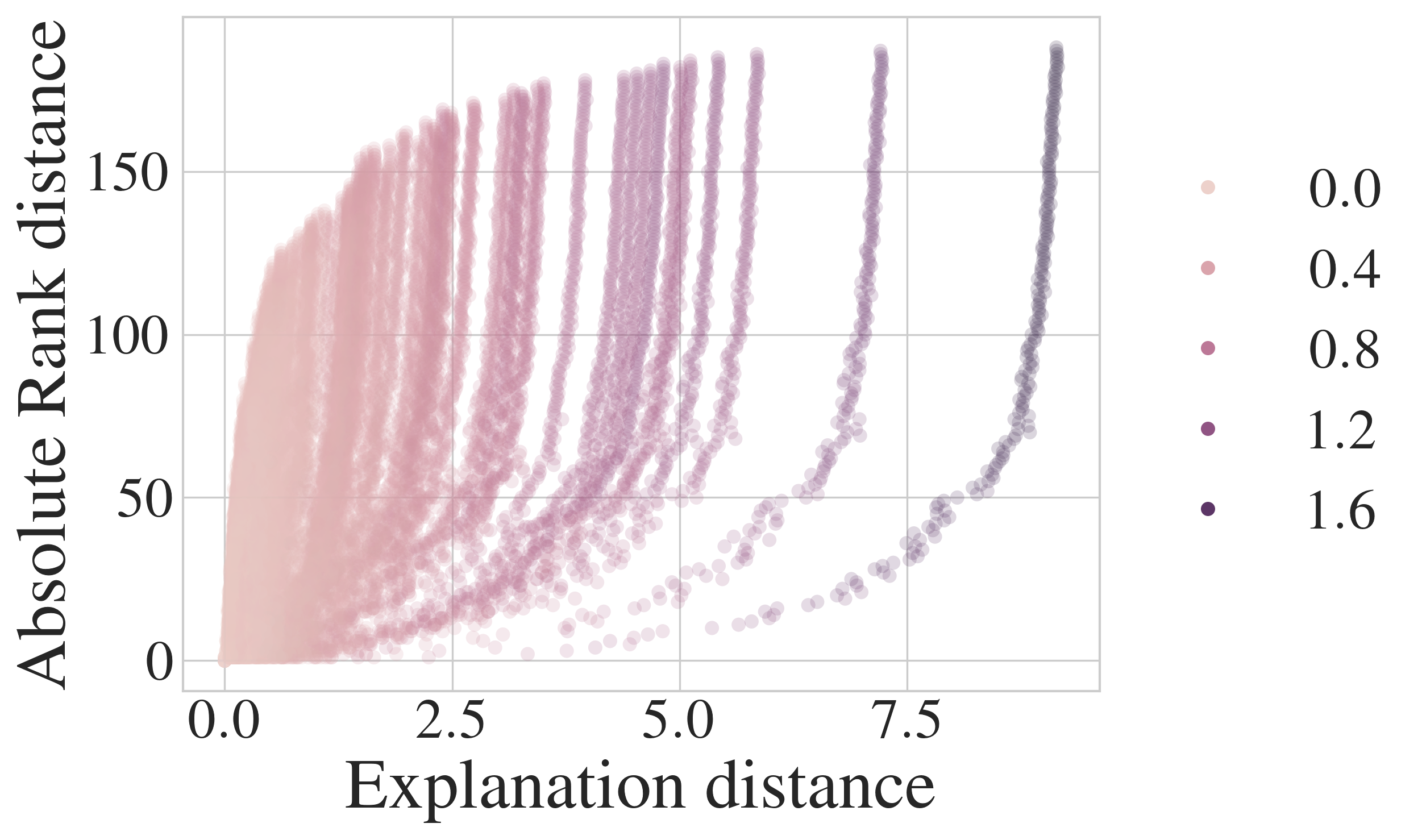}
		\label{img:ShaRP_SCORE-CSRank}}
        \subfloat[ShaRP Rank]{
	    \includegraphics[width=0.5\columnwidth]
        {images/sensitivity-scatterplot-euclidean-CSRank-ShaRP_RANK.png}
		\label{img:ShaRP_RANK-CSRank}}
        \hfill
        \subfloat[HIL Std Score]{
        \includegraphics[width=0.5\columnwidth]
        {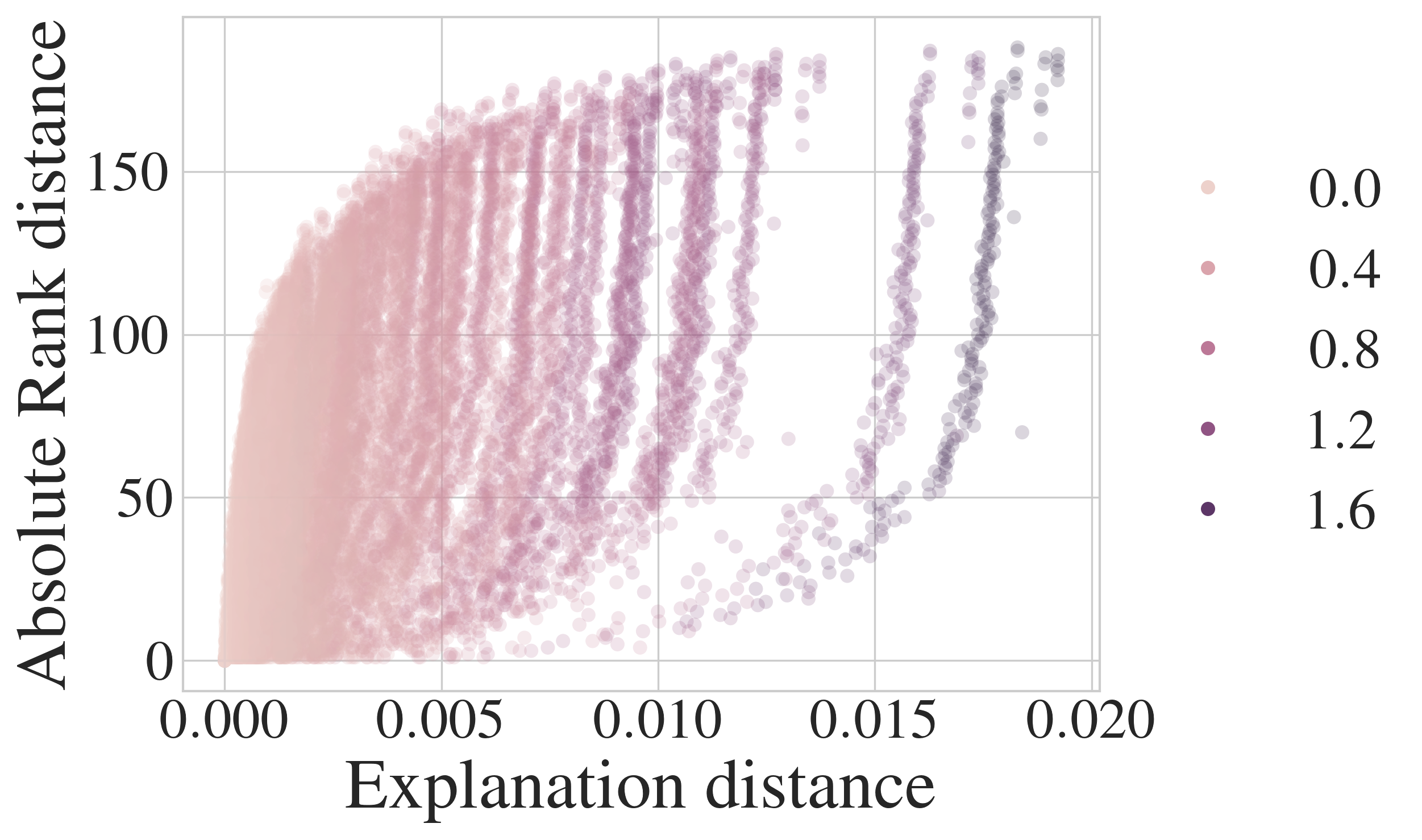}
            \label{img:HIL-Score-CSRank}}
        \subfloat[HIL Std Rank]{
        \includegraphics[width=0.5\columnwidth]
        {images/sensitivity-scatterplot-euclidean-CSRank-HIL_Rank-Shapley.png}
            \label{img:HIL-Rank-CSRank}}
        \hfill
        \subfloat[DEXER]{
        \includegraphics[width=0.5\columnwidth]
        {images/sensitivity-scatterplot-euclidean-CSRank-DEXER.png}
            \label{img:DEXER-CSRank}}
    \caption{Comparison of the sensitivity metric results for the \csr dataset for all methods.}
    \label{fig:sensitivity-CSRank}
\end{figure}

We provide the results of the sensitivity metric for all methods for \csr in Fig.~\ref{fig:sensitivity-CSRank}. In addition to HRE LR, DEXER, LIME, SHAP, ShaRP rank, and HIL std rank that we presented in Fig.~\ref{fig:sensitivity-CSRank-partial}, we also plot HRE DT in Fig.~\ref{img:HRE-DT-CSRank}, HRE OLS in Fig.~\ref{img:HRE-OLS-CSRank}, HRE PLS in Fig.~\ref{img:HRE-PLS-CSRank}, ShaRP score in Fig.~\ref{img:ShaRP_SCORE-CSRank}, and HIL std score in Fig.~\ref{img:HIL-Score-CSRank}. We see that all HRE methods perform similarly or worse than HRE-LR. This is unsurprising as all these methods are used locally. We also see that both HIL std score and ShaRP score perform similarly to SHAP, which is also expected. HIL std score and DEXER are very similar, which reveals our inability to predict the rank using the ranked output of the model. Specifically, the HIL std score assumes knowledge of the weights used by the model and uses them directly to compute the feature importance. DEXER is assuming black-box access to the ranked output only and fits a linear regression model on the ranking. Nevertheless, judging from these results, it appears that DEXER is explaining the score (and not the rank) and is learning the model weights to do so. Inadvertently, we also show that the choice of the explanation method makes a big difference to the final explanation.

\begin{figure}
    \centering
        \subfloat[ShaRP Rank ATP]{
	\includegraphics[width=0.5\columnwidth]
        {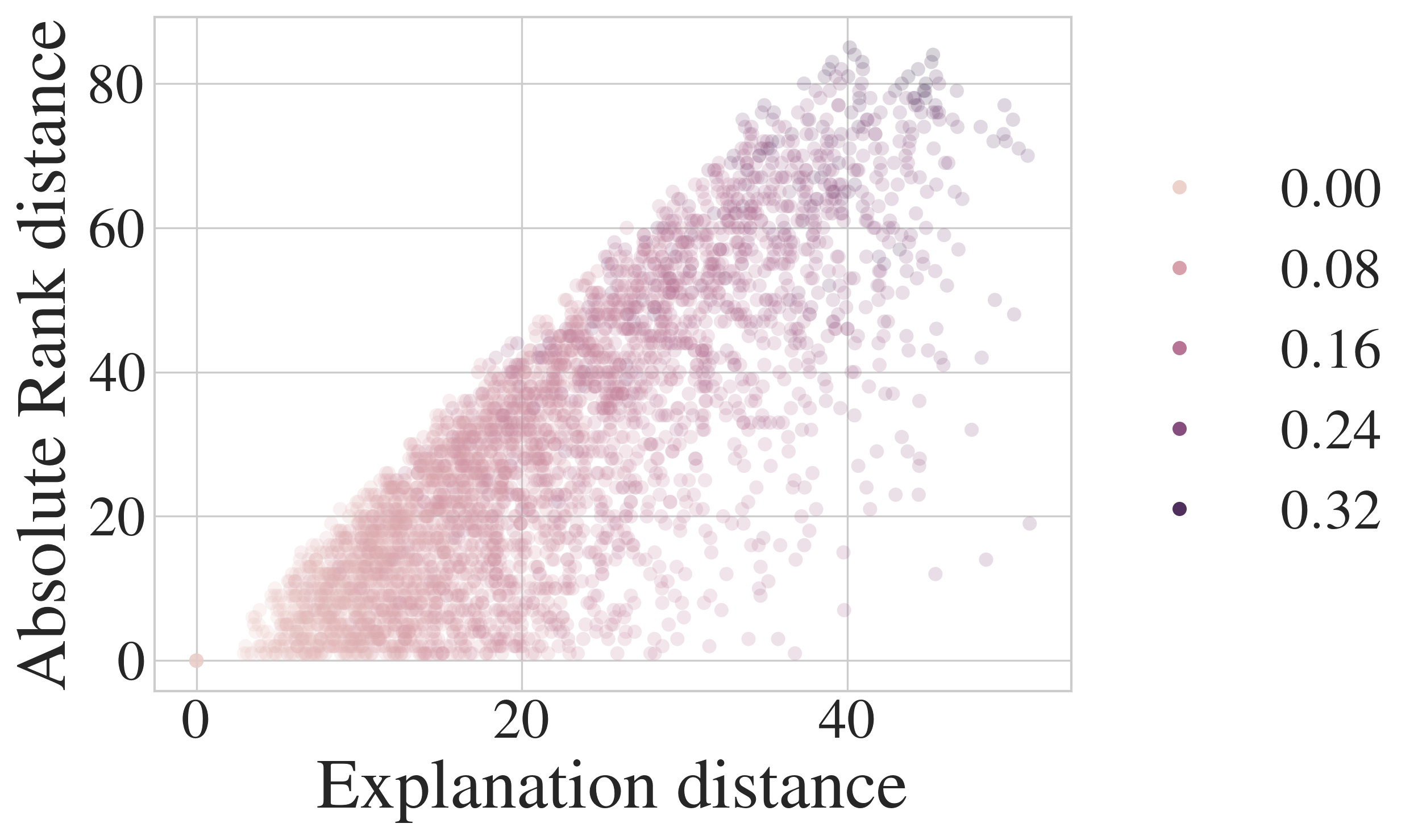}
		\label{img:ShaRP_RANK-ATP-2}}
        \subfloat[HIL Std Rank ATP]{
        \includegraphics[width=0.5\columnwidth]
        {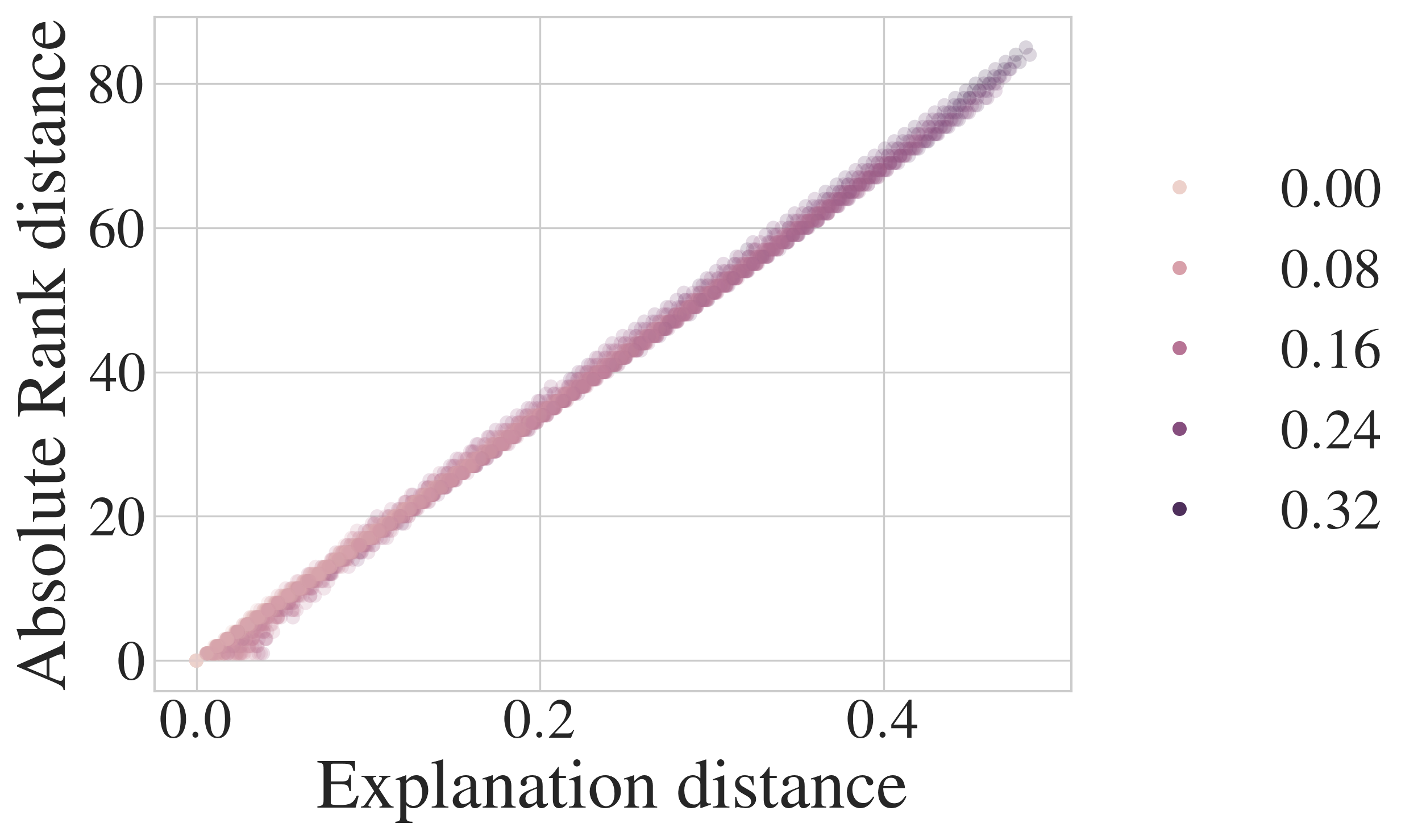}
            \label{img:HIL-Rank-ATP-2}}
        \hfill
        \subfloat[ShaRP Rank THE]{
	\includegraphics[width=0.5\columnwidth]
        {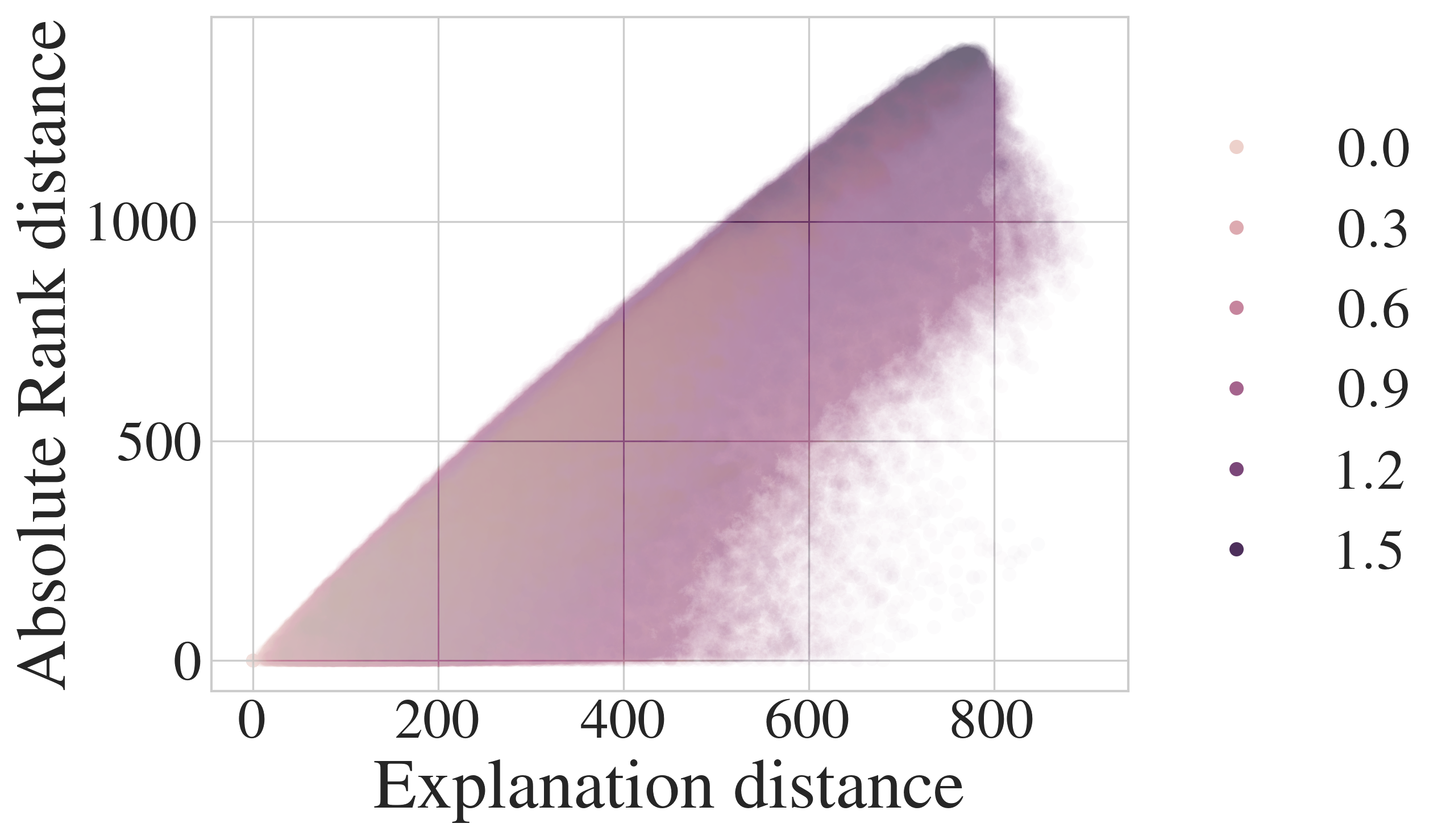}
		\label{img:ShaRP_RANK-THE-2}}
        \subfloat[HIL Std Rank THE]{
        \includegraphics[width=0.5\columnwidth]
        {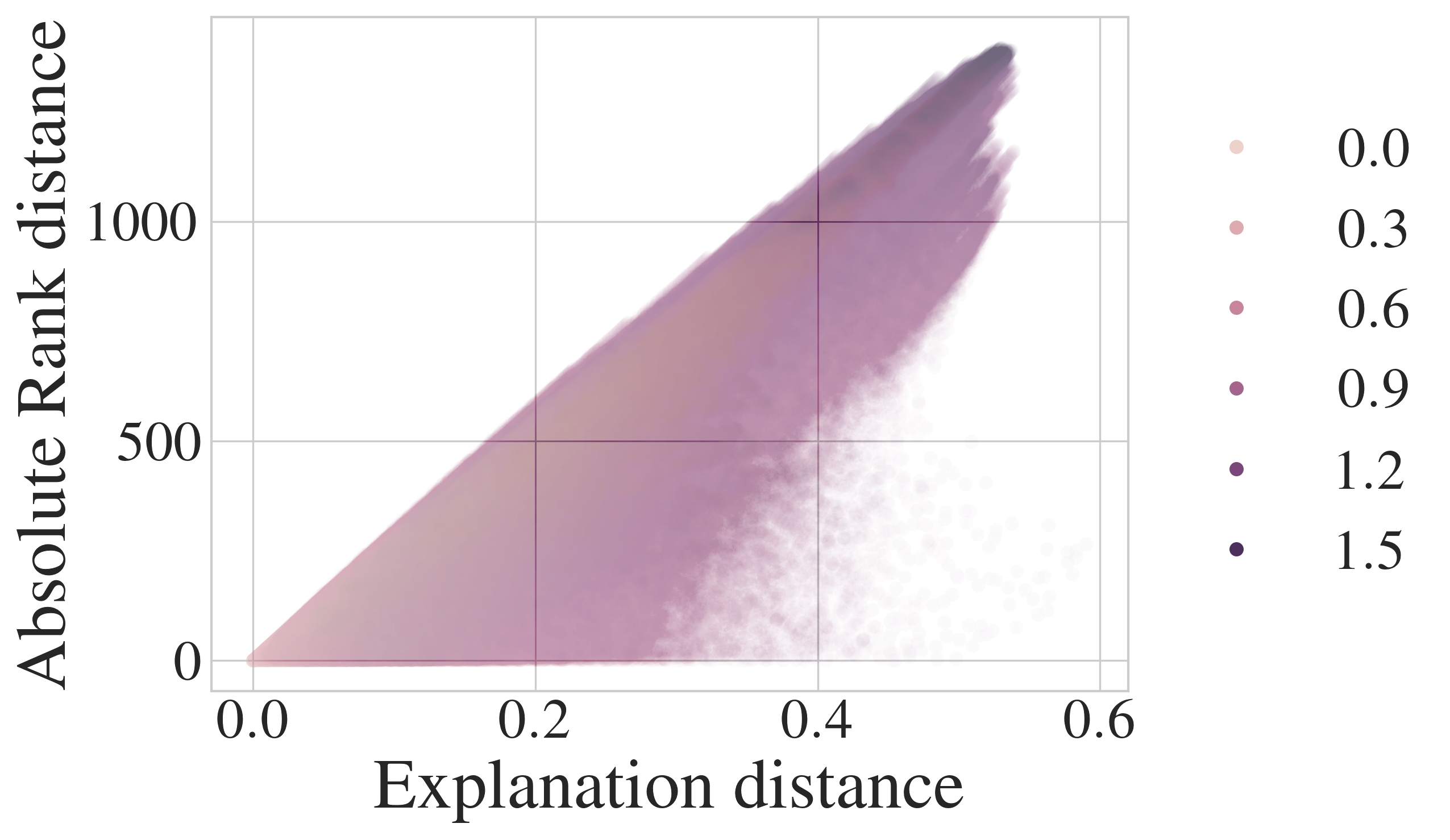}
            \label{img:HIL-Rank-THE-2}}
        \hfill
        \subfloat[ShaRP Rank Synthetic-0]{
	\includegraphics[width=0.5\columnwidth]
        {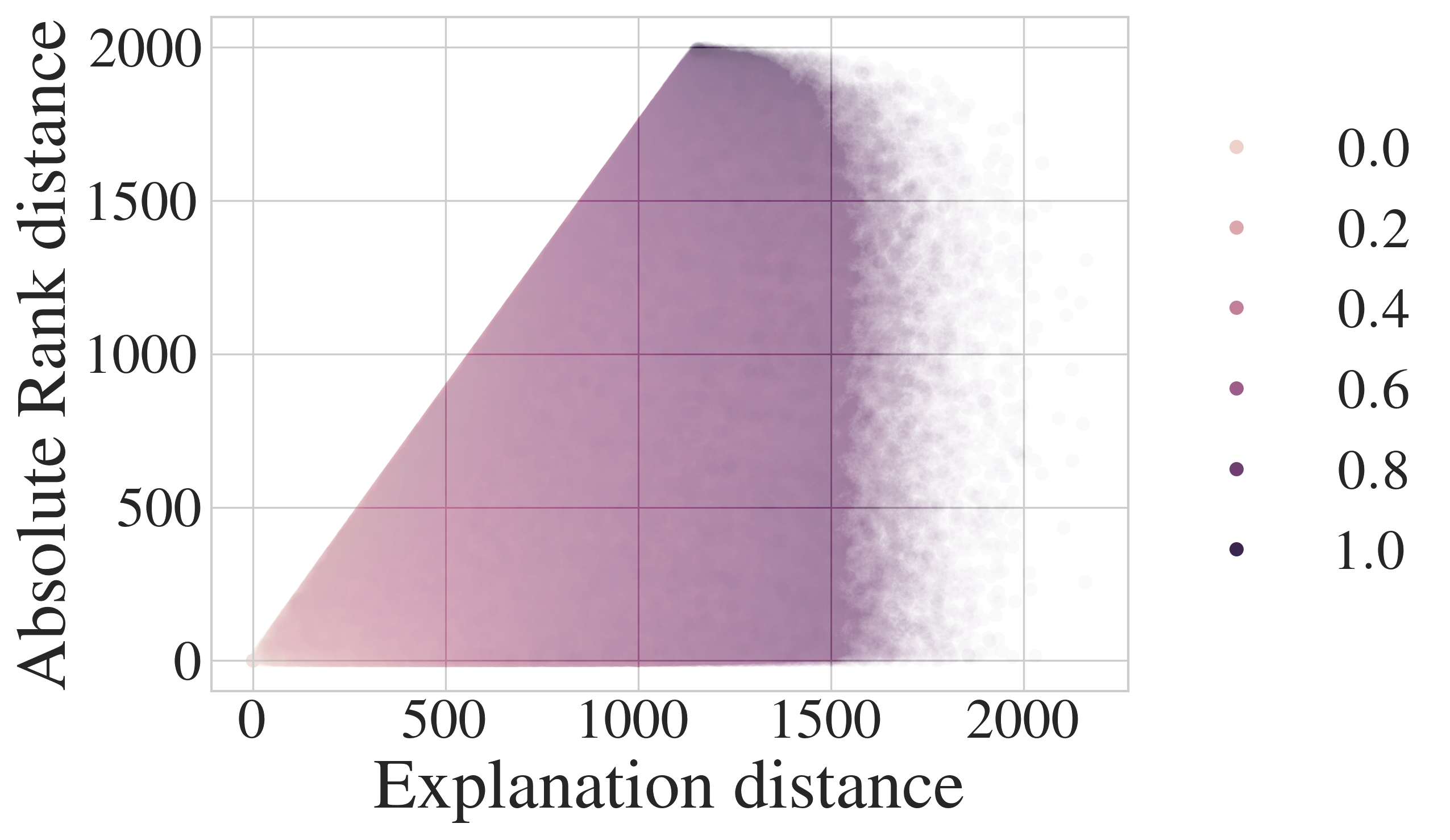}
		\label{img:ShaRP_RANK-Syn0-2}}
        \subfloat[HIL Std Rank Synthetic-0]{
        \includegraphics[width=0.5\columnwidth]
        {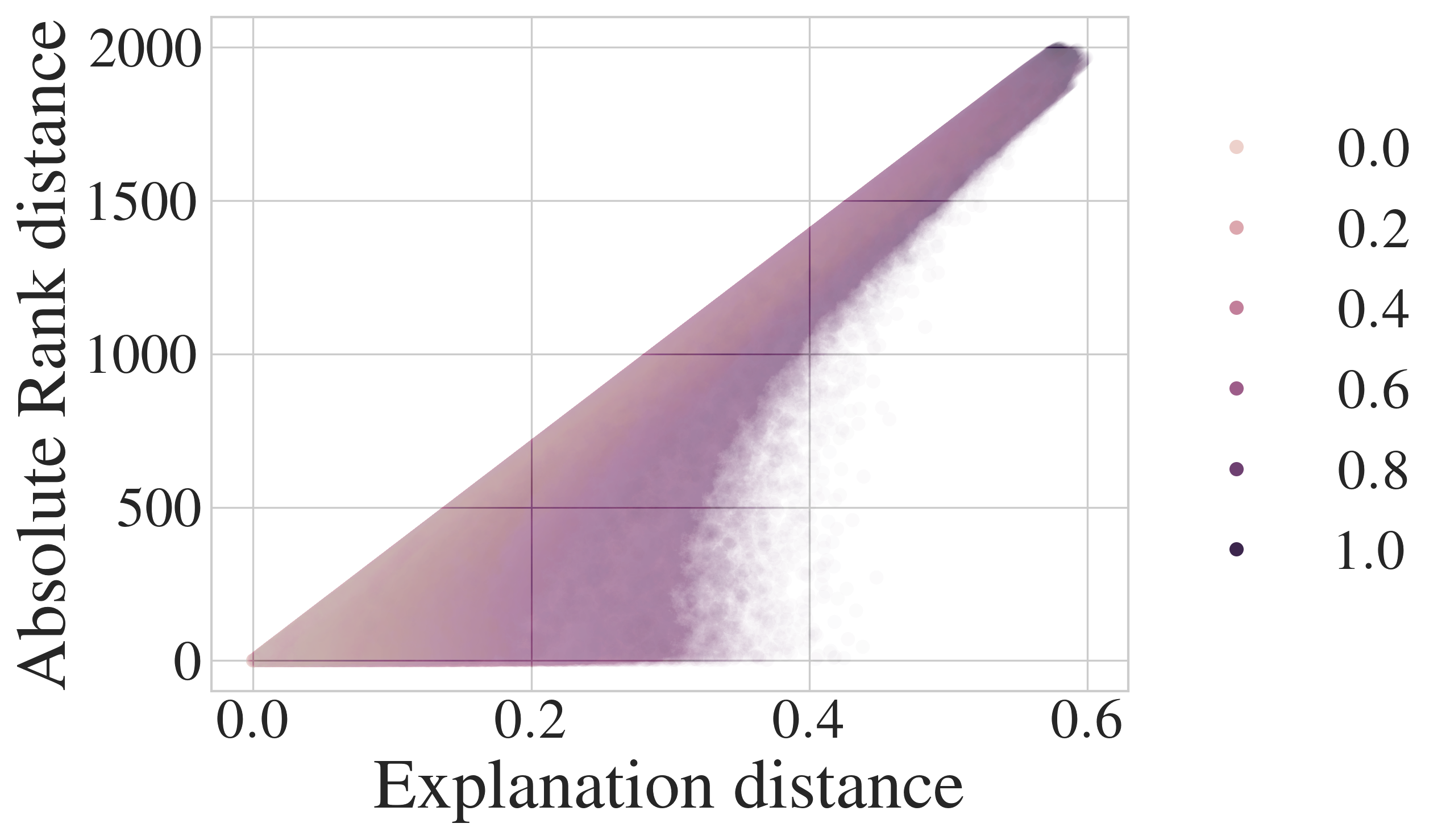}
            \label{img:HIL-Rank-Syn0-2}}
        
    \caption{Comparison of the sensitivity metric results for the ATP, THE, and Synthetic dataset 0 for the methods using the rank QoI.}
    \label{fig:sensitivity-rank-only}
\end{figure}

To further compare~\sys with rank QoI and HIL with Std rank, we present Figure~\ref{fig:sensitivity-rank-only}. Even though both methods are appropriate for the ranking task we are examining, in this figure, we see that \sys with rank QoI (Figures~\ref{img:ShaRP_RANK-ATP-2}, ~\ref{img:ShaRP_RANK-THE-2}, and ~\ref{img:ShaRP_RANK-Syn0-2}) can capture the full range of different ranks and features, and that groups the items more successfully. HIL with Std rank cannot capture the difference of feature values for ATP (Figure~\ref{img:HIL-Rank-ATP-2}), or the similarly ranked items that have different feature values for the Synthetic experiment (the middle area close to the $x$-axis of Figure~\ref{img:HIL-Rank-Syn0-2}. Both methods perform similarly for THE (Figures~\ref{img:ShaRP_RANK-THE-2} and ~\ref{img:HIL-Rank-THE-2}).

Finally, we present an analysis of ShaRP using the score QoI and the rank QoI for the CSR dataset, but for a score task (instead of rank). The goal of this analysis is to show that the sensitivity of the methods that use a score QoI is very high when we are explaining a score task. In other words, if we are trying to explain the score, then the methods that use a score-based profit function perform the best, as is fully expected.

The task we are going to explain is the score of the \csr scoring function. We choose this task for two reasons: first, we already provided the results of the \csr ranking task, and we can draw a direct comparison. Secondly, we have a ranking for that dataset, and we can plot the methods that use the rank QoI for juxtaposition. Note that it is entirely redundant to use a rank-based QoI method in this case. In fact, it is redundant to even produce a ranking as we are asking an explainability question about the score. But we are choosing to provide this information to showcase that each explainability task needs its own profit function, and the choice of the profit function makes a big difference to the final explanation.

\begin{figure}
    \centering
        \subfloat[ShaRP Score]{
		\includegraphics[width=0.5\columnwidth]
        {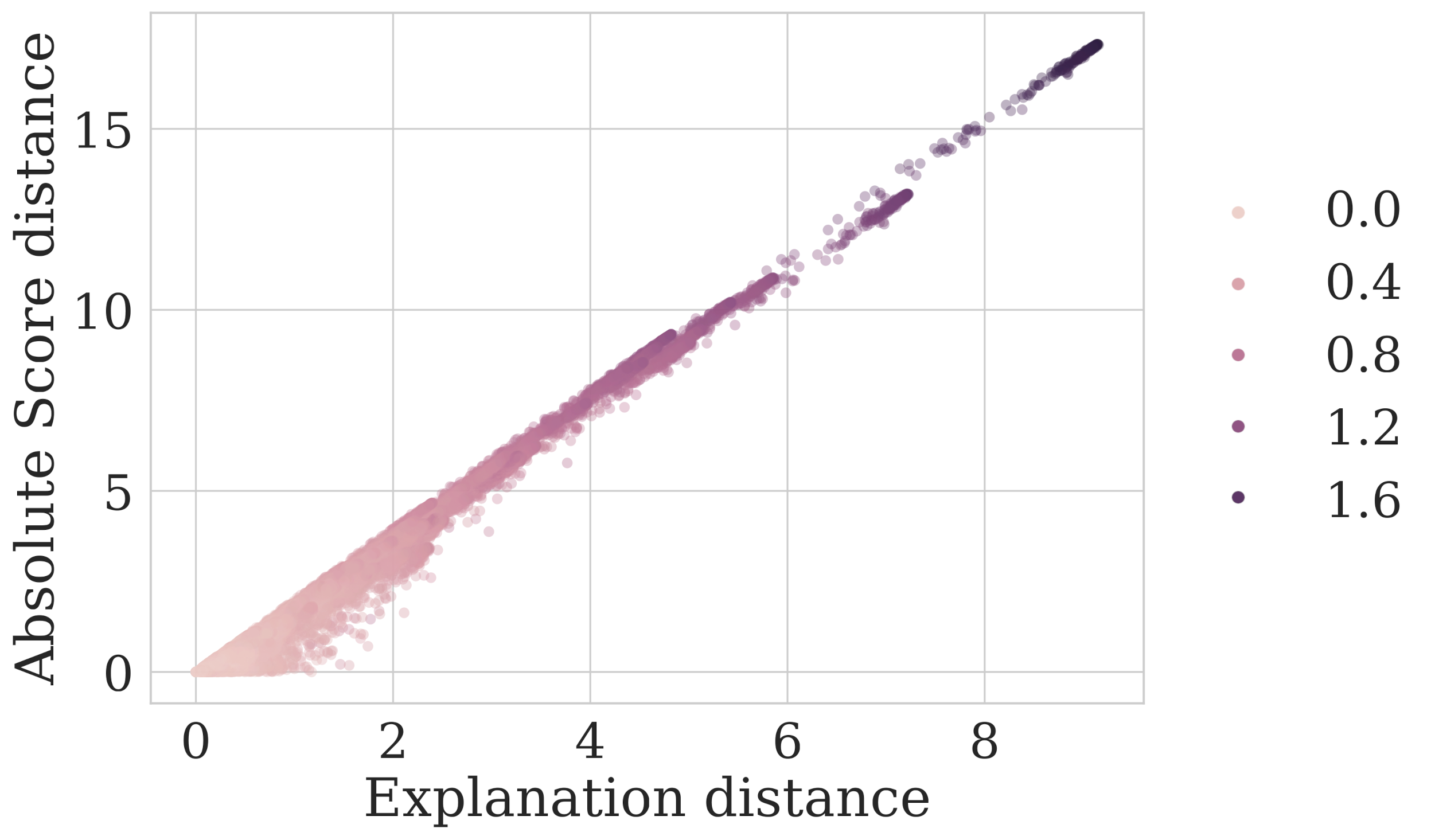}
		\label{img:ShaRP_SCORE-CSRank-scores}}
        \subfloat[ShaRP Rank]{
	\includegraphics[width=0.5\columnwidth]
        {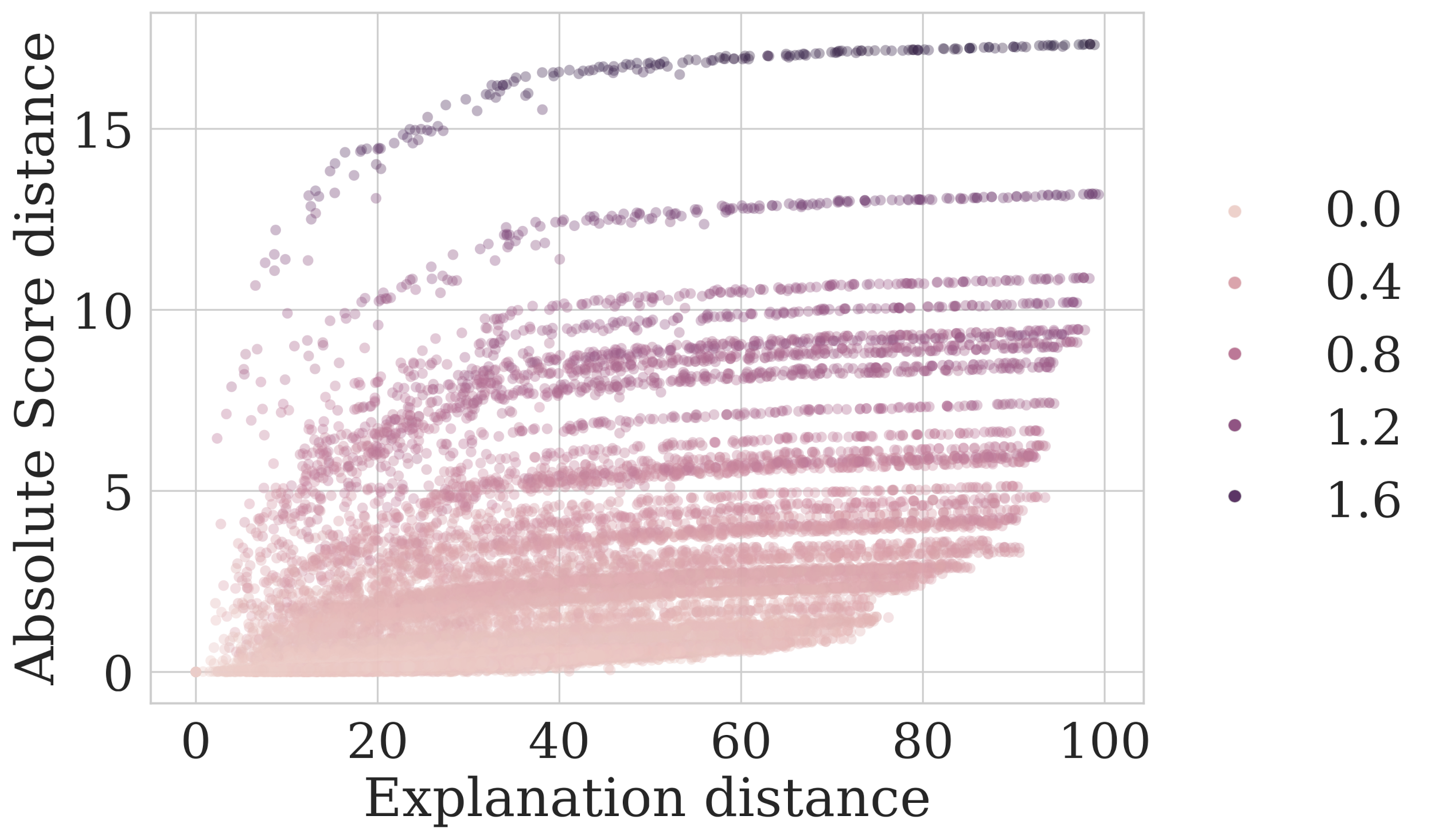}
		\label{img:ShaRP_RANK-CSRank-scores}}
    \caption{Comparison of the sensitivity metric results for the \csr dataset for all methods when the task we are trying to explain is a score task. Compared to Figure~\ref{fig:sensitivity-CSRank}, we see that the methods that use a score QoI are performing better.}
    \label{fig:sensitivity_scores}
\end{figure}

In Fig.~\ref{fig:sensitivity_scores}, we evaluate the similarity of explanations for pairs of similar items \textit{when we attempt to explain the score}. For each pair of items, we compute three distances: (1)  Euclidean distance between the explanations (x-axis); (2)  distance between the \textit{scores} (instead of rank) of the two items (y-axis); and (3) Euclidean distance between the items in terms of their feature values (hue, where lighter means closer). To make the plot, we place one item (the reference item) at position (0,0) and use a scatter point for each other item (neighbor), indicating the distance in ranks and the distance of the explanations. The color of the scatter point indicates the distance between the features of the reference point and the neighbor. We then overlay the plots for all items in the dataset, so that all items are used as reference points.

Unlike Fig.~\ref{fig:sensitivity-CSRank}, we now expect to see items that are both similar in terms of their features and \textit{scored} near each other to have similar explanations. We would still expect all points to be on or near the diagonal line $y=x$, with the hue getting darker as we move away from the origin, \textit{if their explanations successfully explain the score}. 

In Fig.~\ref{fig:sensitivity_scores}, we see that indeed the score-based method has the desired shape we discussed in Section~\ref{sec:exp:sensitivity}. The ShaRP score is extremely similar and almost entirely fits the $y=x$ line. The ShaRP rank appears to be providing explanations that do not depend on the score distance between the items' outcomes (y-axis) or the feature distance between the items (hue), as expected.

\textit{This analysis shows how QoI selection is important when providing an explanation.} The score is unable to perform well for a ranking task since it estimates the impact of each feature on the score outcome, and similarly, it is completely unreasonable to use a rank QoI when explaining the score.

\section{Additional Results for ACSIncome}
\label{app:exp:scored}
\begin{figure*}[h!]
    \centering
        \subfloat[ACS strata]{
        \includegraphics[width=0.72\linewidth]{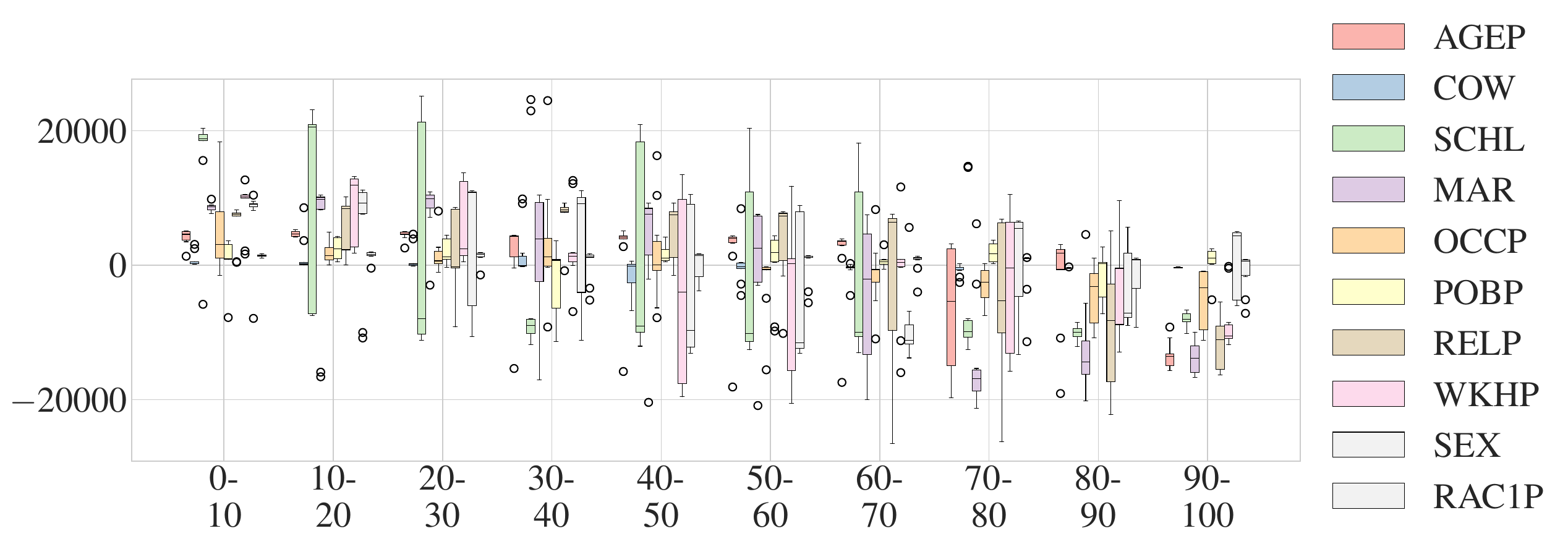}
            \label{img:acs-strata-texas}}
        \hfill
        \subfloat[Overall]{
        \includegraphics[width=0.24\linewidth]{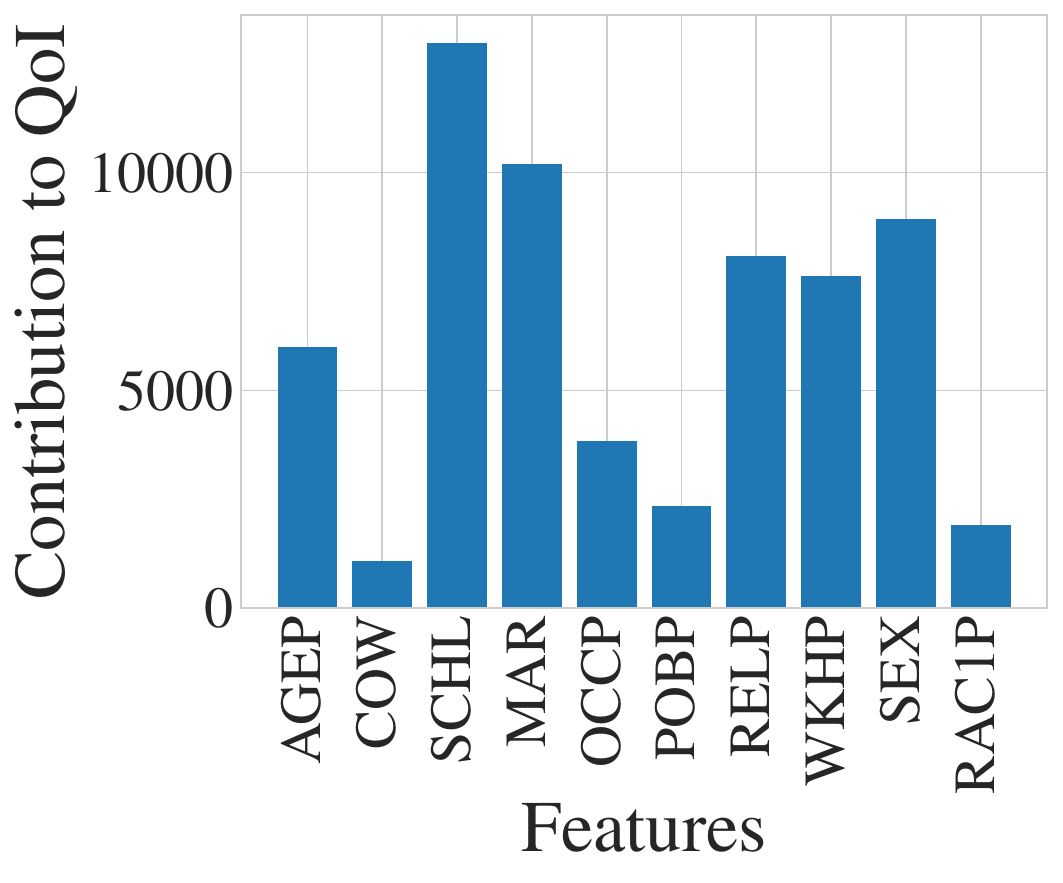}
            \label{img:acs-overall-texas}}
      \caption{Feature contribution on  ACS Income (Texas) to the rank QoI (a) across strata and (b) overall.}
    \label{fig:acs-overall-texas}
\end{figure*}

In Figure~\ref{fig:acs-overall-texas}, we present the overall and strata results for the second ACSIncome dataset we used, Texas, that was previewed in Section~\ref{sec:exp:scored}. As discussed in that section, the feature importance shifts notably compared to Alaska, shown in Figure~\ref{fig:acs-overall-alaska}. The biggest changes are in age (AGEP), education (SCHL), work hours per week (WKHP), and race (RAC1P). These differences highlight the usefulness of explanations, the necessity of working with multiple subsets of similar data, and the ability of our method to capture distributional shifts.

\section{Additional Results on Efficiency and Approximation}
\label{sec:app:approx}
\begin{table*}[h!]
    \caption{Time experiment results over the ACS Income (Alaska) dataset. Running times (reported in seconds) for varying coalition sizes are measured using a fixed sample size of 100, while running times for varying sample sizes are measured using a fixed coalition size of 9. All results are reported by averaging results over 10 tuples, 3 runs each.}
    \begin{tabular}{cr|rrr|rrr}
        \toprule
         &  & \multicolumn{3}{c}{Rank} & \multicolumn{3}{c}{Score} \\
        \makecell{max coal. \\ size} & sample size & Time (cold) & Time (warm) & Fidelity & Time (cold) & Time (warm) & Fidelity \\
        \midrule
        1 & 100 & 1.98 & 1.87 & 0.810 & 0.37 & 0.17 & 0.850 \\
        3 & 100 & 6.46 & 3.14 & 0.857 & 5.410 & 1.41 & 0.887 \\
        5 & 100 & 18.57 & 5.82 & 0.904 & 17.70 & 4.13 & 0.924 \\
        7 & 100 & 24.51 & 7.06 & 0.951 & 23.34 & 5.41 & 0.961 \\
        9 & 100 & 24.86 & 7.18 & 0.991 & 23.66 & 5.52 & 0.993 \\
        \midrule
        9 &   20 & 45.10 & 12.60 & 0.994 & 43.35 & 11.02 & 0.996 \\
        9 &   50 & 95.27 & 28.79 & 0.995 & 92.87 & 27.58 & 0.996 \\
        9 &  100 & 160.19 & 55.80 & 0.997 & 154.56 & 54.54 & 0.997 \\
        9 &  250 & 292.22 & 137.50 & 0.998 & 282.04 & 136.82 & 0.999 \\
        9 &  500 & 445.00 & 271.93 & 0.999 & 428.96 & 270.44 & 0.999 \\
        9 & 1,000 & 708.37 & 542.77 & 0.999 & 689.55 & 536.26 & 0.999 \\
        9 & 3348 & 1,956.79 & 1,830.24 & 1.000 & 1,960.67 & 1,816.78 & 1.000 \\
        \bottomrule
    \end{tabular}
    \label{tbl:acs-ak-times-full}
\end{table*}
\begin{figure*}[h!]
    \centering
        \subfloat[Speedup vs. sample size, CSR]{
        \includegraphics[width=0.55\columnwidth]{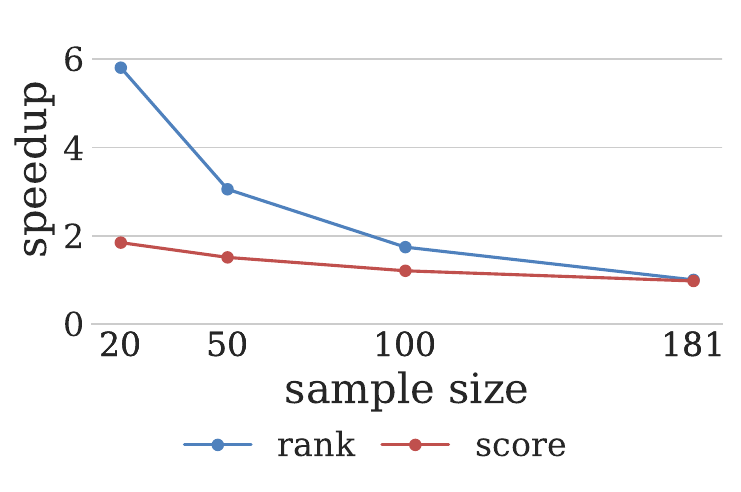}
            \label{img:CSR-sample-speedup-full}}
        \hfill
        \subfloat[Speedup vs. sample size, ATP]{
        \includegraphics[width=0.55\columnwidth]{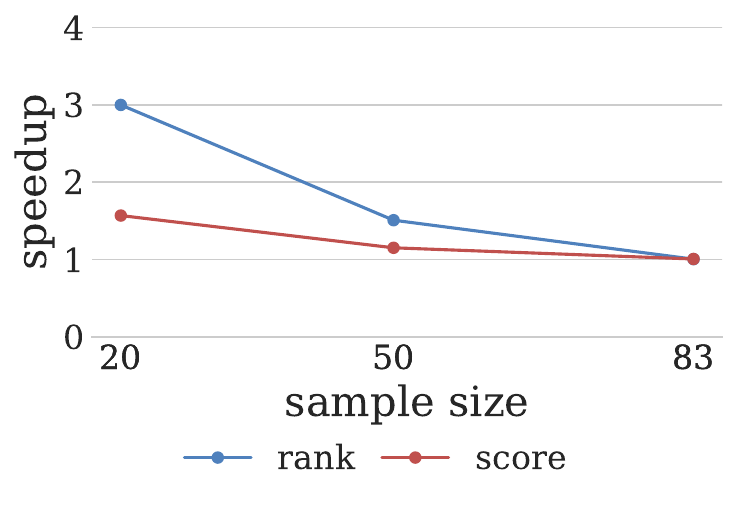}
            \label{img:ATP-sample-speedup-full}}
        \hfill
        \subfloat[Speedup vs. sample size, THE]{
        \includegraphics[width=0.55\columnwidth]{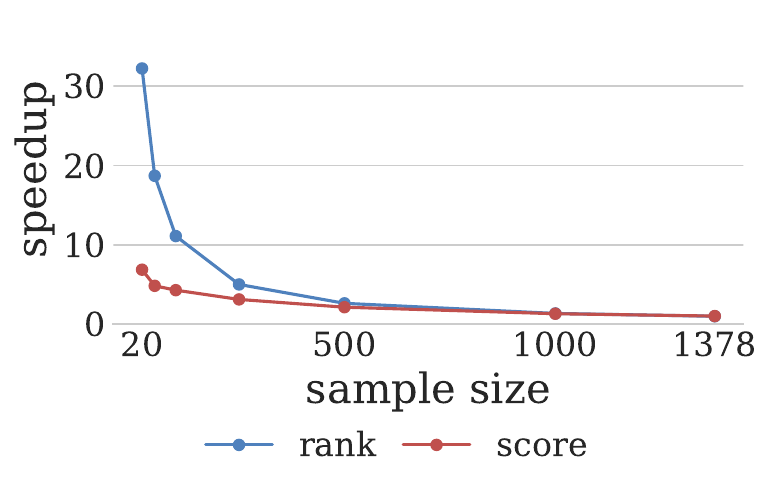}
            \label{img:THE-sample-speedup-full}}
        \hfill
        \subfloat[Speedup vs. maximum coalition size, CSR]{
        \includegraphics[width=0.55\columnwidth]{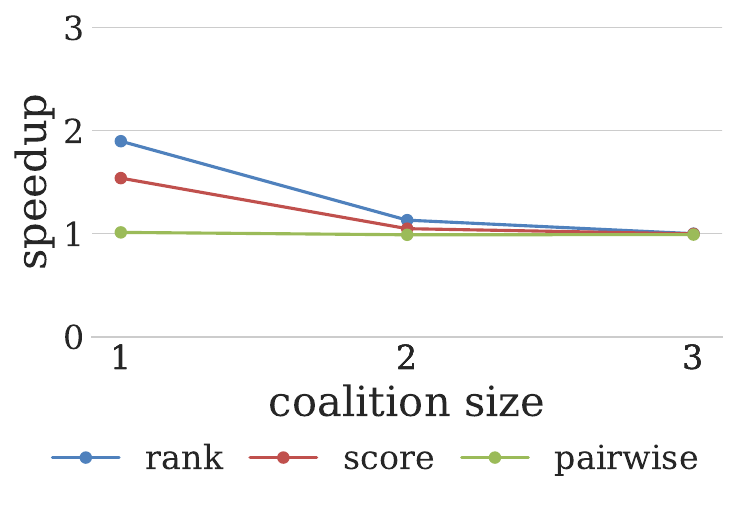}
            \label{img:CSR-coalition-speedup-full}}
        \hfill
        \subfloat[Speedup vs. maximum coalition size, ATP]{
        \includegraphics[width=0.55\columnwidth]{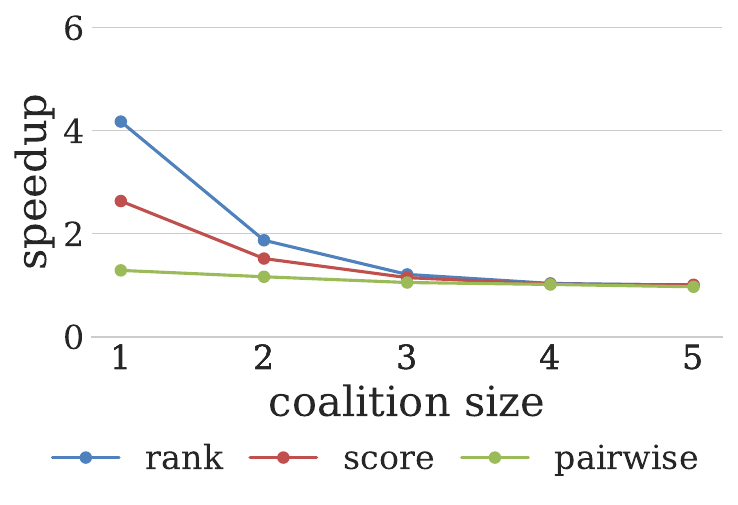}
            \label{img:ATP-coalition-speedup-full}}
        \hfill
        \subfloat[Speedup vs. maximum coalition size, THE]{
        \includegraphics[width=0.55\columnwidth]{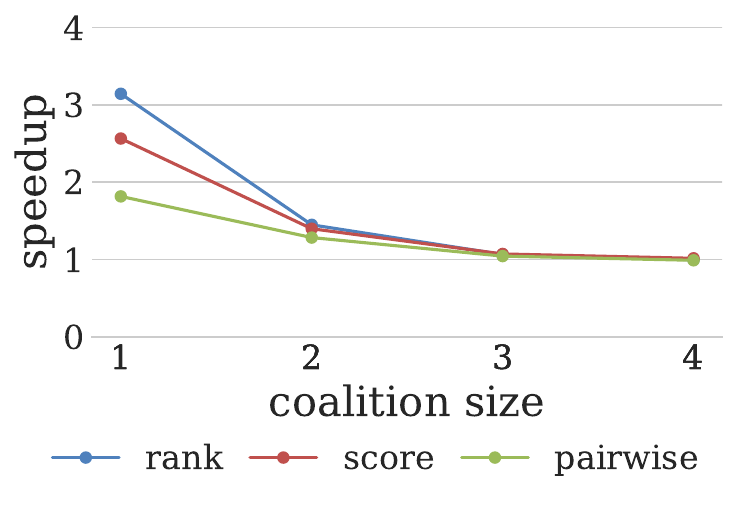}
            \label{img:THE-coalition-speedup-full}}
      \caption{Computational time performance of approximation for \csr (CSR), ATP Tennis (ATP), and Times Higher Education (THE).  Speedup is computed in comparison to exact computation times, reported in Table~\ref{tab:time:exact}.}
    \label{fig:time-approx-full}
\end{figure*}

\begin{figure*}[h!]
    \centering
        \subfloat[Fidelity vs. maximum sample size, CSR]{
        \includegraphics[width=0.55\columnwidth]{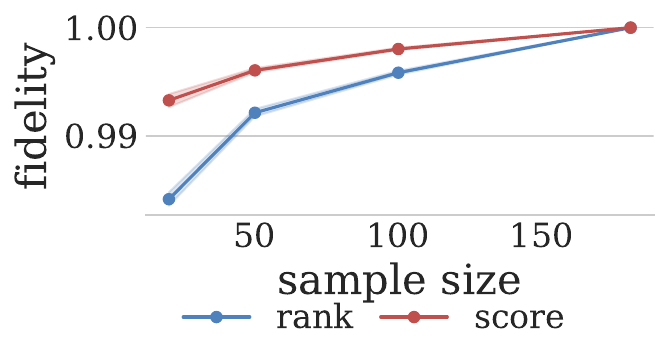}
            \label{img:CSR-sample-fidelity-full}}
        \hfill
        \subfloat[Fidelity vs. maximum sample size, ATP]{
        \includegraphics[width=0.55\columnwidth]{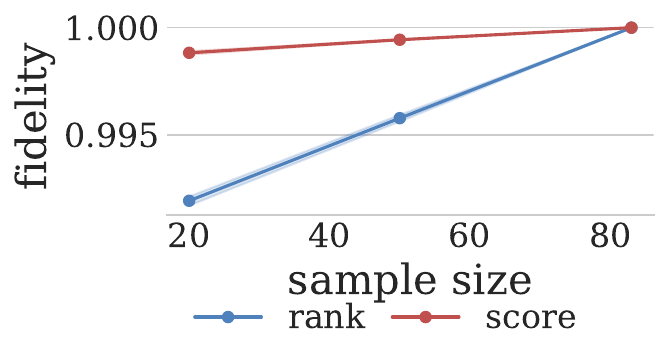}
            \label{img:ATP-sample-fidelity-full}}
        \hfill
        \subfloat[Fidelity vs. maximum sample size, THE]{
        \includegraphics[width=0.55\columnwidth]{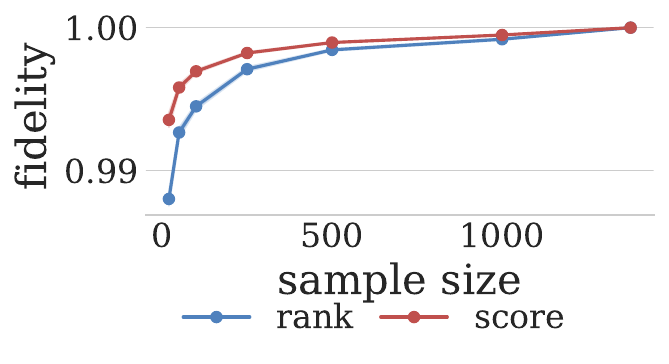}
            \label{img:THE-sample-fidelity-full}}
        \hfill
        \subfloat[Fidelity vs. maximum coalition size, CSR]{
        \includegraphics[width=0.55\columnwidth]{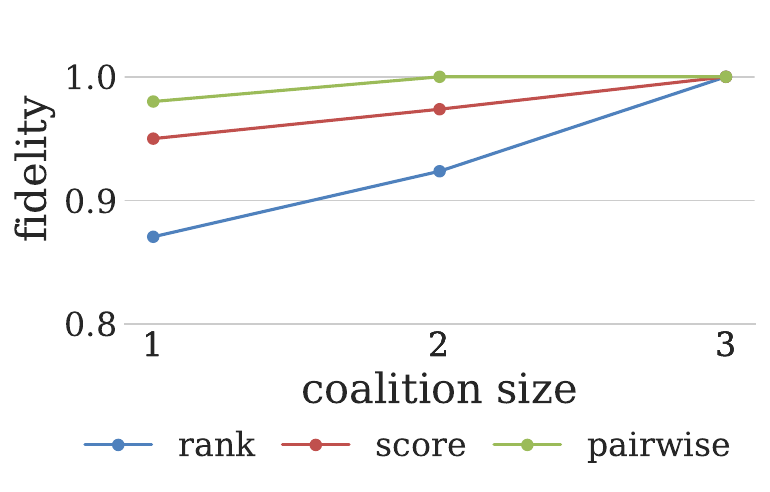}
            \label{img:CSR-coalition-fidelity-full}}
        \hfill
        \subfloat[Fidelity vs. maximum coalition size, ATP]{
        \includegraphics[width=0.55\columnwidth]{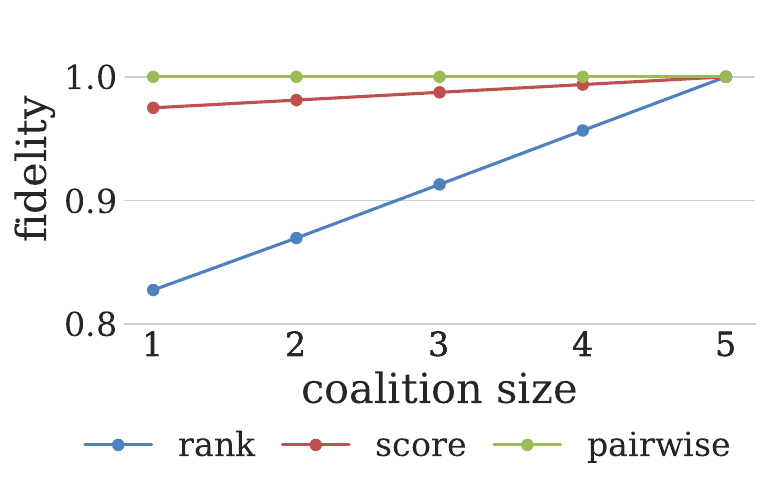}
            \label{img:ATP-coalition-fidelity-full}}
        \hfill
        \subfloat[Fidelity vs. maximum coalition size, THE]{
        \includegraphics[width=0.55\columnwidth]{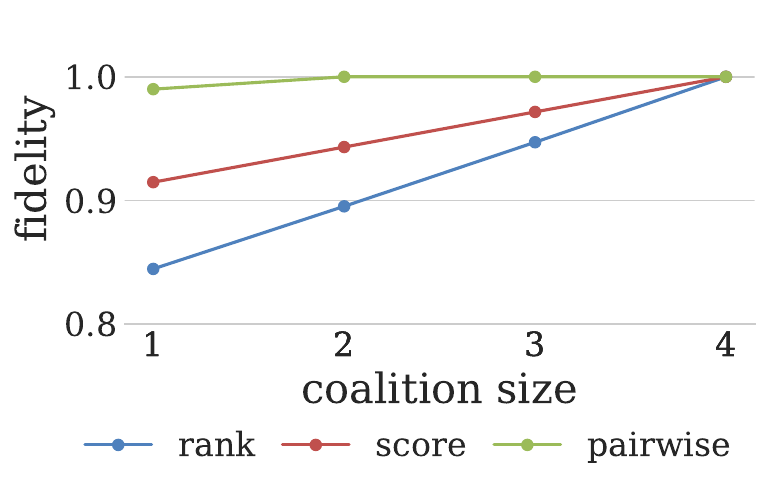}
            \label{img:THE-coalition-fidelity-full}}
      \caption{Fidelity of approximation for \csr (CSR), ATP Tennis (ATP), and Times Higher Education (THE) varying sample sizes and maximum coalition sizes.}
    \label{fig:fidelity-approx-full}
\end{figure*}

In this section, we present the extended results previewed in Section~\ref{sec:exp:efficiency:approx}.

In Table~\ref{tbl:acs-ak-times-full} we include the running times of~\sys for ACSIncome, AK, when varying the maximum coalition size or the sample size. As discussed in ~\ref{sec:exp:efficiency:approx}, we include both cold and warm start results, and the fidelity for each setting. Fidelity is high for any sample size for this dataset, and while it declines more when varying the coalition size, it remains over 0.8 for both the score and rank QoIs for any coalition size and is over 0.9 for both QoIs for coalition size 5 and above.

In Figure~\ref{fig:time-approx-full} we present the speedup vs. sample size, and speed-up vs max coalition size for THE, CSR, and ATP. We already presented the results for ACSIncome, AK in Figure~\ref{fig:time-approx}. We observe similar results, but scaled down due to the dataset sizes. In Figure~\ref{fig:fidelity-approx-full}, we present the corresponding fidelity for both sample size and max coalition size. 
We observe that fidelity is very high for all sample sizes, and almost identical or better to the fidelity of ACSIncome, AK for all max coalition sizes.

\begin{figure}[t!]
    \centering
        \subfloat[Rank QoI]{
		\includegraphics[width=0.5\columnwidth]
        {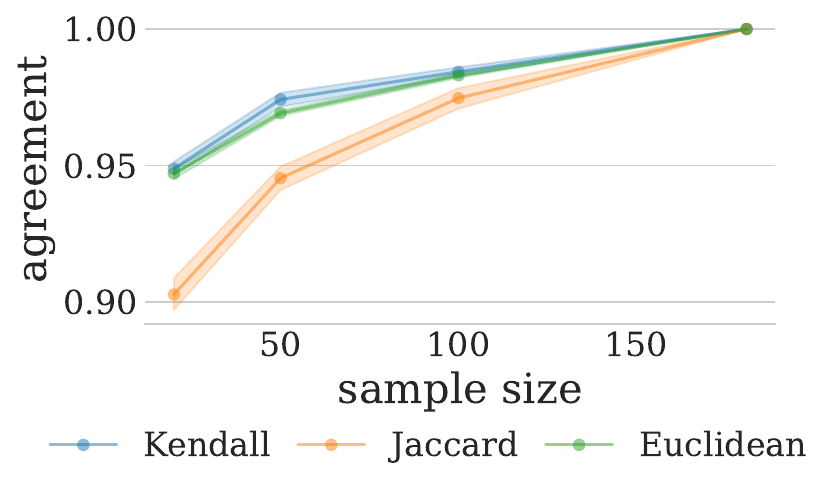}
		\label{img:CSR-samples-agreement-rank-full}}
        \hfill
        \subfloat[Score QoI]{
		\includegraphics[width=0.5\columnwidth]
        {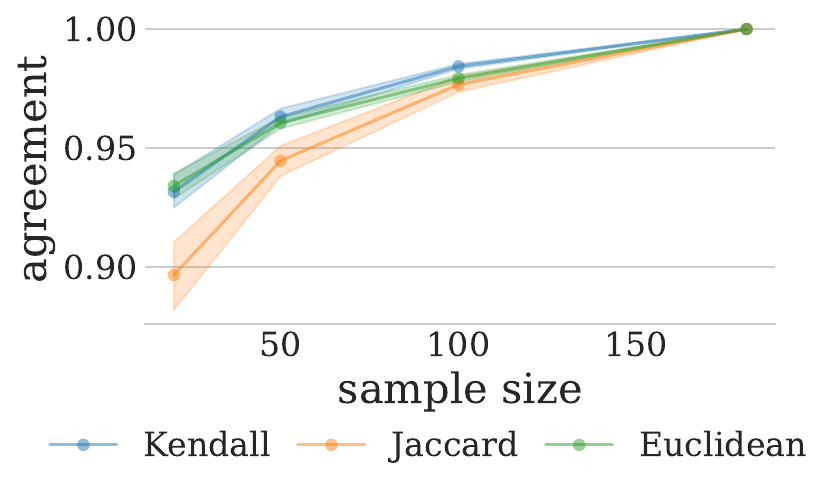}
		\label{img:CSR-samples-agreement-score-full}}
    \caption{Agreement of ShaRP for \csr when varying the sample size and using maximum coalition size for various QoIs.}
    \label{fig:CSR-agreement-sample-full}
\end{figure}

\begin{figure}[h!]
        \subfloat[Rank QoI]{
        \includegraphics[width=0.5\columnwidth]
        {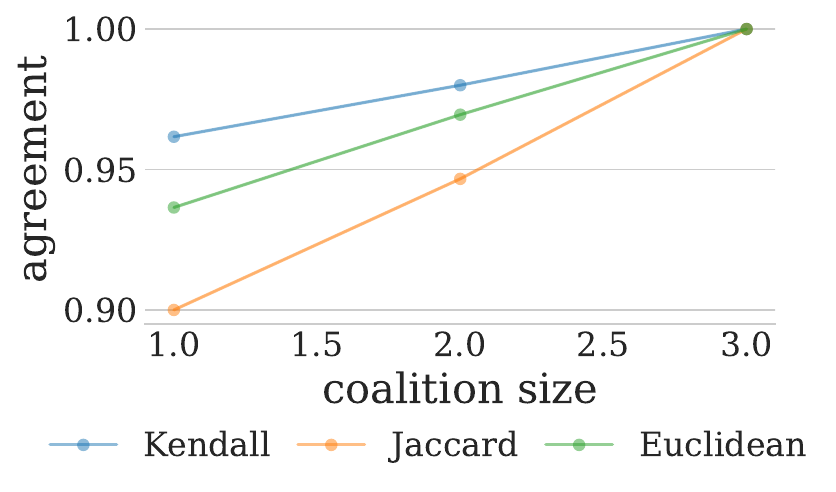}
            \label{img:CSR-coalitions-agreement-rank-full}}
        \subfloat[Score QoI]{
        \includegraphics[width=0.5\columnwidth]
        {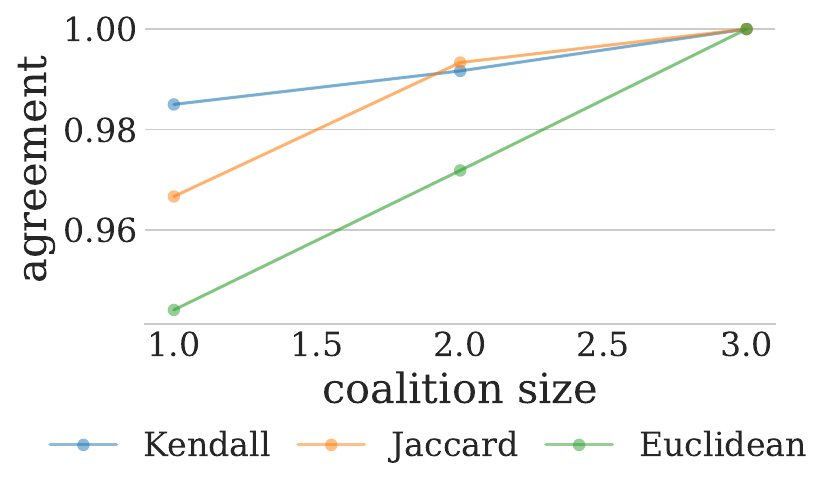}
            \label{img:CSR-coalitions-agreement-score-full}}
        \hfill
        \subfloat[Pairwise QoI]{
        \includegraphics[width=0.5\columnwidth]
        {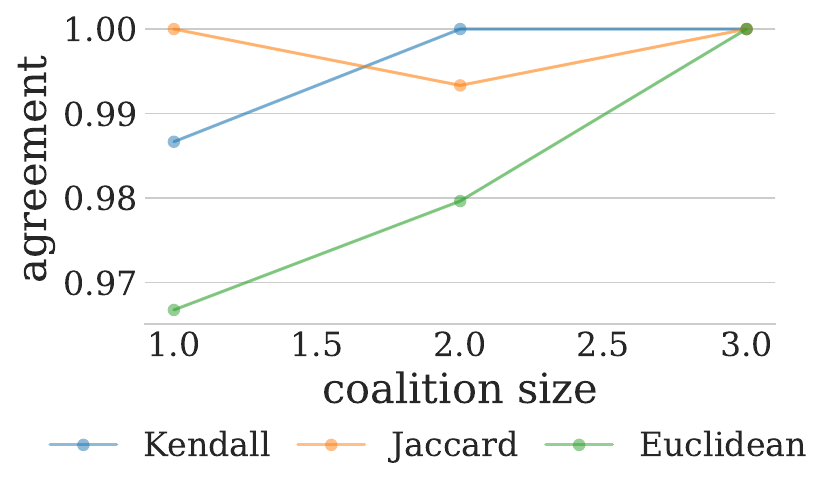}
            \label{img:CSR-coalitions-agreement-pairwise_rank-full}}
    \caption{Agreement of ShaRP for \csr when varying the coalition size and using maximum sample size for various QoIs.}
    \label{fig:CSR-agreement-coalition-full}
\end{figure}

In Figures~\ref{fig:CSR-agreement-sample-full} and~\ref{fig:CSR-agreement-coalition-full}, we present the method agreement between the approximation and the exact computation for \csr (CSR). We omit method agreement results for the other datasets, where \sys performs similarly. In~\ref{img:CSR-samples-agreement-rank-full} and~\ref{img:CSR-samples-agreement-score-full}, we present the agreement of the approximation when we vary the sample size for the rank and the score QoI. We evaluate the agreement using the Jaccard Index (considering the top-2 features), Kendall's tau distance, and the Euclidean distance of the feature vectors (converted to unit vectors). Here, we see that performance is similar for both QoIs. The Jaccard index is over 0.9 for any sample size, and is the distance metric with the worst performance for both QoIs. This is worth noting as shorter explanations are often considered more interpretable~\cite{molnar2020interpretable}. Agreement is similar or higher for all QoIs when we vary maximum coalition size, see Figure~\ref{img:CSR-coalitions-agreement-rank-full}-~\ref{img:CSR-coalitions-agreement-pairwise_rank-full}.

\section{User Study Protocol and Results}
\label{app:study_protocol}
In this section, we provide more details on the user study (NYU IRB-FY2025-9983) study described in Section~\ref{sec:focus_group}.

The goal of the study was to evaluate the usability of rank-based and score-based explanations. We conducted the study among members of our institution. For this reason, we chose \csr as the dataset since we assumed that it would be of interest to the participants. To understand the user understanding of group-based and rank-based explanations, we randomly selected a subset of the \csr schools, we produced explanations for each school using either the score or the rank QoI, we divided the participants into two groups \grpScore and \grpRank, and presented each group with a series of identical questions about the score or the rank explanations correspondingly.

In this section, we detail the study protocol in Subsection~\ref{app:study_protocol:protocol} and then we present the extended results in Subsection~\ref{app:study_protocol:results}.

\subsection{Study Protocol}
\label{app:study_protocol:protocol}
The study consisted of the four parts listed below. In this section, we provide details for each part.
\begin{enumerate}
    \item Enrollment form
    \item Introductory document
    \item Score-based or rank-based tasks
    \item Exit discussion
\end{enumerate}

\paragraph{Enrollment form} The enrollment form collected the educational background of the participants (optional text box), their highest academic degree (BS/BA, MS/MA, PhD, Other), their field of study (required text box), their relevant background (text box optional), their familiarity with AI explainability (scale 1-5, where 1 means unfamiliar and 5 expert), their familiarity with Shapley value-based methods (scale 1-5, where 1 means unfamiliar and 5 expert), and their familiarity with the \csr dataset (scale 1-5, where 1 means unfamiliar and 5 expert).

\paragraph{Introductory document} We provide the introductory document in Section~\ref{app:study_materials} and briefly summarize it here. The introductory document provided a short description of algorithmic rankers, the \csr dataset, and ~\sys, and then proceeded to explain the task. The task involves the interpretation of individual or sets of Shapley value explanations. So, using example figures, we provided information on how to read Shapley value explanations to perform tasks such as, distinguishing the features that negatively or positively impact the outcome, understanding the magnitude of the importance of a feature, understanding the metric-unit of the explanation (which depends on the QoI), and finally the Shapley value efficiency property.

\begin{figure}[b!]
        \subfloat[\grpRank category (i) visualization 1]{
        \includegraphics[width=0.45\columnwidth]
        {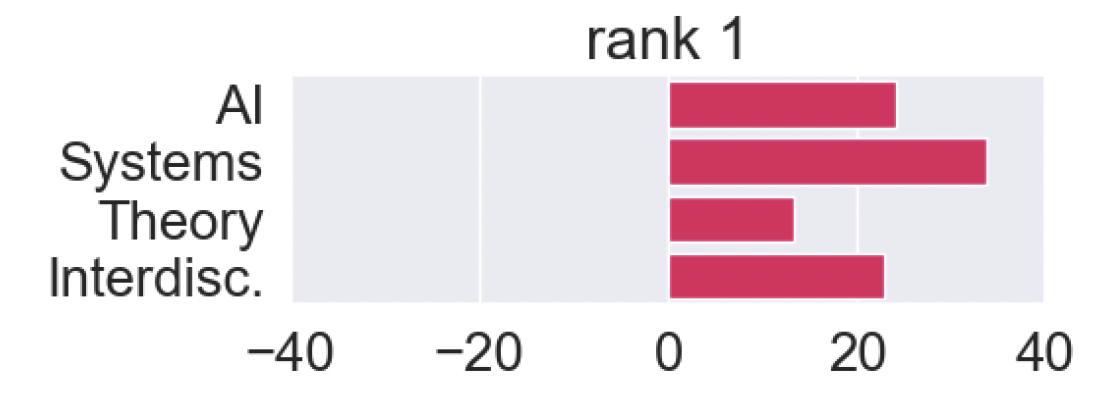}
            \label{img:study_single_rank1}}
        \hfill
        \subfloat[\grpScore category (i) visualization 1]{
        \includegraphics[width=0.45\columnwidth]
        {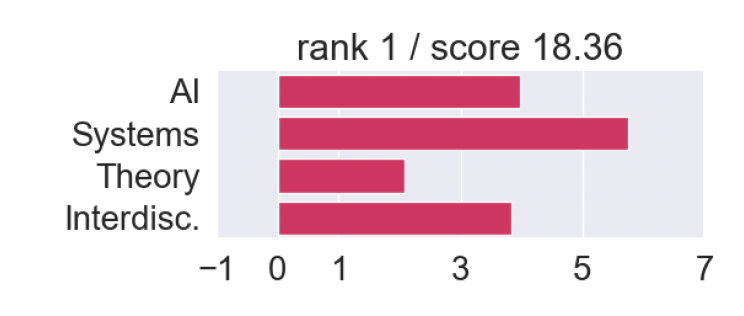}
            \label{img:study_single_score1}}
        \hfill
        \subfloat[\grpRank category (i) visualization 2]{
        \includegraphics[width=0.45\columnwidth]
        {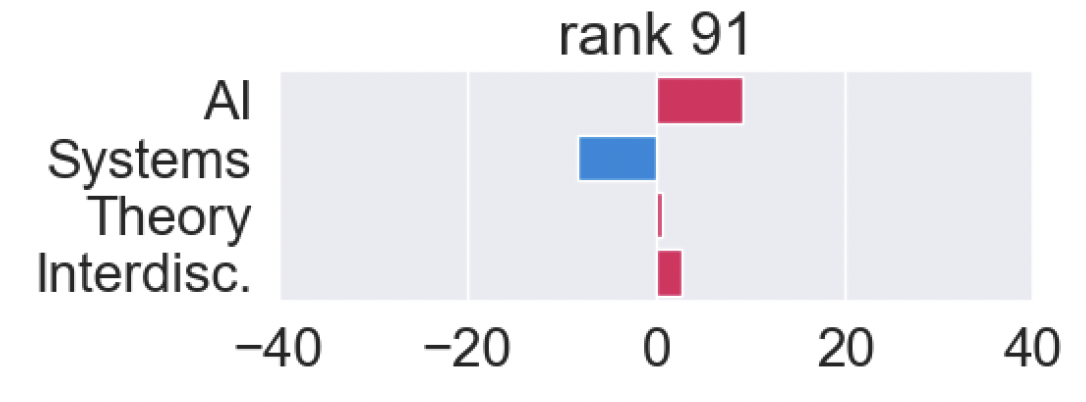}
            \label{img:study_single_rank2}}
        \hfill
        \subfloat[\grpScore category (i) visualization 2]{
        \includegraphics[width=0.45\columnwidth]
        {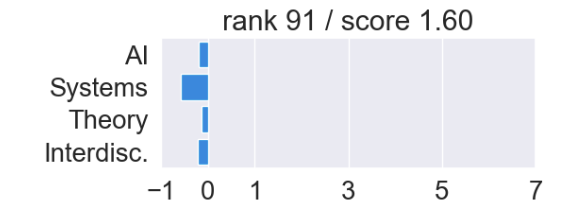}
            \label{img:study_single_score2}}
        \hfill
        \subfloat[\grpRank category (i) visualization 3]{
        \includegraphics[width=0.45\columnwidth]
        {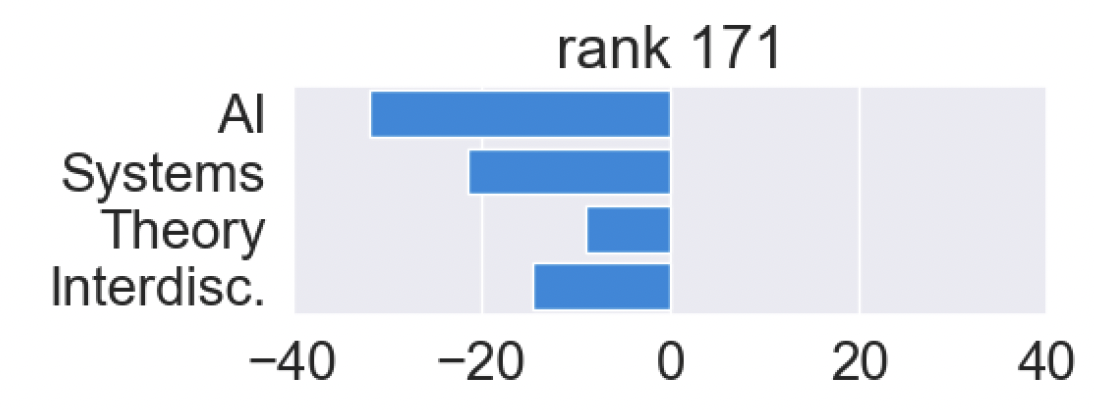}
            \label{img:study_single_rank3}}
        \hfill
        \subfloat[\grpScore category (i) visualization 3]{
        \includegraphics[width=0.45\columnwidth]
        {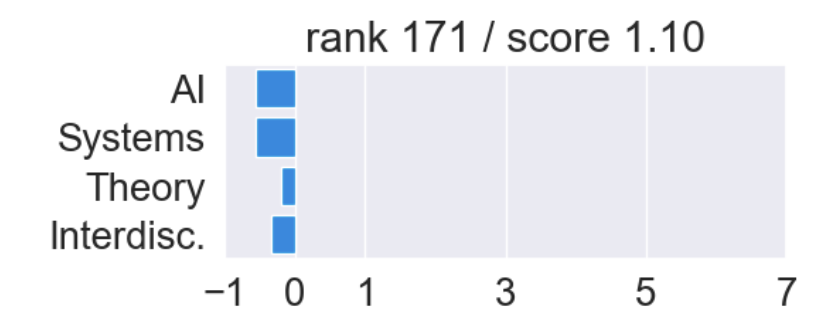}
            \label{img:study_single_score3}}
    \caption{Example figures for questions of type (i): understanding the rank of a specific department}
    \label{fig:study_images_single}
\end{figure}

\begin{figure*}[h!]
        \subfloat[\grpRank category (ii) visualization 1]{
        \includegraphics[width=0.9\columnwidth]
        {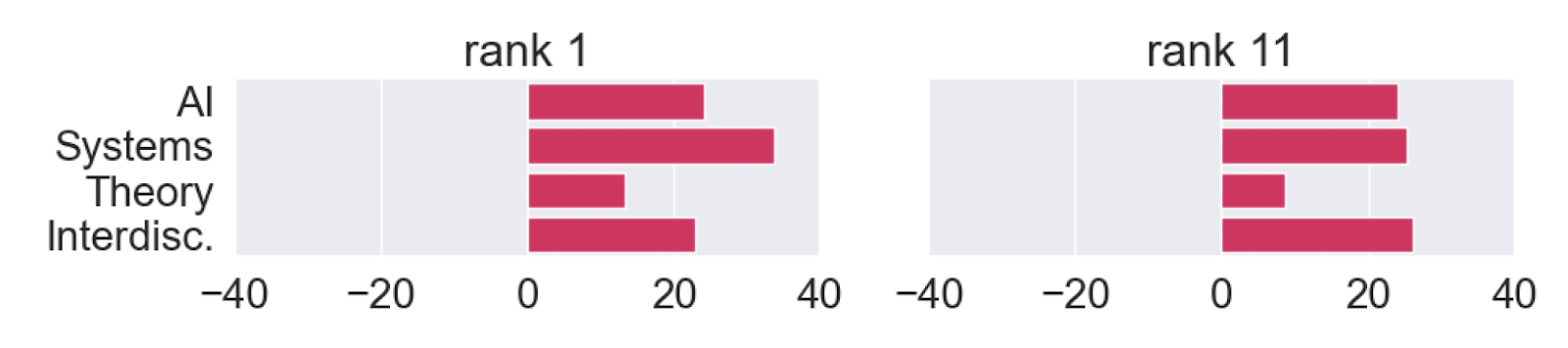}
            \label{img:study_pair_rank1}}
        \hfill
        \subfloat[\grpScore category (ii) visualization 1]{
        \includegraphics[width=0.9\columnwidth]
        {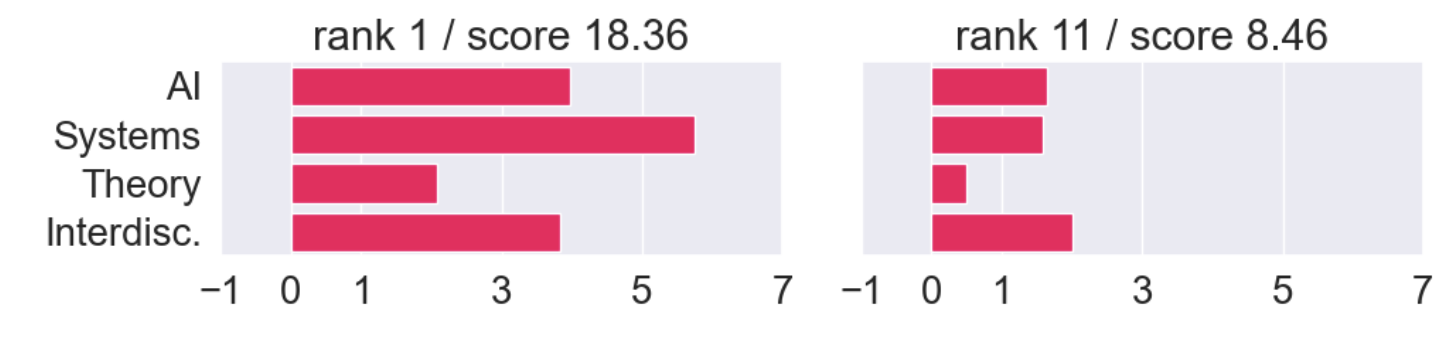}
            \label{img:study_pair_score1}}
        \hfill
        \subfloat[\grpRank category (ii) visualization 2]{
        \includegraphics[width=0.9\columnwidth]
        {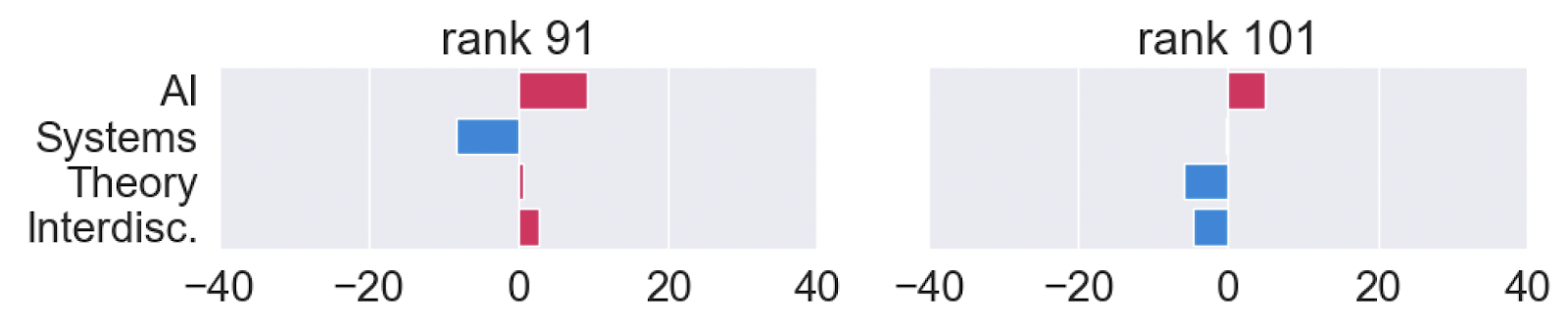}
            \label{img:study_pair_rank2}}
        \hfill
        \subfloat[\grpScore category (ii) visualization 2]{
        \includegraphics[width=0.9\columnwidth]
        {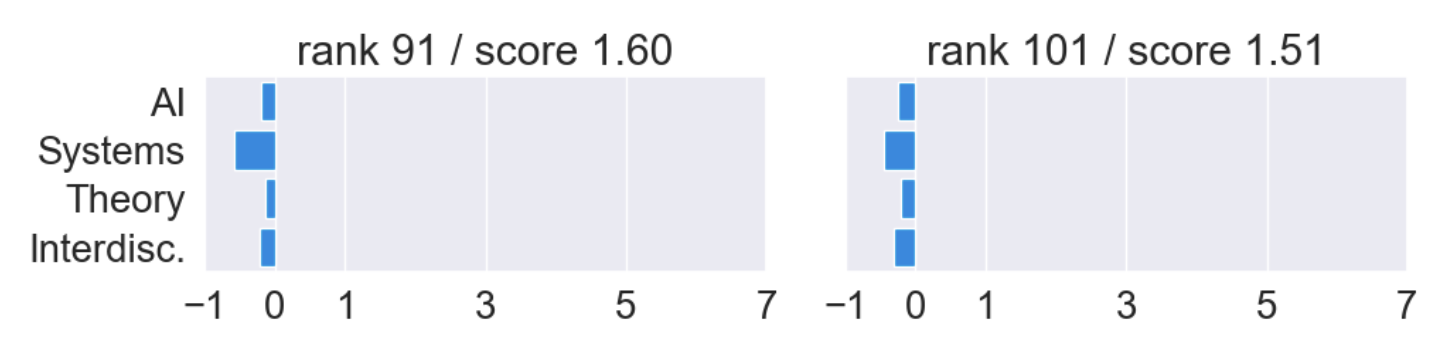}
            \label{img:study_pair_score2}}
        \hfill
        \subfloat[\grpRank category (ii) visualization 3]{
        \includegraphics[width=0.9\columnwidth]
        {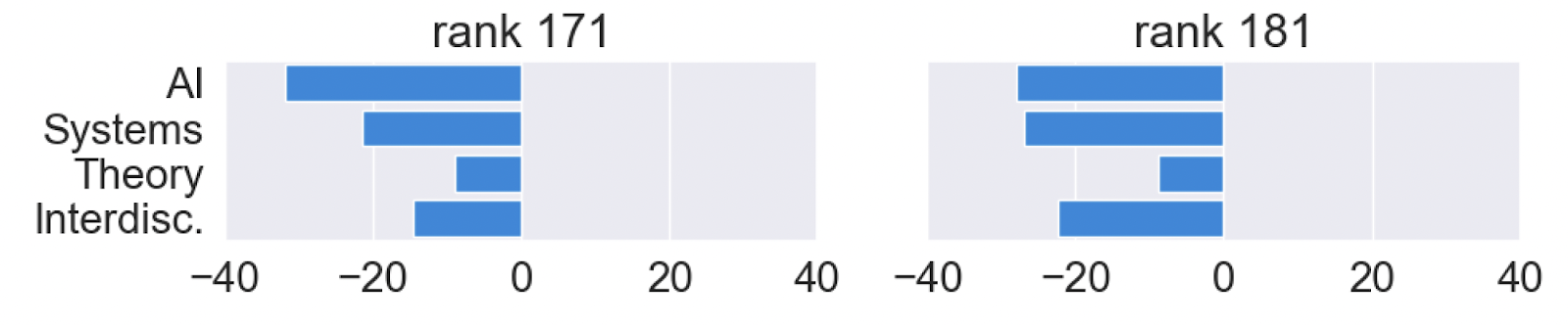}
            \label{img:study_pair_rank3}}
        \hfill
        \subfloat[\grpScore category (ii) visualization 3]{
        \includegraphics[width=0.9\columnwidth]
        {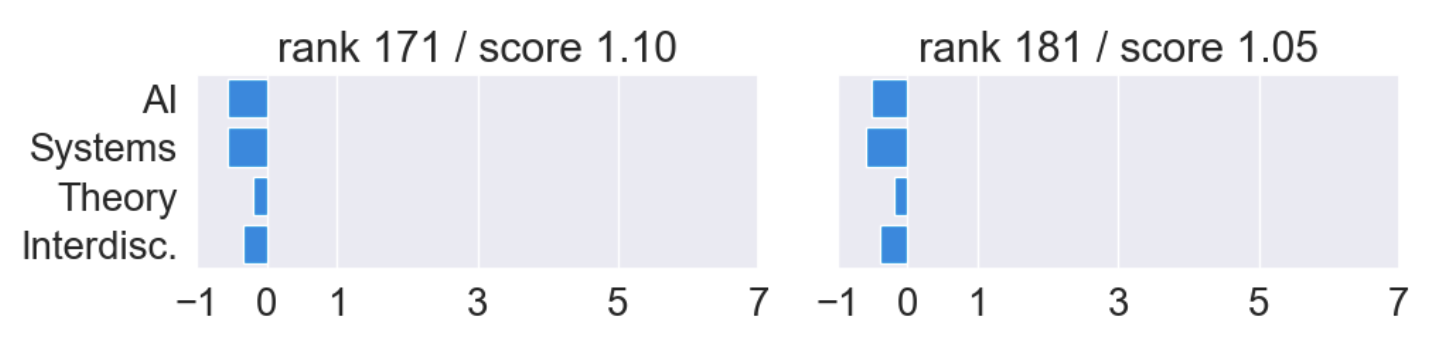}
            \label{img:study_pair_score3}}
    \caption{Example figures for questions of type (ii): understanding why one department is ranked higher than another}
    \label{fig:study_images_pairs}
\end{figure*}

\begin{figure*}[h!]
        \subfloat[\grpRank category (iii) visualization 1]{
        \includegraphics[width=0.9\columnwidth]
        {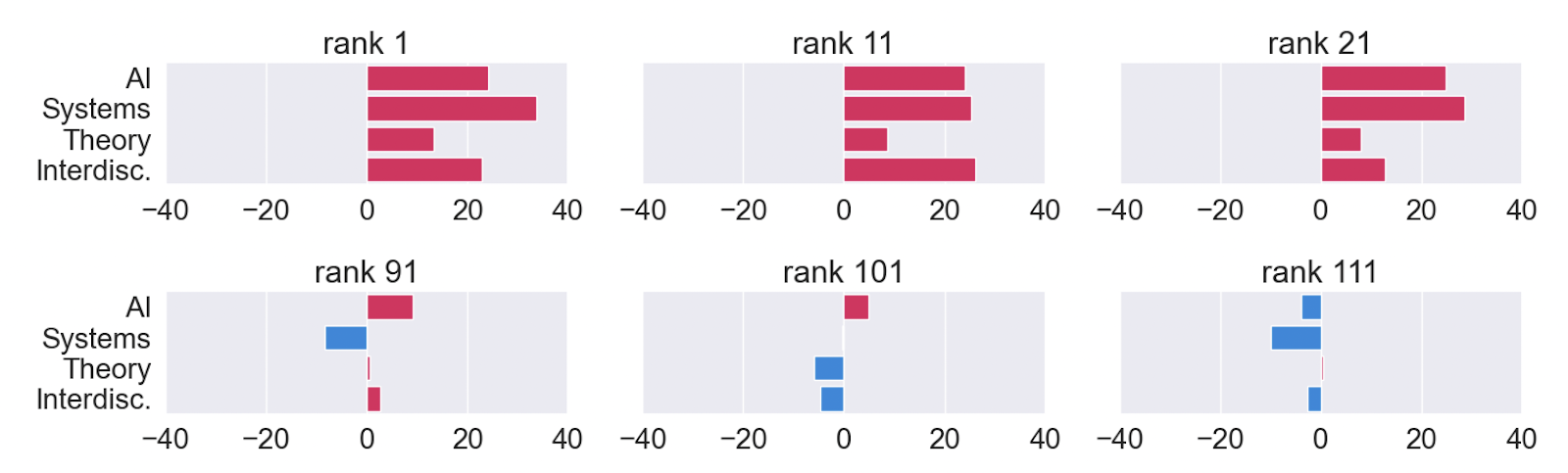}
            \label{img:study_group_rank1}}
        \hfill
        \subfloat[\grpScore category (iii) visualization 1]{
        \includegraphics[width=0.9\columnwidth]
        {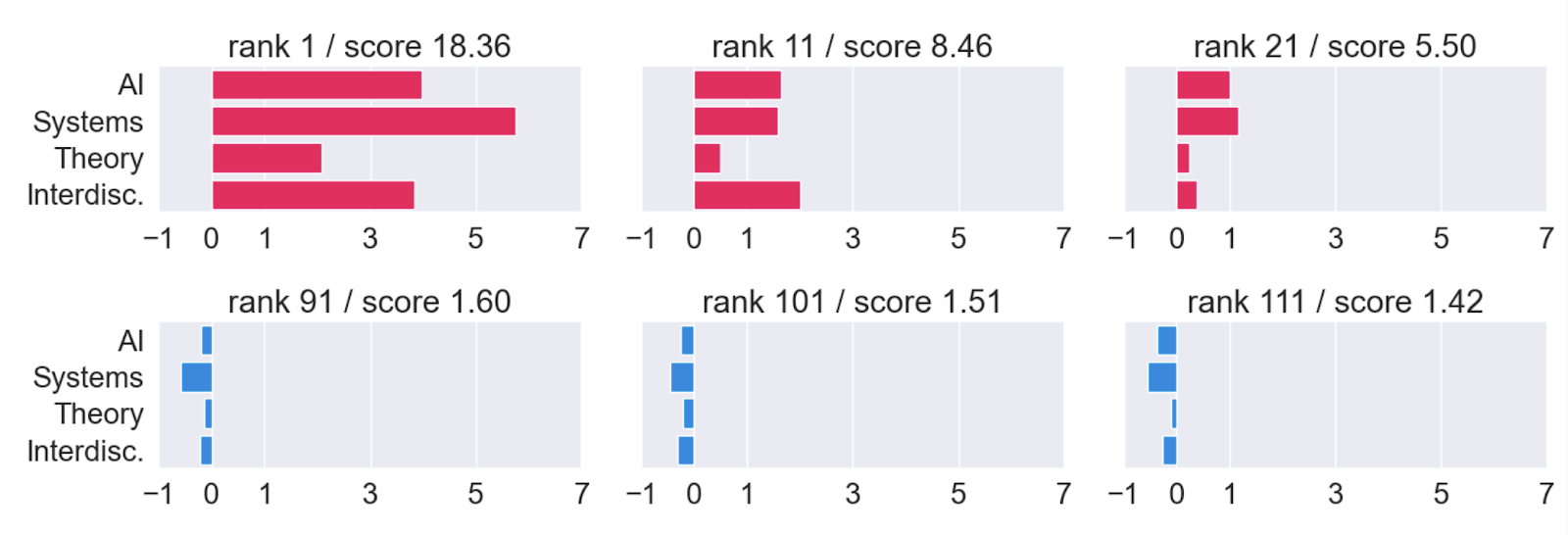}
            \label{img:study_group_score1}}
        \hfill
        \subfloat[\grpRank category (iii) visualization 2]{
        \includegraphics[width=0.9\columnwidth]
        {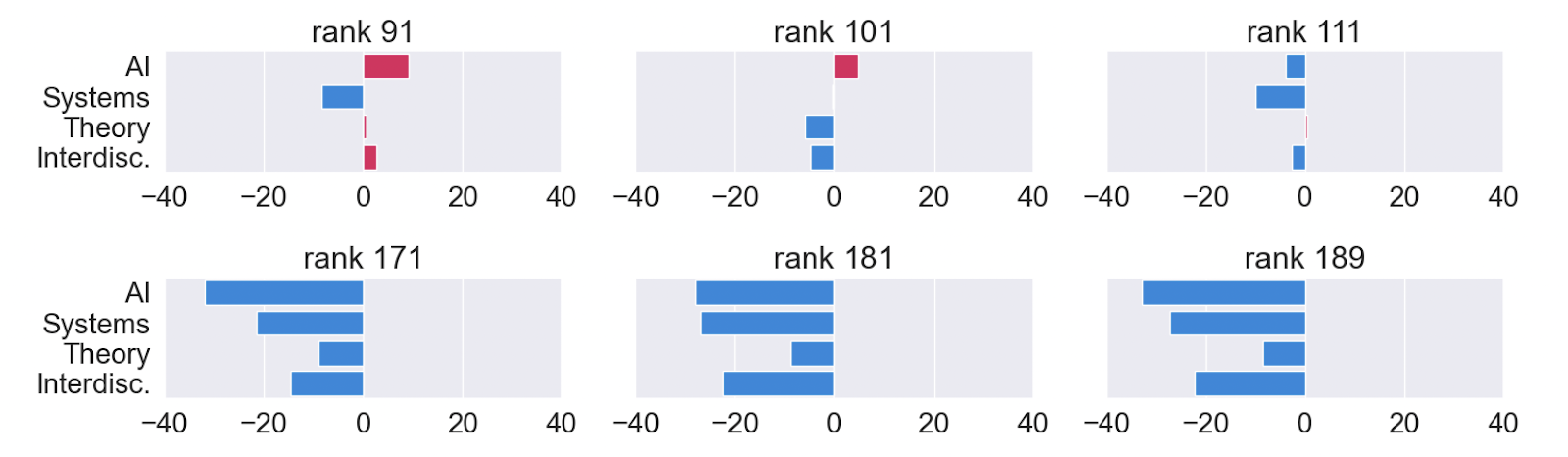}
            \label{img:study_group_rank2}}
        \hfill
        \subfloat[\grpScore category (iii) visualization 2]{
        \includegraphics[width=0.9\columnwidth]
        {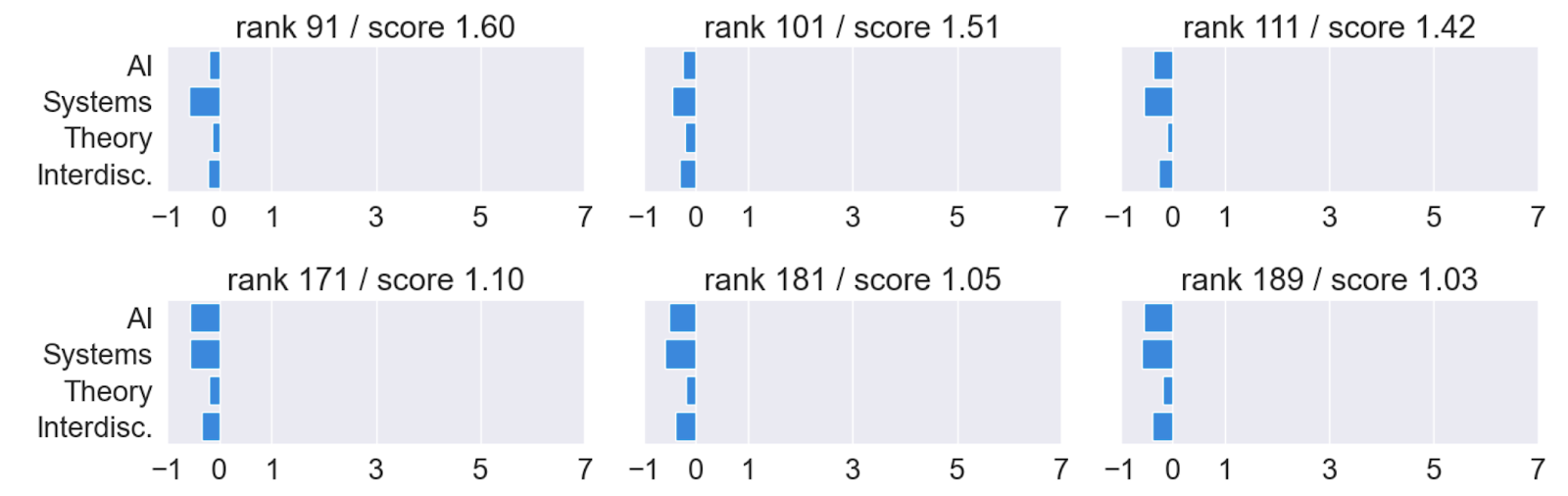}
            \label{img:study_group_score2}}
    \caption{Example figures for questions of type (iii): understanding feature importance trends across the ranking}
    \label{fig:study_images_groups}
\end{figure*}

\paragraph{Tasks} The tasks consisted of three categories. Each category had a different objective and different questions. The categories were (i) understanding the rank of a specific department (3 departments, 4 questions for each), (ii) understanding why one department is ranked higher than another (3 department pairs, 2 questions for each), and (iii) understanding feature importance trends across the ranking (2 sets of 6 departments, 2 questions for each). To select the items presented in the study, we sampled 9 universities from \csr, 3 from the top, 3 from the middle, and 3 from the bottom of the ranking at random. We generated explanations for all of them using our method and plotted them on the same axes so they are comparable.

Figure~\ref{fig:study_images_single} contains the images used in the study for the questions of type (i) for both \grpRank (left column) and \grpScore (right column). Each image was presented separately, accompanied by four questions. Each question was followed by a 5-point Likert-scale confidence question. The first question for this category asked the participants to select the feature that contributed to the department being at its respective rank the most \emph{overall}. The second asked for the feature that contributed the least \emph{overall}. The third, for the feature that contributed the most \emph{positively}. And, finally, the fourth one for the feature that contributed the most \emph{negatively}. All questions asked the participants to select the correct answer among the options. The options listed all features (AI, Systems, Theory, Interdisciplinary) and also included ``Don't know'' as an option. The last two questions also had ``No features contributed positively/negatively'' as an option.

Figure~\ref{fig:study_images_pairs} contains all pairs of images used in the study for the questions of type (ii) for both \grpRank (left column) and \grpScore (right column). Each pair of images was presented separately, accompanied by two multiple-choice questions. Each question was followed by a 5-point Likert-scale confidence question. The first question for this category asked the participants to select all features that were helping the department on the left outrank the department on the right. The second task asked the participants to select all features that were hurting the department on the left relative to the department on the right. The answers listed all features (AI, Systems, Theory, Interdisciplinary), ``None,'' and ``Don't know.''

Finally, Figure~\ref{fig:study_images_groups} contains the images used in the study for the questions of type (iii) for both \grpRank (left column) and \grpScore (right column). Each group of images was presented separately, accompanied by two multiple-choice questions. Each question was followed by a 5-point Likert-scale confidence question. The first question for this category asked the participants to select up to 2 features that were helping the departments in the top row the most in comparison to the departments in the bottom row. The second task asked the participants to select the features that were hurting the rank/score of the departments in the bottom row the least in comparison to the departments in the top row. The answers listed all features (AI, Systems, Theory, Interdisciplinary), and ``Don't know.''

\paragraph{Discussion} The last part of the user study was an open discussion that lasted approximately 30 minutes. During the discussion, prompting questions were asked, and the participants were encouraged to expand on their thoughts. The prompting questions were the following:
\begin{enumerate}
\item What are your impressions of the explanations you just reviewed?
\item Do you feel the explanations provided sufficient information to answer the questions accurately?
\item Is there any additional or alternative information you would have preferred to receive?
\item Do you have any other comments or feedback regarding the explanations or your overall experience?
\end{enumerate}

\begin{table*}[bp!]
\caption{Performance in total and for each type of question. Confidence is reported on a 5 Likert scale.}
\begin{tabular}{ c c c | c c | c c }
 & & & \multicolumn{2}{ c }{Rank-only} & \multicolumn{2}{ c }{Score-only} \\
 Type  & Visualization & Which feature(s) & \% correct & Avg. Conf. & \% correct & Avg. Conf. \\
 \midrule
 (i) & Figure \ref{img:study_single_rank1}/\ref{img:study_single_score1} & Contributed the most & 100.00\% & 4.43 & 100.00\% & 4.83 \\
 (i) & Figure \ref{img:study_single_rank1}/\ref{img:study_single_score1} & Contributed the least & 100.00\% & 4.57 & 100.00\% & 4.83 \\
 (i) & Figure \ref{img:study_single_rank1}/\ref{img:study_single_score1} & Contributed most positively & 100.00\% & 4.71 & 100.00\% & 4.83 \\
 (i) & Figure \ref{img:study_single_rank1}/\ref{img:study_single_score1} & Contributed most negatively & 85.71\% & 4.57 & 100.00\% & 4.67 \\
 (i) & Figure \ref{img:study_single_rank2}/\ref{img:study_single_score2} & Contributed the most & 85.71\% & 4.14 & 66.67\% & 4.00 \\
 (i) & Figure \ref{img:study_single_rank2}/\ref{img:study_single_score2}  & Contributed the least & 71.43\% & 4.29 & 83.33\% & 4.00 \\
 (i) & Figure \ref{img:study_single_rank2}/\ref{img:study_single_score2}  & Contributed most positively & 85.71\% & 4.57 & 83.33\% & 4.33 \\
 (i) & Figure \ref{img:study_single_rank2}/\ref{img:study_single_score2} & Contributed most negatively & 85.71\% & 4.57 & 83.33\% & 4.67 \\
 (i) &  Figure \ref{img:study_single_rank3}/\ref{img:study_single_score3} & Contributed the most & 100.00\% & 4.29 & 66.67\% & 2.50 \\
 (i) &  Figure \ref{img:study_single_rank3}/\ref{img:study_single_score3} & Contributed the least & 100.00\% & 3.71 & 100.00\% & 4.17 \\
 (i) &  Figure \ref{img:study_single_rank3}/\ref{img:study_single_score3} & Contributed most positively & 85.71\% & 4.43 & 83.33\% & 4.33 \\
 (i) &  Figure \ref{img:study_single_rank3}/\ref{img:study_single_score3} & Contributed most negatively & 85.71\% & 4.29 & 66.67\% & 3.00 \\
 \midrule
 (i) & & Total & 90.48\% & 4.38 & 86.11\% & 4.18 \\
 \midrule
 (ii) & Figure \ref{img:study_pair_rank1}/\ref{img:study_pair_score1} & Helped the 1st of the pair & 14.29\% & 4.29 & 0.00\% & 4.67 \\
 (ii) & Figure \ref{img:study_pair_rank1}/\ref{img:study_pair_score1}  & Hurt the 1st of the pair & 71.43\% & 4.14 & 0.00\% & 4.17 \\
 (ii) & Figure \ref{img:study_pair_rank2}/\ref{img:study_pair_score2}  & Helped the 1st of the pair & 100.00\% & 3.86 & 66.67\% & 4.17 \\
 (ii) & Figure \ref{img:study_pair_rank2}/\ref{img:study_pair_score2}  & Hurt the 1st of the pair & 100.00\% & 3.86 & 83.33\% & 4.17 \\
 (ii) & Figure \ref{img:study_pair_rank3}/\ref{img:study_pair_score3}  & Helped the 1st of the pair & 57.14\% & 4.00 & 16.67\% & 2.17 \\
 (ii) & Figure \ref{img:study_pair_rank3}/\ref{img:study_pair_score3}  & Hurt the 1st of the pair & 0.00\% & 4.00 & 50.00\% & 2.67 \\
 \midrule
 (ii) & & Total & 57.14\% & 4.02 & 36.11\% & 3.67 \\
 \midrule
(iii) &  Figure \ref{img:study_group_rank1}/\ref{img:study_group_score1}  & Helped the top row the most & 42.86\% & 3.71 & 33.33\% & 3.67 \\
 (iii) &  Figure \ref{img:study_group_rank1}/\ref{img:study_group_score1} & Hurt the bottom row the least & 14.29\% & 3.50 & 83.33\% & 3.67 \\
 (iii) &  Figure \ref{img:study_group_rank2}/\ref{img:study_group_score2} & Helped the top row the most & 42.86\% & 3.57 & 16.67\% & 3.33 \\
 (iii) &  Figure \ref{img:study_group_rank2}/\ref{img:study_group_score2} & Hurt the bottom row the least & 71.43\% & 3.71 & 83.33\% & 3.00 \\
 \midrule
 (iii) & & Total & 42.86\% & 3.63 & 54.17\% & 3.42 \\
 \midrule
 All & & Total & 72.73\% & 4.15 & 66.67\% & 3.90 \\
 \bottomrule
\end{tabular}
\label{tab:study_results}
\end{table*}
\subsection{Results}
\label{app:study_protocol:results}
We found that \grpScore performed  worse than \grpRank. The results are presented in detail in Table~\ref{tab:study_results}. \grpRank participants managed to answer correctly 73\% of the time, in contrast to 67\% for \grpScore participants. Additionally, \grpScore participants were less confident in their answers, 4.15/5 and 3.90/5 measured in a 5-Likert scale, for \grpRank and \grpScore, respectively.

Looking at the results per question category, we see that \grpRank performed better for the questions of category (i) and (ii), scoring 90\% and 57\% correctly versus 86\% and 36\%. However, they performed worse for the questions of category (iii). The reason for \grpScore performing better in the last category appears to be the second question of each group codified as ``Which feature hurt the bottom row the least?'' in the table. Our hypothesis for why this happened is that it is easier to answer this question correctly when looking at the \grpScore plots in ~\ref{fig:study_images_groups}, unlike most of the other questions. (The right answer here is ``Theory''.)

The confidence of the participants in \grpRank is overall higher for all questions. It is worth noting, however, that as the questions get harder, the confidence for either group does not accurately reflect the accuracy of their answers. For example, participants of both groups were overall more confident when answering incorrectly for the questions in category (ii).

The discussion portion of the study also yielded different results for \grpRank and \grpScore. For \grpRank, participants discussed the questions and the visualization choices. While the participants of \grpScore also mentioned these points, they additionally expressed distrust in both the ranking process and the dataset during the discussion. This is consistent with~\cite{aechtner2022comparing}, who used a school admissions dataset and showed that (score-based) SHAP exhibited greater and unexplained variability in the trust of the system by users compared to other methods.

Finally, \grpScore participants noted that a single score-based explanation provides no insight into the overall ranking process. This is expected, as score-based explanations focus solely on the score of an item, without relating it to its position in the ranking. The x-axis represents the score, and the contributions are derived from it, making it difficult to infer how ranks change. Participants emphasized that understanding the ranking process requires viewing multiple explanations. They appreciated that the study allowed them to examine several explanations at once, which helped them form a clearer understanding of how the ranking works.

In summary, our results provide preliminary evidence that rank-based explanations are a better fit for ranking tasks as compared to score-based explanations.  We are working to refine the user study protocol based on participants' feedback and to scale up the sample size to observe clearer trends.

\section{User Study Additional Materials}
\label{app:study_materials}
In this section, we present the introductory materials used for the user study described in Section~\ref{sec:focus_group}.

\includepdf[pages=-]{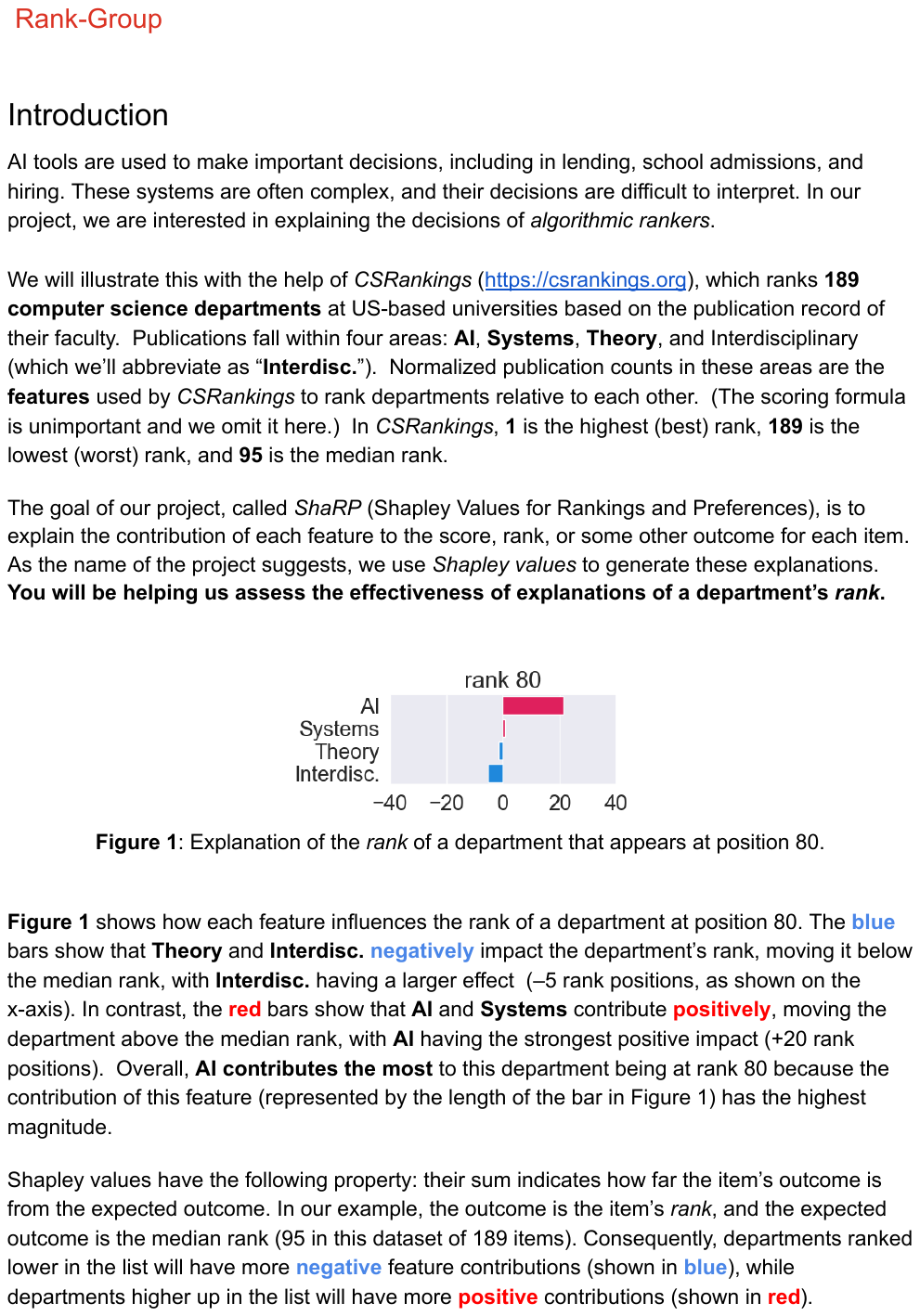}
\includepdf[pages=-]{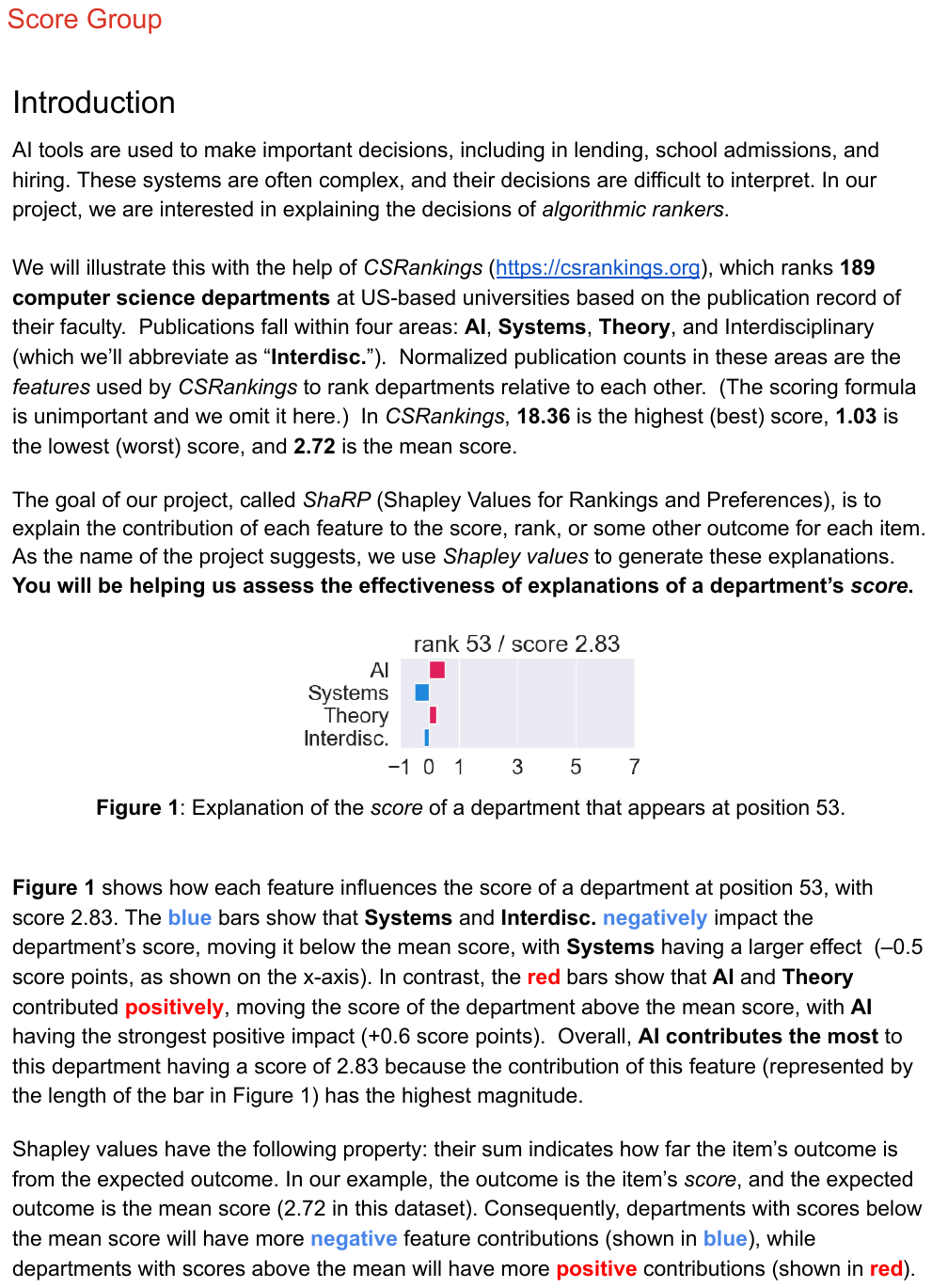}
\end{document}